%% file: fair_template.tex
\PassOptionsToPackage{table}{xcolor}
\documentclass[]{fairmeta}

\input{math_commands.tex}

\usepackage{hyperref}
\usepackage{url}
\usepackage{graphicx}
\usepackage{subcaption}
\usepackage{multirow}
\usepackage{xcolor}
\usepackage{xspace}
\usepackage{tabularx}
\usepackage{cleveref}
\usepackage{booktabs}
\usepackage{tabularx}
\usepackage{array}
\usepackage{microtype}

\usepackage[textsize=scriptsize,textwidth=1.15in,linecolor=gray!60,bordercolor=gray!60]{todonotes}
\usepackage{marginnote}
\usepackage[most]{tcolorbox}
\usepackage{listings}
\usepackage{pifont}
\usepackage[normdefault,type1]{sourceserifpro}

  \newtcblisting{promptbox}[1]{
      enhanced,
      breakable,
      listing only,
      title={#1},
      colback=black!2,
      colframe=black!25,
      coltitle=black,
      fonttitle=\bfseries,
      boxrule=0.6pt,
      arc=2pt,
      left=1mm,
      right=1mm,
      top=1mm,
      bottom=1mm,
      listing options={
          basicstyle=\ttfamily\footnotesize,
          breaklines=true,
          columns=fullflexible,
          keepspaces=true,
          showstringspaces=false
      }
  }
  
\definecolor{TableHeader}{HTML}{EAF1F8}
\definecolor{TableStripe}{HTML}{F7F9FC}
\definecolor{GainColor}{HTML}{087F5B}

\definecolor{darkblue}{rgb}{0, 0, 0.5}
\hypersetup{
  colorlinks = true,
  linkcolor  = darkblue,
  citecolor  = darkblue,
  urlcolor   = darkblue,
}

\title{Don’t Forget What You Just Learned: Self-Directed Repair Turns Agent Mistakes Into Lasting Lessons}

\author[1,2,*]{Yuval Kansal}
\author[2]{Karen Ullrich}
\author[2]{Polina Kirichenko}
\author[2]{Ilia Kulikov}
\author[1]{\textbf{Niraj K. Jha}}
\author[2]{\textbf{Mohammad Pezeshki}}
\author[2]{\textbf{Mark Ibrahim}}

\affiliation[1]{Princeton University}
\affiliation[2]{Meta FAIR}

\contribution[*]{Work done during an internship at Meta FAIR.}
\makeatletter
\DeclareRobustCommand{\sidecomment}[3]{%
  \ifx\@captype\undefined
    \todo[color=#1!12,bordercolor=#1!70]{\textcolor{#1}{\textbf{#2:}} #3}%
  \else
    \marginnote{\raggedright\scriptsize\textcolor{#1}{\textbf{#2:} \textit{#3}}}%
  \fi}
\makeatother

\newcommand{\resampling}{base resampling\xspace}

\newcommand{\trajretry}{Failed Trajectory in Context\xspace}

\newcommand{\icl}{Reflections in Context\xspace}

\newcommand{\method}{%
  {\fontfamily{SourceSerifPro-LF}\fontseries{mb}\selectfont
   \textls[15]{Self-Direction}}%
  \xspace%
}
\abstract{
Intelligent agents should learn from their mistakes so that experience leads to continuous improvement. Yet today’s agents recover from a mistake only temporarily:  in-context reflections are easily forgotten across sessions and compete for limited context, while full-parameter fine-tuning is expensive and can interfere with existing capabilities.
We introduce \method, a mechanism for agents to learn from mistakes that combines the benefits of temporary in-context retries and policy updates. When an agent fails a task, it reflects on its own trajectory in order to extract a lesson and immediately distills it into a lightweight low-rank adapter.  
By keeping the original weights frozen, \method prevents adapter training from overwriting base-model parameters while training additional parameters equivalent to $\sim 0.02\%$ of the base model's parameter count for each adapter.
The agent then saves these adapters in a persistent library of lessons,
 automatically routing them to solve related problems later. Experiments with agents based on open-weight models from the Qwen family across open-ended research (GAIA), tool-use workflows (Claw-Eval), and terminal tasks (Terminal-Bench~2.1) show that our method improves overall success rates by 8.9 to 17.0 percentage points and performs competitively with frontier systems. 
We find that \method reduces repeated mistakes across future encounters of a task, nearly doubling the reliability of an agent's corrections. Learned lessons also transfer: a handful of adapters generalize to dozens of new tasks without additional training.
Ultimately, this self-directed repair transforms everyday mistakes into lasting lessons.

}

\date{October 20, 2026}
\metadata[Website]{\url{https://self-direction.github.io/}}
\correspondence{\email{yuvalkansal@princeton.edu}, \email{marksibrahim@meta.com}}
\begin{document}

\maketitle

% \owner{Yuval then Karen first round by tomorrow 9am; then Mohammad + Mark tomorrow morning}

% \vspace{-0.2cm}
\begin{figure}[h]
    \centering
    \includegraphics[width=\linewidth]{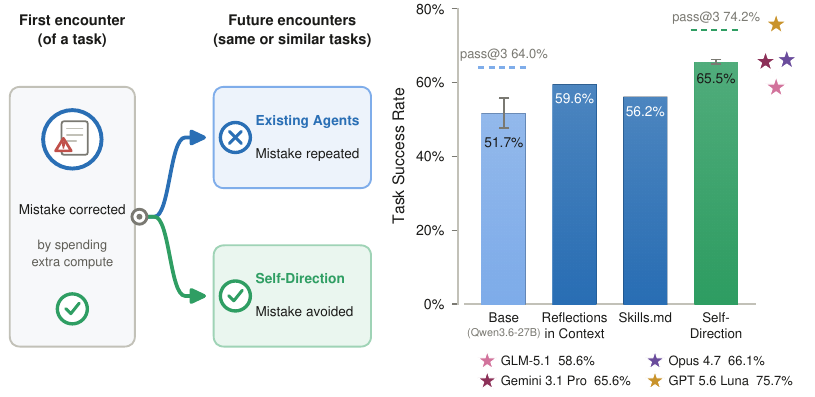}
    \caption{\textbf{\method turns an agent's own failures into reusable LoRA
    lessons.} \textbf{Left:} Current agents fail to reliably reuse lessons learned by spending extra resources correcting a mistake. \method enables storage and reuse of lessons to avoid repeated mistakes. \textbf{Right:} On Terminal-Bench 2.1, \method lifts performance from 51.7\% to 65.5\%, outperforming existing baselines and frontier models\protect\footnotemark. %\MP{What is base? Is it retry or initial try?}
    }
    \label{fig:headline}
\end{figure}
\section{Introduction}
\label{sec:intro}

Modern agents can often correct their mistakes and complete the task at hand. 
% \Mark{failure or mistake?}
Yet, this recovery is ephemeral: In a new session, the agent is just as likely to repeat the same mistake 
\citep{rabanser2026towards, gonzalez2026reliability}. As agents tackle longer-horizon workflows, each additional step introduces another opportunity for error. Even frontier agents make mistakes across tool use, coding, and general assistant tasks \citep{yuan2026osworld2, bandi2602mcp}. Repeatedly correcting the same mistakes is costly, leads to inconsistent performance, and leaves the potential for continuous improvement unrealized. Our goal is not just to fix an agent’s mistake in the current session, but to \textit{learn persistent lessons to avoid repeating the mistake in the future}.

Existing approaches encapsulate lessons either in context or in model weights, but enable only part of the desired behavior. Context-based methods, e.g., demonstrations, prompt optimization, reflection, hard-coded skill files, and persistent agent harnesses, can change behavior quickly during deployment \citep{shinn2023reflexion, majumder2023clin, zhao2024expel}. They enable agents to accumulate experience without access to model parameters, but their lessons remain external: They must be retrieved when relevant, compete for limited context, and may not reliably influence behavior on related future tasks.

Conversely, parameter-based methods, such as supervised fine-tuning and reinforcement learning, can produce more persistent behavioral change, but typically require separate training processes, substantial data and optimization, and careful management of interference to reduce the risk of catastrophic forgetting \citep{uoEmpiricalStudyCatastrophic2025,gModelEditingHarms2024,venContinualLearningCatastrophic2025}.
% They have enabled agents to accumulate experience even when model parameters are inaccessible. Their lessons, however, remain external; they must be retrieved and reintroduced when relevant\Mark{Combine this and previous sentence to highlight the shortcoming directly}, compete for limited context, and may not consistently influence behavior on related future tasks \citep{wynterInContextLearningLearning2026,minRethinkingRoleDemonstrations2022,uFantasticallyOrderedPrompts2022,iuLostMiddleHow2023,atil2024non}. Conversely, parameter-based methods, such as supervised fine-tuning and reinforcement learning, can produce more persistent behavioral change, but typically require separate training processes, substantial data and optimization, careful management of interference, and are prone to catastrophic forgetting \citep{uoEmpiricalStudyCatastrophic2025,gModelEditingHarms2024,venContinualLearningCatastrophic2025}. 
What is missing is a continuous mechanism to fix mistakes in the current session and avoid repeating mistakes in future sessions, without impacting existing capabilities.

To address this gap, we introduce \textbf{\method}, a mechanism for agents to learn from their mistakes. After a failed execution, the agent examines its trajectory to extract a lesson and then distills the lesson into a lightweight Low-Rank Adaptation (LoRA) repair adapter. \method turns feedback about a mistake from the environment or a user into a LoRA lesson that can be employed immediately to retry the current task and retained for future use. \method uses a router to determine if an existing lesson is relevant to the current task based on the similarity between the current task description and descriptions of previously seen tasks, enabling reuse of lessons on future tasks. Each lesson is encoded in a separate adapter while the base model and previously learned adapters remain frozen, preventing new training from overwriting existing parameters. Furthermore, each lesson requires only a modest amount of memory (e.g., $\sim 85$ MB for Qwen3.6-27B), allowing the library of lessons to grow with the agent's experience. An overview of \method is presented in Fig.~\ref{fig:pipeline-overview}.

% Occupying the sweet spot between transient textual retry reflection and computationally heavy full-model rollouts, this targeted in-weight repair is both persistent and safely contained. The base model preserves its general capabilities without catastrophic forgetting, while a growing library of modular LoRA adapters accumulates specialized experience, dynamically selected by a routing mechanism when relevant. \Mark{do we / can we show this easily with what we have? This would be a great addition to connect the work to continual learning}. Self-direction thus makes failure productive, transforming an unsuccessful execution into a capability that remains available after the episode ends.
\footnotetext{Frontier model numbers were obtained from \url{https://www.tbench.ai/?version=2.1}.}

We evaluate \method on open-weight agents across terminal coding, tool-use, web search, and personal assistant tasks.   
We find:

\begin{enumerate}
    \item \textbf{\method increases success across task types.}  By repairing mistakes, \method raises the task success rate by 17.0 percentage points on GAIA \citep{mialonGAIABenchmarkGeneral2023}, 13.8 points on Terminal-Bench~2.1 \citep{merrillTerminalBenchBenchmarkingAgents2026}, and 8.9 points on Claw-Eval \citep{yeClawEvalTrustworthyEvaluation2026} (tool-using workflows), outperforming in-context reflections and reflections distilled into \texttt{Skills.md}. On Terminal-Bench~2.1, official scores show that \method with a 27B-parameter Qwen agent outperforms GLM-5.1, a model with approximately $30\times$ more parameters, and performs competitively with frontier closed-weight systems including Claude Opus 4.7 and Gemini 3.1 Pro (see \Cref{sec:results}).
    
    \item \textbf{\method prevents repeated mistakes across future encounters of a task.} \method repairs are reliable across repeated encounters. It nearly doubles the recovery rate across three encounters compared with base-agent retrying. On Claw-Eval, \method raises $\mathrm{pass}^3$, which measures success in all three independent runs, from $51.6\%$ to $64.6\%$, achieving performance competitive with that of frontier agents that use GPT-5.4, Kimi-K2.6, and Gemini 3.1 Pro (see \Cref{sec:results-reliability}).

    \item \textbf{\method generalizes to new tasks without sacrificing existing capabilities.} With a library capped at just three adapters, occupying 256 MB of memory in total, \method raises task success on GAIA from 36.4\% to 52.7\%. Lessons learned exclusively from easier GAIA tasks improve success on unseen, harder tasks by 10.5 points, whereas three adapters learned from Claw-Eval operations tasks raise success on workflow tasks from $75.0\%$ to $87.5\%$, without additional training. Notably, \method enables these gains without interfering with existing capabilities (Section~\ref{sec:backward-transfer}).
    
\end{enumerate}

Together, these results bring open-weight agents closer to continuous improvement, where lessons accumulate with continued use.

\section{Related Work}
\label{sec:related-work}
% \owner{Karen first pass on harness; Mohammad second pass on continual learning}

% An agent that fails on a task can always be run again. Retrying by independently sampling reliably increases coverage, the fraction of tasks solved through at least one attempt \citep{brown2024large}. Allocating more compute at inference time can substitute for a larger model \citep{snell2025scaling}. Neither approach retains any information. Once the episode ends, the effort spent recovering from failure is discarded and must be repeated. Work that does retain this information differs primarily in where the lesson is kept: in the context the agent reads, in the harness that runs it, or in the weights themselves.

% \vspace{-0.2cm}
After failing to complete a task, an agent could simply re-run the same task \citep{brown2024large} or spend additional inference-time compute \citep{snell2025scaling}. Both methods improve the fraction of tasks solved by at least one attempt. However, these methods retain no lesson from the recovery after the episode ends, leaving the agent just as likely to repeat the same mistake. Methods that do retain lessons differ mainly in where the lesson is stored: in the agent’s context, its harness, or its weights.

% \paragraph{Lessons kept in context.} The most direct way to preserve a lesson is to write it down. Approaches like Reflexion \citep{shinn2023reflexion}, ExpeL \citep{zhao2024expel}, CLIN \citep{majumder2023clin}, and Voyager \citep{wangVoyagerOpenEndedEmbodied2023} maintain verbal reflections, causal abstractions, or executable skill files in an episodic memory that conditions later trials. While these methods enable rapid adaptation without parameter updates, their lessons remain external to the policy. Consequently, reuse depends on continuous retrieval and fitting text into finite context windows – processes that are notoriously fragile. In-context performance is highly sensitive to prompt composition \citep{uFantasticallyOrderedPrompts2022} and models use relevant information less reliably as contexts grow \citep{iuLostMiddleHow2023}. We provide a more comprehensive discussion of the expressive limits of context-based adaptation in Appendix \ref{sec:appendix_related_work}.

% \vspace{-0.3cm}
\paragraph{Lessons kept in context.} The most direct way to preserve a lesson is to write it down. Reflexion \citep{shinn2023reflexion}, ExpeL \citep{zhao2024expel}, CLIN \citep{majumder2023clin}, and Voyager \citep{wangVoyagerOpenEndedEmbodied2023} store verbal reflections, causal abstractions, or executable skills in episodic memory to condition later trials. This enables rapid adaptation without parameter updates but leaves lessons external to the policy: Reuse requires repeatedly retrieving them into a finite context window. This process is fragile, as performance is sensitive to prompt composition \citep{uFantasticallyOrderedPrompts2022} and models use relevant information less reliably in longer contexts \citep{iuLostMiddleHow2023}. Appendix \ref{sec:appendix_related_work} further discusses the expressive limits of context-based adaptation.

\paragraph{Lessons kept in the harness.}
An orthogonal approach modifies the agent's harness---its prompts, tools, and memory---while leaving the model fixed \citep{chen2026harnessx, ning2026code}. Recent systems search over agent designs or self-improve by rewriting their code in response to environmental feedback \citep{hu2025adas, zhang2026dgm, robeyns2025sica}. Closest to our setting, \citet{zhang2026selfharness} let a fixed model mine its failure traces to propose minimal, regression-tested harness edits. We share this failure-driven, teacherless trigger but instead incorporate repairs into the policy itself. Harness-only methods change \emph{how} a fixed policy is invoked, not \emph{what} it has learned; gains remain in the scaffold, leaving the underlying model no more capable after deployment. Appendix \ref{sec:appendix_related_work} reviews self-synthesizing and evolving harnesses more broadly.

\paragraph{Lessons kept in weights.} Parameter updates write lessons directly into the policy, though full fine-tuning can interfere with existing capabilities \citep{uoEmpiricalStudyCatastrophic2025}. Parameter-efficient methods, such as LoRA \citep{uLoRALowRankAdaptation2021}, constrain these updates, motivating approaches that convert failures into weights. FailForge \citep{lvFailForgeDistillingProcedural2026} and SKILL0 \citep{uSKILL0InContextAgentic2026}, however, typically rely on global fine-tuning, privileged reference solutions, or stronger external teachers. Test-Time Training (TTT) instead adapts parameters online to the current input \citep{sun2020testtime, akyurek2025surprising} but applies updates unconditionally and discards them after the episode. \method differs from both: Triggered only by a verified failure, it derives supervision from the agent's own failed trajectory without a stronger teacher, then uses a targeted adapter \citep{ren2026scaling} to store each repair as a routable entry in a persistent library of lessons. Appendix \ref{sec:appendix_related_work} compares other parameter-efficient, self-rewarding, and TTT methods.

\section{Learning from Mistakes}
\label{sec:method}

% Intelligent agents should treat a mistake not as a cost to be incurred again on the next similar task, but as an opportunity to acquire a persistent, reusable lesson. 
% \vspace{-0.2cm}
Consider an agent governed by an initial policy $\pi_0$ deployed across a task set $\mathcal{T}$. During execution, the tasks on which the agent fails are denoted by a mistake subset $\mathcal{T}_{fail}$. When an agent encounters a task $t \in \mathcal{T}_{\text{fail}}$, the goal is to repair the immediate mistake while updating the system so that similar mistakes are avoided in the future and base capabilities are preserved. 
%\MP{Fixed Mark's terminology points. Settled on: a `mistake' is the event, we `repair' it, and the thing we store is a `skill'. Dropped `dead end' and `failure recovery'. So I updated this subsection, the Fig.~\ref{fig:gaia-headline} caption, and one line in the abstract. Left `failure' alone where it is already part of a term we use (failure-conditioned, failed trajectory).} 

% When an agent makes a mistake, we aim for it to repair the mistake now, learn from the mistake to avoid repeating similar mistakes in the future, and not sacrifice existing capabilities. We consider various approaches for repairing a mistake:
% \vspace{-0.2cm}
\subsection{Existing Paradigms for Repairing Mistakes}
\begin{figure*}[t]
      \centering
      \includegraphics[width=0.92\textwidth]{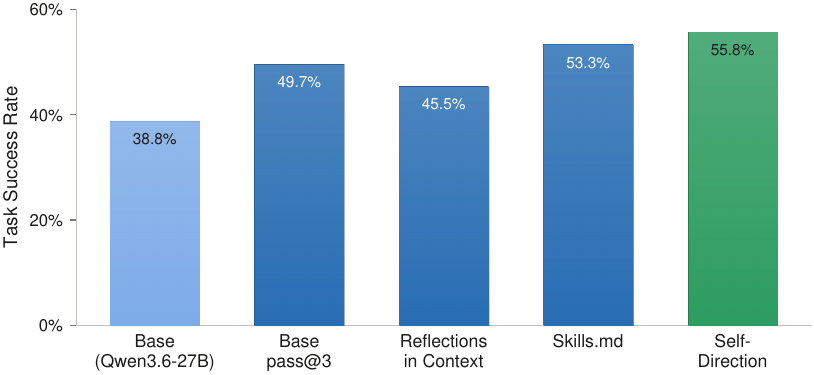}
      \caption{\textbf{\method outperforms alternative mistake-repair strategies on GAIA.} SVD-initialized LoRA achieves success on $92/165$ tasks ($55.8\%$), outperforming Retrying (pass@3), \icl, and Accumulating Reflections in Skills.md. \method recovers 28 mistakes, yielding a $17.0$-point gain over the initial base execution. We also compare our results with \trajretry and doubling the tool call budget in Appendix~\ref{app:gaia-by-level}.}
      \label{fig:gaia-headline}
      % \vspace{-0.3cm}
\end{figure*}

% \vspace{-0.2cm}
\label{sec:repair-paradigms}
Motivated by current mechanisms for repairing mistakes outlined in Section~\ref{sec:related-work}, we consider four strategies. They differ in where lessons are encoded and whether they persist after the episode: nowhere, in the context, in the weights, or in the harness (see Fig.~\ref{fig:gaia-headline}). 
%\MP{Reframed around where the lesson is kept, matching the Related Work split. These are four discrete places rather than points on a curve.}

\begingroup
\setlength{\leftmargini}{1.2em}
\begin{itemize}
    \item \textbf{Resampling} \textit{(lesson not stored)}: Agents spend extra test-time compute by drawing multiple independent samples from the unmodified base model $\pi_0$ 
    %in hopes of generating 
    to hopefully generate a new, correct rollout~\citep{muennighoffS1SimpleTesttime2025}. No lesson is stored, even in the current session: Each agent action is selected from the same policy, $\pi_0$. 
    \item \textbf{Reflection in Context} \textit{(lesson stored in current session's context)}: Reflecting on mistakes and inserting verbal critiques, error rationales, or raw failed trajectories into the prompt context has been shown to be an effective method for boosting agent performance~\citep{shinn2023reflexion, majumder2023clin}. By self-reflecting and loading task-specific procedural lessons into context, agents can avoid mistakes when retrying the task. However, by default, the lesson is forgotten once a new session without the current context begins.
    \item \textbf{Reflection in Weights} \textit{(lessons persist across sessions in weights)}: As an alternative to loading procedural per-task reflections in context as textual lessons, agents can incorporate the same reflections in weights through a lightweight adapter, which enables the lessons to persist in the policy and can lead to corrective behavior.
    \item \textbf{Accumulating reflections in \texttt{Skills.md}} \textit{(lessons are accumulated across sessions in a text file)}: Instead of relying on an external mechanism to determine relevant in-context reflections for a task and retrying, agents can learn from mistakes by distilling and accumulating reflections across sessions  and tasks in a shared Markdown text file, often called \texttt{Skills.md}. Agents can load the relevant \texttt{Skills.md} into context when the same or similar tasks are encountered. 
\end{itemize}
\endgroup

% Retrying the same task can solve the current task success, but does not persist the repair---when the agent encounters the same task again, it is just as likely to repeat its original mistake. In contrast, when an agent repairs a mistake after reflecting on the mistake, the reflection can be persisted through the model's context or by updating the model's weights. As reflections accumulate, however, reflections compete for limited context and can interfere with existing capabilities when incorporated in-weights. 
% \vspace{-0.2cm}
We compare these approaches using open-weight agents across general assistant, coding, and tool-calling agentic tasks.

% \vspace{-0.3cm}
\subsection{Experimental Setup and Evaluation on GAIA}
\label{sec:setup}

% \vspace{-0.2cm}
\paragraph{Experimental setup.} To evaluate mechanisms for repairing mistakes under controlled conditions, we standardize our primary agent architecture on open-weight models from the Qwen family. The main part of this article contains results for Qwen3.6-27B \citep{qwen3.6-27b}. The agent operates inside a ReAct interaction loop using standard benchmark system prompts outlined in Appendix~\ref{app:prompts} \citep{yao2022react}. Appendix~\ref{app:hyperparameters} lists all hyperparameters, Appendix~\ref{app:prompts} lists the prompts, and Appendix~\ref{app:cross-model-results} repeats the loop on the stronger Qwen3.8-27B (with additional results for Qwen3.5-9B) \citep{qwen38}. Open-weight models enable us to directly compare in-weight versus in-context reflections, described in Section~\ref{sec:repair-paradigms}, while keeping the agent's base model fixed. 

% \vspace{-0.2cm}
\paragraph{Metrics.} In addition to accuracy, measured as the \emph{Task Success Rate}, we report \emph{pass@$k$} and \emph{pass$^k$} to separate coverage from consistency across $k$ independent executions of the whole pipeline. Pass@$k$ is the fraction of tasks solved in at least one run; pass$^k$ is the fraction solved in every run.
%\MP{Sec.~\ref{sec:results-efficiency} says $73.9\%$ (Claw-Eval) and $66.3\%$ (Terminal-Bench~2.1) but Sec.~\ref{sec:results-breadth} is $73.3\%$ and $65.5\%$.}

% \paragraph{Metrics.} To measure improvements, we report \emph{Task success} determined by the criteria of each benchmark. \emph{Agent rollouts per task} measure computational cost, counting task-environment executions only and excluding reflection, synthesis, and adapter training \Mark{Why do we exclude adapter training?}. We compare the loop's accuracy and rollout count against the base pass@$k$ curve. \emph{Pass@$k$ and pass$^k$} separate coverage from consistency across $k$ independent executions of the whole pipeline: pass@$k$ counts a task solved in at least one run, pass$^k$ only when solved in every run. \MP{Sec.~\ref{sec:results-efficiency} says $73.9\%$ (Claw-Eval) and $66.3\%$ (Terminal-Bench~2.1) but Sec.~\ref{sec:results-breadth} is $73.3\%$ and $65.5\%$.}

% \vspace{-0.2cm}
\paragraph{Evaluating assistant, tool-calling, and coding agents.} Our evaluation spans three distinct domains. \textbf{GAIA} \citep{mialonGAIABenchmarkGeneral2023} comprises 165 validation tasks across three difficulty levels. Agents use web search, page retrieval, Python, a shell, and file inspection; success is verified by matching a short final answer against a reference. \textbf{Claw-Eval} \citep{yeClawEvalTrustworthyEvaluation2026} evaluates tool-use workflows over mock email, calendar, task-tracking, and business services. Verification considers the whole trace, including tool calls, audit logs, and produced artifacts, together with an LLM judge. A task can therefore fail because of the actions the agent took even when its final message looks correct. \textbf{Terminal-Bench~2.1} \citep{merrillTerminalBenchBenchmarkingAgents2026} comprises 89 tasks executed through a persistent shell inside a fresh container. Success is verified by hidden tests run after the attempt. We provide additional experiments for browser agents on MiniWoB++ in Appendix~\ref{app:additional-browser-benchmarks}. Additional details on benchmark-specific tool calls are presented in Appendix~\ref{app:benchmark-tools}.

In this section, we employ GAIA as a diagnostic testbed to motivate our pipeline design, reserving comprehensive cross-domain evaluations on Claw-Eval and Terminal-Bench~2.1 for Section~\ref{sec:results}.
\begin{figure}[t]
    \centering
    \includegraphics[width=0.95\linewidth]{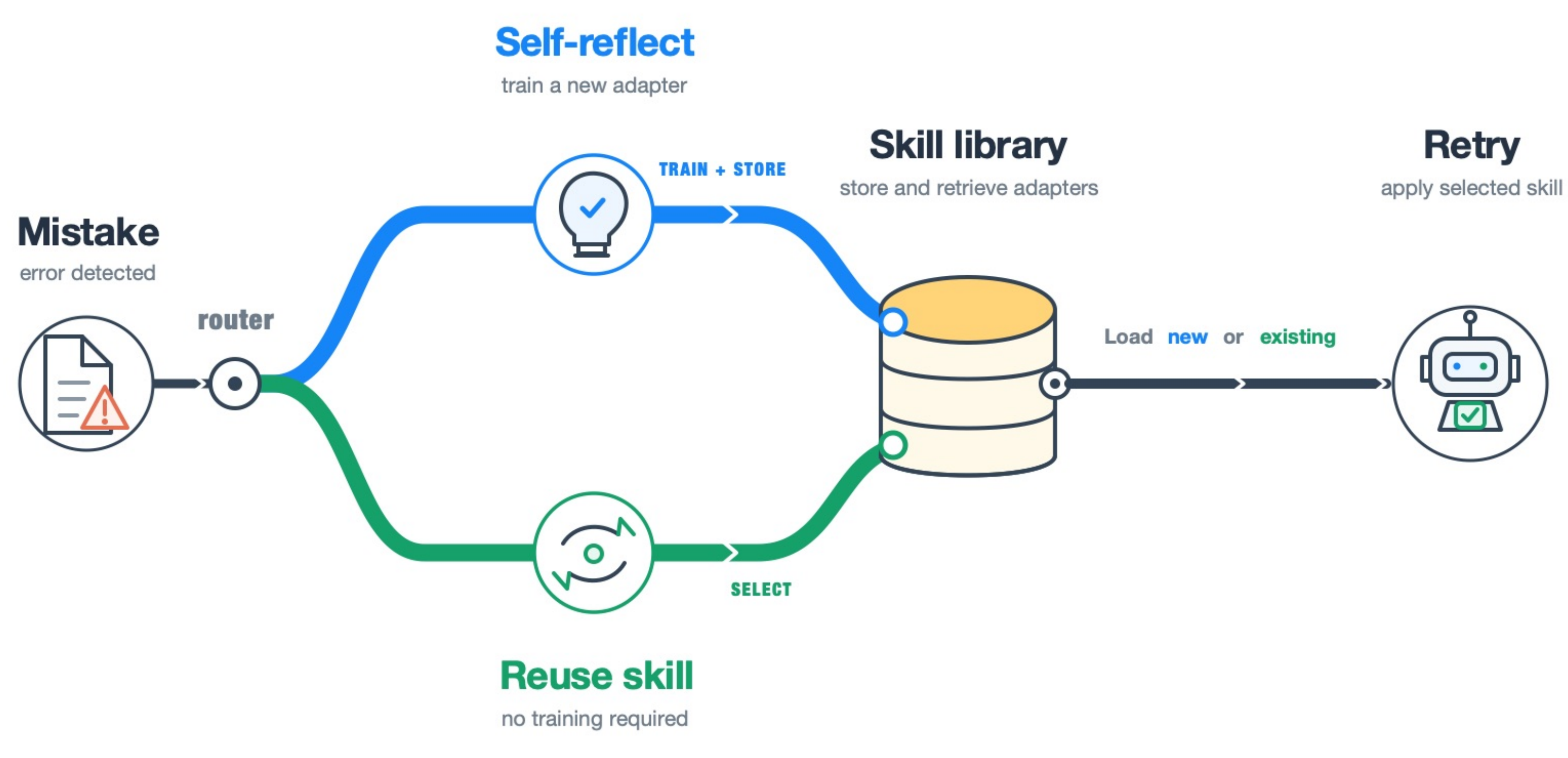}
    \caption{\textbf{\method pipeline overview.} On detecting a mistake, the agent either self-reflects on the
    failed trajectory and trains a new LoRA adapter or retrieves an adapter already stored in the library of lessons; the selected adapter is then loaded and the task retried.}
    \label{fig:pipeline-overview}
    % \vspace{-0.3cm}
\end{figure}

% \paragraph{Baselines.} We compare retrying, retrying with reflections in context, retrying with reflection in-weights via LoRA updates, and retrying with reflections distilled into a skill.md file. We also compare training a single LoRA adapter across reflections from many mistakes.  
% \vspace{-0.4cm}
\paragraph{Findings on GAIA.} We evaluate task success across standard retrying, in-context reflections, and in-weight parametric updates on the 165 GAIA validation tasks (see Fig.~\ref{fig:gaia-headline}).
\label{sec:gaia-results}
% We find on GAIA tasks, retrying is an effective method for boosting task success with a gain of $+9.5$pp with three retires compared to the base agent as shown in \Cref{fig:cross-domain-headline}. However, after correcting a mistake by retrying, agents are just a likely to repeat mistakes in the future. 

% (moved Fig.~\ref{fig:gaia-headline} earlier to appear on page 4)

% \subsection{Self-Direction retries with in-weight repairs}

% retrying is great, but agents are just a likely to repeat mistakes on future encounters.
% we search for the best way to persist skills learned from mistakes: in-weight is better; LoRA is sufficient
% accumulatng skills can be tricky (saturated context if raw lessons are loaded or not as good if distilled into a skill), so we use a selector and load LoRA. This also prevents any interference with existing capabilties as mistakes accumulate allowing us to build a larger library of agents.
% we incorproate the power of retrying by adding a base fallback.

\textit{Retrying boosts coverage but lacks persistence:} Resampling from the base model by doing parallel rollouts is effective for immediate task recovery, yielding a $10.9$ pp gain with pass@3 compared to the initial base execution. 
% \MP{Three names for one baseline: "Retrying with Reflections in Context",  "Reflections in Context", "in-context learning".} 
%\MP{Why maj@8 here, when always comparing against base pass@k? That is two different resampling baselines for the same thing. My worry is that maj@8 gives the baseline no verifier, while our loop needs one to notice the mistake in the first place, so a reviewer will say we picked the comparator we beat. Yuval, do we have GAIA pass@8?} 
However, retrying leaves model parameters untouched; once an episode ends, the agent remains just as likely to repeat the same mistake when encountering a similar task in future sessions. We also compare our results with \trajretry in Appendix~\ref{app:gaia-by-level}. 

\textit{In-weight updates outperform in-context reflections:}  To enable lessons to persist, we compare adding failure reflections to context versus incorporating them in weights. Encoding reflections in weights achieves higher task success than loading them into context ($+10.3\text{ pp}$). As mistakes accumulate, we find that encoding reflections into a single LoRA adapter is not as effective as individual LoRA adapters per task (see Fig.~\ref{fig:gaia-recovery-ablation}), suggesting that LoRA updates per task are effective in capturing repairs without using global parameter updates. A qualitative example of a successful recovery from a mistake on a Level~1 task is presented in Fig.~\ref{fig:qual-gaia-example}.
% \MP{Not obvious from the get go whether we are introducing our method or another in-weight update. Here we are mixing method explanations and the findings.}

\textit{Harness summaries (Skills.md) underperform and constrain context budget:} As execution failures accumulate, storing textual reflections in context hits severe scaling limits. Storing full failure trajectories and reflections across GAIA tasks exceeds Qwen3.6-27B’s native context window by over $10\times$ (Appendix~\ref{app:context-scaling}). Distilling reflections into a high-level Skills.md file avoids context overflow but task performance is lower than when reflections are retained in weights (see Fig.~\ref{fig:gaia-headline}).

Across 165 GAIA tasks, \method improves accuracy from 38.8\% to 55.8\%, a gain of 17.0 points (paired task-level $t$-test: $t(164)=5.79$, $p=3.50\times10^{-8}$; $95\%$ CI: $[11.2, 22.8]$ pp). We also find that loading an incorrect or mismatched adapter degrades performance below the clean base policy (Appendix~\ref{app:adapter-specificity}), motivating us to include a \texttt{router} to select an adapter relevant to the task at hand based on the similarity between the current task description and the source-task descriptions associated with stored adapters. Additional details about the $t$-test are presented in Appendix~\ref{app:t-test-gaia}.

% To persist learnings we also compare adding reflections after a mistake in-context versus in-weight. 
% We find encoding reflections in-weights yields higher task success compared to adding reflections in-context (+8.9pp vs. +5.6pp for in-context); light-weight LoRA updates are sufficient (specifically SVD LoRA, see Appendix \ref{app:svd_lora}). 

% As mistakes accumulate, we find reflections in-context can exceed the limited context window with reflections from GAIA exceeding Qwen3.6's context window by $10\times$ \Cref{tab:gaia-context-capacity}. 
% We also explore distilling reflections in a \texttt{skill.md} by asking the agent to summarize a high level lesson for each mistake rather than multiple reflections. We find while \texttt{skill.md} can scale across the set of GAIA tasks, the performance is lower than adding the full set of reflections for a mistake in context as shown in \Cref{fig:cross-domain-headline}. Based on task success and context window limits, we elect to encode reflections in-weights via LoRA adapters rather than in-context for Self-Direction. Furthermore, we find loading the wrong adapter to be degrade performance, motivating us to include a \texttt{router} to select adapter relevant to the task at-hand based on the similarity of the goal and goal of a previous mistake \Cref{fig:pipeline-overview}.
% \vspace{-0.3cm}
\subsection{The \method Loop}
\label{self-direction-loop}
% \vspace{-0.2cm}
Motivated by these diagnostic findings, we introduce \textbf{\method}: a mechanism that enables agents to learn from a failed execution by reflecting on their mistakes and distilling the resulting lessons into a LoRA adapter, which can later be loaded for similar tasks (see Fig.~\ref{fig:pipeline-overview}). This enables learning from mistakes to scale while preserving the agent’s existing capabilities (see Section~\ref{sec:backward-transfer}).
% (moved Fig.~\ref{fig:pipeline-overview} earlier to appear on page 5)
% Given the effectiveness of retrying, we add a final retry with a prompt noting the agent previously failed to further boost the Self-Direction loop. 
Self-Direction operates in two main modes:

% \vspace{-0.3cm}
\paragraph{\textit{Training repair adapters.}}  When an agent encounters a failure, it performs self-directed reflection to synthesize a concise set ($N \le 24$) of procedural instruction-response examples. Crucially, the model acts as its own teacher without relying on external supervisor models and does not require full rollouts; the procedural examples are behavioral corrections (see Appendix~\ref{app:reflection-examples}). These examples are distilled into a lightweight LoRA adapter. To maintain numerical stability and keep the base weights unchanged, adapter matrix $A$ is initialized from the leading right-singular vectors of the targeted base weights and frozen, whereas matrix $B$ (initialized to zero) is optimized (see Appendix~\ref{app:svd_lora}). Training each adapter updates additional parameters equivalent to at most approximately $0.02\%$ of the base model's parameter count, enabling continuous accumulation in a persistent library of lessons $\mathcal{L}$ with a modest memory footprint ($\sim 85\,\text{MB}$ for Qwen3.6-27B). Adapter optimization takes $\sim1.5$ minutes on an H200 GPU in our setup; reflection and example synthesis require additional time. Additional details about the reflection, synthesis, and training stages are presented in Appendix~\ref{app:adapter-compute}.

% \vspace{-0.2cm}
\paragraph{\textit{Reusing adapters.}} When the same or a similar failed task is encountered, Self-Direction uses a sentence-embedding router\protect\footnotemark ~\citep{reimers2019sentence} to compare the incoming task with the task descriptions stored alongside each lesson in $\mathcal{L}$ (the similarity is 1 if a new adapter is trained for the current task). The corresponding adapter is then loaded into the frozen base model for re-execution.

% (moved Fig.~\ref{fig:pipeline-overview} earlier to appear on page 5) sentence-transformers/all-MiniLM-L6-v2
\footnotetext{We use the sentence-transformers/all-MiniLM-L6-v2 embedding model (\url{https://huggingface.co/sentence-transformers/all-MiniLM-L6-v2}).}

\textit{Base retry fallback: } Motivated by the benefits of resampling, if an adapter-guided retry fails, Self-Direction unloads the adapter and executes a final attempt on the base model with a prompt that indicates that a prior attempt failed (Appendix~\ref{app:prompts}). This fallback uses the remaining rollout budget for immediate recovery after the adapter is unloaded.

% \vspace{-0.3cm}
\section{Results}
\label{sec:results}
% \vspace{-0.2cm}
Beyond GAIA, we expand our evaluation to coding, tool-calling assistant tasks, and browser agent tasks. We find \method can repair mistakes across diverse settings, avoid repeating mistakes in the future, and scale learning while preserving existing capabilities, with repair adapters that generalize to unseen task categories and difficulties.

\footnotetext{Frontier model numbers are quoted verbatim from original benchmark numbers at \url{https://claw-eval.github.io/}.}

% \vspace{-0.2cm}
\subsection{Self-Direction Repairs Mistakes Across Various Task Types}
\label{sec:results-breadth}
\begin{figure*}[t]
    \centering
    \includegraphics[width=\textwidth]{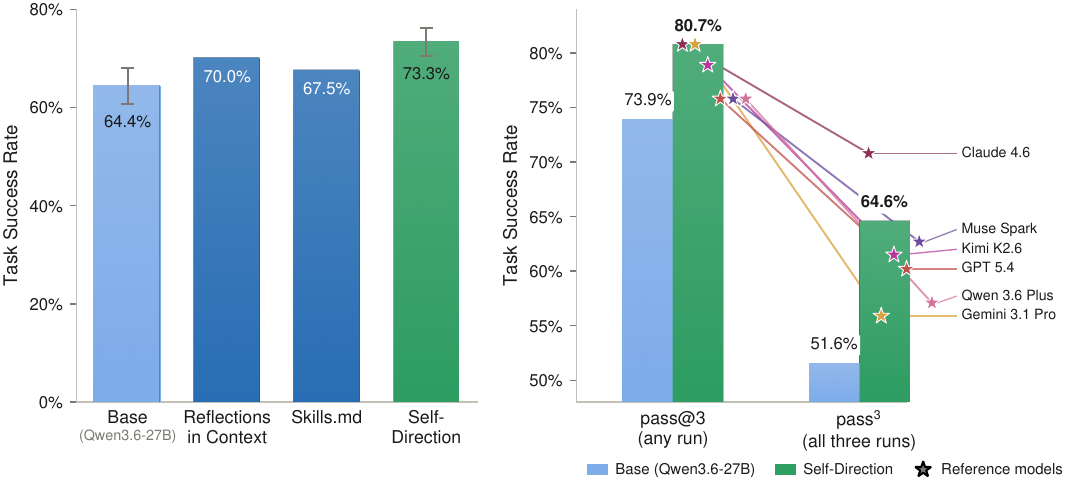}
    \caption{\textbf{Self-Direction improves task success across distinct agentic environments.} \textbf{Left:} Across three runs, Self-Direction improves mean success on Claw-Eval from $64.4\%$ to $73.3\%$ and outperforms other baselines. Error bars denote one sample standard deviation; cross-run pass@$3$ and pass$^3$ results are reported in Appendix~\ref{app:three-run-results}. Terminal-Bench 2.1 and GAIA results are reported in Figs.~\ref{fig:headline} and \ref{fig:gaia-headline}, respectively. \textbf{Right:} Self-Direction enables a 27B model to outperform several larger frontier models across pass@$3$ and pass$^3$ metrics for Claw-Eval tasks.\protect\footnotemark}
    \label{fig:cross-domain-headline}
    % \vspace{-0.4cm}
\end{figure*}

First, we measure whether Self-Direction improves task success across diverse agentic environments. We compare it to the base agent over three independent runs across Claw-Eval, GAIA, Terminal-Bench~2.1, as well as browser gym tasks from MiniWoB++. We find Self-Direction improves task success across all task types. On Claw-Eval, mean success increases from $64.4\% \pm 3.6\%$ to $73.3\% \pm 2.8\%$, a gain of $8.9$ percentage points (Fig.~\ref{fig:cross-domain-headline}). On Terminal-Bench~2.1, mean success increases from $51.7\% \pm 4.1\%$ to $65.5\% \pm 0.6\%$, a gain of $13.8$ points (Fig.~\ref{fig:headline}). On GAIA, the gain is $17.0$ points (Fig.~\ref{fig:gaia-headline}). We observe additional gains compared to the base agent on MiniWoB++ browser tasks (Appendix~\ref{app:miniwob-results}).

% \vspace{-0.3cm}
\paragraph{\method outperforms repair baselines and is agent-rollout efficient.}
We also compare Self-Direction with other strategies, such as adding reflection in context or distilling reflections in a \texttt{Skills.md} file. As shown in Fig.~\ref{fig:cross-domain-headline}, Self-Direction outperforms both settings. In addition, we find that Self-Direction compares favorably when controlling compute in terms of the number of rollouts (see Fig.~\ref{fig:rollout-efficiency}); e.g., Self-Direction achieves 55.8\% task success on GAIA with only 2.10 agent rollouts per task, matching the base agent's pass@6. 

% \vspace{-0.3cm}
\paragraph{\method narrows the gap to reported frontier agents.}
For Claw-Eval and Terminal-Bench~2.1, we also compare Self-Direction with officially reported numbers for frontier agents. We find Self-Direction with a 27B-parameter Qwen model achieves a competitive task success rate. For example, on Terminal-Bench~2.1, Self-Direction outperforms GLM-5.1, a model with $30\times$ more parameters, and is competitive with closed-weight models, such as Opus 4.7 and Gemini 3.1 Pro (see Fig.~\ref{fig:headline}). Similarly, on Claw-Eval, Self-Direction outperforms GLM-5.1 and is competitive with Gemini 3.1 Pro and Claude Opus 4.7 (see Fig.~\ref{fig:cross-domain-headline}). 
%These experiments establish that Self-Direction repairs mistakes across diverse environments. 

The gains extend across GAIA difficulty levels (Appendix~\ref{app:gaia-by-level}), Claw-Eval task categories and execution structures (Table~\ref{tab:claw-eval-stratified}), and Terminal-Bench~2.1 difficulty levels (Table~\ref{tab:terminal-bench-2-1-stratified}).  The gains also extend to agents with Qwen3.5 and Qwen3.8 model variants (\Cref{tab:terminal-bench-2-1-qwen3-5-9b} and Appendix~\ref{app:cross-model-results}). Together, these results demonstrate that Self-Direction improves diverse agents across open-ended research, tool-using workflows, terminal coding, and browser interaction. 
% \MP{Too much cross-reference to the point that the arguments are buried.}

Next, we study whether a repair persists across repeated encounters of a task. 
%\Mark{We can mention these comparisons include the effect of the harness too and reference the appendix?}

% \vspace{-0.3cm}
\subsection{Learned Lessons Prevent Repeated Mistakes}
\label{sec:results-reliability}

% once a mistake is corrected, how likely is the agent to repeat mistakes?
% 
% \vspace{-0.1cm}
After repairing a mistake, we study how consistently Self-Direction prevents repetitions of the same mistakes in the future. We evaluate whether a repair is merely the result of a lucky retry or a persistent lesson in two ways. First, we measure whether a repair remains effective over future repeated encounters of a task. Second, we assess variability in how often a learned lesson repairs a mistake across multiple independent runs. 

% \vspace{-0.3cm}
\paragraph{\method avoids repeating mistakes across future encounters of a task.} Across 185 prior mistakes from the three benchmarks, \method nearly doubles the fraction solved consistently across three encounters: pass$^{3}$ increases from 9.7\% to 18.9\% ($+9.2$ pp) across GAIA, Terminal-Bench~2.1, and Claw-Eval (see \Cref{tab:reliability}). At each encounter, both base retrying and \method receive up to two full rollouts. We report \emph{mistakes avoided}, $\mathrm{pass}^{k}$, as the fraction of prior mistakes solved in every encounter through encounter $k$. We also report \emph{corrections preserved}, $\mathrm{pass}^{k}/\mathrm{pass}^{1}$, as the fraction of first-encounter successes that remain successful through encounter $k$. More importantly, after \method repairs a mistake once, $76.9\%$ of the repairs remain effective through a second encounter and $67.3\%$ through a third, compared with only $52.4\%$ and $42.9\%$ for the base agent. 

\begin{figure*}[t]
      \centering
      \includegraphics[width=0.98\textwidth]{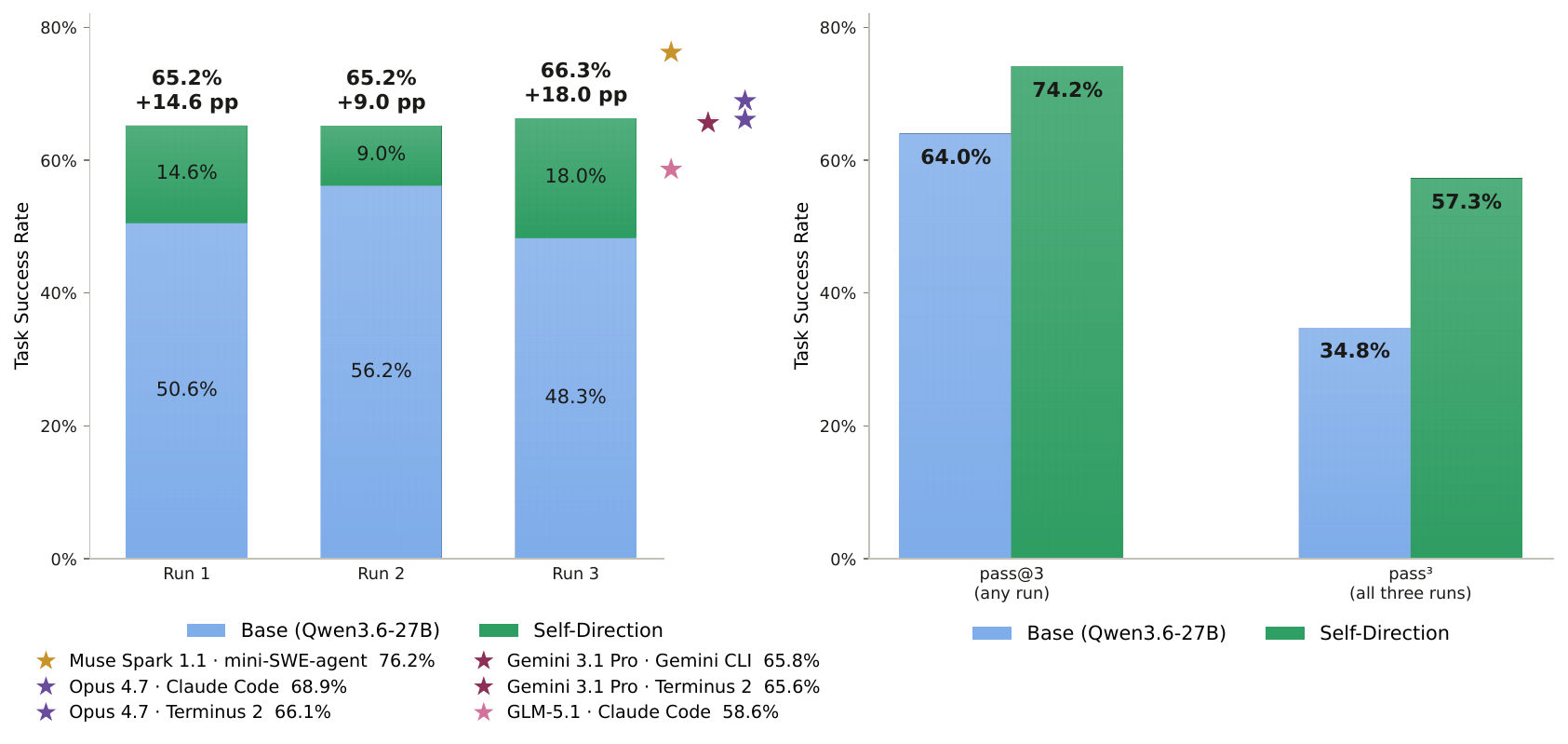}
      \caption{\textbf{Self-Direction reduces the performance variability of the base agent.}
      \textbf{Left:} Base agent performance varies within the same benchmark in the exact same setting, ranging from 48.3\% to 56.2\%. Self-Direction helps maintain reliable performance across all runs. \textbf{Right:} Self-Direction increases pass@$3$ from $64.0\%$ to $74.2\%$ and pass$^3$ from $34.8\%$ to $57.3\%$. Reference frontier scores are provided for context to highlight differences between different harnesses and agentic settings.}
      \label{fig:terminal-three-runs}
\end{figure*}

In Fig.~\ref{fig:terminal-three-runs}, we illustrate the variable performance of the base agent on Terminal-Bench~2.1 across three independent runs, which range from 48.3\% to 56.2\%; Self-Direction improves every run with gains ranging from +9.0 to +18.0 pp. Notably, pass$^3$, which requires a task to succeed in all three runs, increases from 34.8\% to 57.3\% (+22.5 pp). We show a similar plot for eight runs in \Cref{app:terminal-eight-runs}. 

Therefore, Self-Direction spends compute on repairing the current mistake, retaining the resulting lesson, and helping the agent avoid that mistake across future task encounters.

Next, we ask whether a growing library of lessons preserves the agent’s existing capabilities.

\begin{table*}
  \centering
  \caption{\textbf{Self-Direction prevents repaired mistakes from recurring.}
  Learned lessons increase success across repeated encounters and preserve
  repairs more reliably than base-model retrying. Results cover 101 GAIA,
  44 Terminal-Bench~2.1, and 40 Claw-Eval prior mistakes.}
  \label{tab:reliability}

  \small
  \renewcommand{\arraystretch}{1.24}
  \setlength{\tabcolsep}{5pt}

  \begin{tabularx}{\textwidth}{@{}
    >{\raggedright\arraybackslash}X
    c
    r r r
    r r@{}}
    \toprule
    \rowcolor{TableHeader}
    \textbf{Benchmark} &
    \textbf{Encounters} &
    \multicolumn{3}{c}{\textbf{Mistakes avoided} ($\mathrm{pass}^{k}$)} &
    \multicolumn{2}{c}{\textbf{Corrections preserved}} \\
    \cmidrule(lr){3-5}\cmidrule(l){6-7}
    \rowcolor{TableHeader}
      & $k$
      & \textbf{Base retry}
      & \textbf{Self-Direction}
      & \textbf{Gain}
      & \textbf{Base retry}
      & \textbf{Self-Direction} \\
    \midrule

    \textbf{GAIA}
      & 2 & $9.9\%$  & \textbf{$21.8\%$} & \textcolor{GainColor}{\textbf{+11.9\,pp}} & $45.5\%$ & \textbf{$78.6\%$} \\
    & 3 & $6.9\%$  & \textbf{$19.8\%$} & \textcolor{GainColor}{\textbf{+12.9\,pp}} & $31.8\%$ & \textbf{$71.4\%$} \\

    \rowcolor{TableStripe}
    \textbf{\mbox{Terminal-Bench~2.1}}
      & 2 & $13.6\%$ & \textbf{$20.5\%$} & \textcolor{GainColor}{\textbf{+6.8\,pp}} & $50.0\%$ & \textbf{$69.2\%$} \\
    \rowcolor{TableStripe}
      & 3 & $13.6\%$ & \textbf{$15.9\%$} & \textcolor{GainColor}{\textbf{+2.3\,pp}} & $50.0\%$ & \textbf{$53.8\%$} \\

    \textbf{Claw-Eval}
      & 2 & $15.0\%$ & \textbf{$22.5\%$} & \textcolor{GainColor}{\textbf{+7.5\,pp}} & $75.0\%$ & \textbf{$81.8\%$} \\
      & 3 & $12.5\%$ & \textbf{$20.0\%$} & \textcolor{GainColor}{\textbf{+7.5\,pp}} & $62.5\%$ & \textbf{$72.7\%$} \\

    \midrule
    \rowcolor{TableHeader}
    \mbox{\textbf{Total}}
      & 2 & \textbf{$11.9\%$} & \textbf{$21.6\%$} & \textcolor{GainColor}{\textbf{+9.7\,pp}} & \textbf{$52.4\%$} & \textbf{$76.9\%$} \\
    \rowcolor{TableHeader}
      & 3 & \textbf{$9.7\%$}  & \textbf{$18.9\%$} & \textcolor{GainColor}{\textbf{+9.2\,pp}} & \textbf{$42.9\%$} & \textbf{$67.3\%$} \\
    \bottomrule
  \end{tabularx}

  % \vspace{2pt}
  % \begin{minipage}{\textwidth}
  %   \footnotesize\color{black!65}
  %   \textit{Note.} Mistakes avoided is the fraction of prior mistakes solved in
  %   every encounter through $k$ ($\mathrm{pass}^{k}$). Corrections preserved is
  %   the fraction of first-encounter successes that remain successful through
  %   encounter $k$ ($\mathrm{pass}^{k}/\mathrm{pass}^{1}$).
  %   Both methods receive up to two attempts per encounter.
  % \end{minipage}
  % \vspace{-0.3cm}
\end{table*}

% \vspace{-0.4cm}
\subsection{Self-Direction Learns New Lessons Without Impacting Existing Capabilities}
\label{sec:backward-transfer}

% \MP{This section is not empirical. I would say merge it.}
Continual lesson acquisition is useful only if learning a new lesson does not disrupt the agent's existing capabilities. \method limits this interference in two ways. First, it routes every task to the unchanged base model for the initial attempt; adapter retrieval occurs only after a base agent makes a mistake. Second, each learned lesson is stored as a separate lightweight adapter, leaving the base model and previously learned adapters frozen. Learning a new lesson, therefore, cannot overwrite previously acquired behavior.

On a retry, \method loads an existing adapter only when the current task goal is sufficiently similar to the adapter's source-task goal under threshold $\tau$. An agent with $\tau=1$ will load an existing adapter only if the same task is encountered again; otherwise, the agent will train a new adapter. In this setting, \method trains new adapters for each new task and preserves stored base-model and adapter parameters by keeping them frozen. When $\tau < 1$, \method loads adapters for similar but not necessarily identical tasks, enabling reuse of adapters across new tasks. We find that empirically, except for low thresholds $\tau < 0.30$, the router preserves or improves existing capabilities even with a library of lessons containing adapters from the two other unrelated benchmarks. The detailed findings are presented in Appendix~\ref{sec:app-preserving}.

% We next show empirically that across a range of appropriate thresholds, no unrelated skills from other benchmarks are ever loaded, allowing \method to preserve existing capabilities even as the library of skills grows. 

% On a retry, \method loads an existing adapter only when its task-goal similarity exceeds a threshold $\tau$. Setting $\tau=1$ disables reuse and recovers unbiased, per-failure \method: a newly trained adapter is loaded directly for the current retry, while the frozen base agent remains unchanged. Lower thresholds enable reuse; if no adapter qualifies, a new one is trained. We next show empirically that an appropriate threshold prevents cross-benchmark retrievals, avoiding activation of unrelated skills and preserving existing capabilities as the library grows.

% For each pair of GAIA, Claw-Eval, and Terminal-Bench~2.1, we score tasks from one benchmark against adapters learned on the other (\Cref{tab:cross_benchmark_thresholds}). The largest observed cross-domain similarity is $0.546$; consequently, a single global threshold of $\tau=0.55$ produces zero unintended retrievals across all 370 cross-domain evaluations. Thus, no evaluated task activates an adapter learned from an unrelated benchmark. Together, the initial base agent attempt, separate lesson storage, and threshold routing preserve existing capabilities while enabling the lesson library to grow to repair mistakes.

Next, we study a simple strategy of selecting the most similar adapter, without a fixed threshold. We find that selecting the most similar adapter leads to generalization across new tasks.

\subsection{Learned Lessons Transfer to Unseen Tasks Across Difficulties and Types} 
\label{sec:results-transfer}

Training a new lesson for each failure may not always be practical. Here, we ask whether lessons acquired through \method transfer to unseen tasks. We explore three types of transfer: to unseen tasks in a continual deployment stream, to unseen harder tasks, and to unseen task categories.
% \vspace{-0.2cm}
\paragraph{A library of three lessons boosts task success across GAIA.}
We simulate a deployment stream by randomly interleaving all 165 GAIA tasks across all levels. The agent begins with an empty library of lessons and applies \method to learn new lessons. With the library capped at three adapters, the complete \method loop increases task success across the full 165-task GAIA stream from 36.4\% to 52.7\%, a gain of 16.3 points, as shown in \Cref{tab:skill-transfer-summary}. The complete three-adapter library occupies only 256 MB, showing that useful experience can be retained with a modest incremental memory footprint. We report additional deployment-stream experiments with libraries containing six or twelve adapters in Appendix~\ref{app:gaia-adapters-visual}.

% \vspace{-0.3cm}
\paragraph{Lessons learned on easier tasks transfer to harder tasks.} 
We study how well lessons learned on easier Level-1 GAIA tasks transfer to unseen, harder Level-2 tasks. Specifically, we train 28 adapters on failed Level-1 tasks, freeze the resulting library, and evaluate it on 86 unseen Level-2 tasks. We find that reusing Level-1 lessons recovers nine failures, increasing overall accuracy from $40.7\%$ to $51.2\%$ – a gain of $10.5$ percentage points. The base agent achieves a pass@2 rate of \(43\%\) on this subset, demonstrating that the learned repairs remain effective even on more challenging tasks. 

% \vspace{-0.2cm}
\paragraph{Lessons transfer across related task categories.}
Finally, we test whether lessons cross category boundaries within Claw-Eval. We freeze a library of three adapters learned exclusively from operations tasks and evaluate it on 24 unseen workflow tasks (\Cref{tab:skill-transfer-summary}). The base agent solves 18 tasks on the initial execution. On the six remaining failures, the operations library solves three, recovering half of the failures without training on a single workflow task. This raises accuracy from $75.0\%$ to $87.5\%$, a gain of $12.5$ percentage points. Qualitative examples of these transfers are presented in Appendix~\ref{app:qualitative-transfer}. % and a mathematical analysis of why the adapters transfer meta capabilities is presented in Appendix~\ref{app:repair-cone}.

Collectively, our findings present \method as a promising foundation for agents that improve through continual use. Across research, tool-use, browser, and coding environments,  \method repairs mistakes now, prevents repeating mistakes in the future, preserves existing capabilities, and generalizes to harder tasks and tasks in related categories. \method offers a path towards agents whose deployment experience compounds over time---where each mistake can contribute not only to solving the task at hand, but to building lasting capabilities.

\section{Conclusion}
A growing library of lessons holds promise for scaling continuous agent self-improvement. Encoding each lesson as a LoRA adapter provides a practical way to store and reuse experience across deployments and domains. Besides self-improvement, \method may support model specialization for tailored applications, personalization, and patching security vulnerabilities. Lessons could also be implemented through self-evolving harnesses or stored in \texttt{Skills.md} files for use in context. Future research can explore how to select the most suitable medium for retaining a given lesson and how to compose multiple lessons when addressing a mistake.

\paragraph{Limitations.} \method begins with a mistake, which must be detected through self-evaluation or grader feedback, or reported by a user. It then retains the resulting lesson. Encoding that lesson in model weights requires GPU resources and storage capacity beyond standard inference to train LoRA adapters. Textual reflections can instead be stored and accumulated until sufficient resources become available. Compared with in-context or textual lessons, which transfer across model families and remain human-interpretable, in-weight repairs avoid consuming the limited context window but remain model-specific.
\clearpage

\bibliography{iclr2027_conference,references_fixed,paper}
\bibliographystyle{assets/plainnat}

\newpage
\beginappendix

\section{Expanded Related Work}
\label{sec:appendix_related_work}
\paragraph{Lessons kept in context.}
Expanding on the context-based methods discussed in Section~\ref{sec:related-work}, systems frequently rely on iterative self-correction without training. For instance, Self-Refine allows a single model to critique and revise its own output \citep{madaanSelfRefineIterativeRefinement2023}, whereas GEPA turns this reflective step into an optimizer by attributing outcomes to specific prompt content and evolving candidate pools \citep{agrawal2026gepa}. Maintaining textual skills has also appeared in deployed systems as markdown skill files \citep{Claude_skill_2026}, with recent work co-evolving such skills alongside persistent knowledge bases for cross-model transfer \citep{tangWikiSkillCompilingAgent2026}. Other approaches, like Skill-Pro, refine a skill memory using a PPO-style objective while keeping the underlying model frozen \citep{miSkillProLearningReusable2026}. However, relying on externalized skills has significant operational downsides. Retrieved skills have been shown to introduce functional failures and efficiency regressions even when they appear relevant \citep{dong2026agent}. Furthermore, the limitation is fundamentally expressive: \citet{gupta2026reasoncache} argue that concatenating exemplars is only a shallow form of adaptation, as the context can only reach the internal states that a literal prompt induces through frozen weights.

\paragraph{Lessons kept in the harness.}
Expanding on the harness-based methods discussed in Section~\ref{sec:related-work}, there is extensive debate regarding how to best represent an agent's harness \citep{pan2026nlah, chen2026harnessx, wang2026handbook} and what elements the harness should own at runtime \citep{jiang2026harness1}. While early work focused on meta-agents programming new agents in code \citep{hu2025adas}, recent systems have heavily specialized this self-evolving loop. For instance, \citet{lou2026autoharness} synthesize code harnesses automatically, \citet{sengupta2026harbor} treat harness configuration as a direct optimization target, and \citet{lee2026metaharness} optimize a weaker agent's harness using a stronger external one. \citet{lin2026ahe} pair every harness edit with a prediction that is later verified against task outcomes. Beyond single-episode optimization, \citet{nie2026evolvenet} distribute this process across deployments and \citet{nie2026tthe} evolve harness code over an unlabeled test stream, though adaptation is scheduled per batch rather than conditioned on a verified failure. Other approaches synthesize harnesses just in time \citep{zhang2026jitagent}, adapt the model-environment interface offline \citep{xu2026adapt}, compose harnesses into multi-day development loops \citep{yan2026hoh}, or expose this evolving surface directly in deployed systems \citep{primeintellect2026primeagent}. Ultimately, all of these methods accumulate a lineage of scaffold code rather than a library of parametric repairs.

\paragraph{Lessons kept in weights.}
As discussed in Section~\ref{sec:related-work}, low-rank adaptation reduces the risk of catastrophic forgetting. However, empirical analysis shows that routing among several adapters can still underperform an oracle \citep{backUnderstandingLoRAKnowledg2026}. Variations like PiSSA initialize an adapter from the principal singular components of a pretrained weight and train those components, refactoring the stored weight into a residual \citep{mengPiSSAPrincipalSingular2025}. In contrast, we freeze the spectral factor and train only the zero-initialized coefficient matrix, leaving base weights completely untouched such that the adapter remains detachable. This follows the initialization introduced for online parametric memory by \citet{ren2026scaling}, though their adaptation is triggered by context-budget overflow and evolves a single intra-episode state rather than a retrievable library.

Furthermore, the training signal for in-weight learning is typically gathered in offline batches, as seen in STaR and ReST \citep{zelikmanSTaRBootstrappingReasoning2022, gulcehre2023reinforced}. Methods that learn specifically from mistakes usually rely on a stronger corrector, such as GPT-4 \citep{LearningMistakesMakes2024}, or the model judging itself \citep{yuanSelfRewardingLanguageModels2025}. When a teacher is given privileged information, the resulting performance gap conflates true transferable capability with information asymmetry \citep{yuDOPDDualOnpolicy2026}. Closest to our setting, FailForge converts persistent failures into a training signal, but uses a stronger teacher, a reference patch as privileged input, and a single global fine-tune after which the textual skill is deliberately discarded \citep{lvFailForgeDistillingProcedural2026}. Finally, Test-Time Training (TTT) is related to our in-weight updates during goal execution. TTT adapts parameters on the current test input under a self-supervised objective \citep{sun2020testtime}, on text retrieved from neighbors before being reset \citep{hardt2024testtime}, or through a per-task adapter trained on rule-based augmentations \citep{akyurek2025surprising}. In each of these cases, adaptation is transient and unconditional, whereas our method conditions strictly on verified failures and preserves the repair for future reuse.

\clearpage

\section{Additional Browser and Computer-Use Benchmarks}
\label{app:additional-browser-benchmarks}

We further evaluate whether \method extends to environments in which success depends on interacting with graphical interfaces rather than producing a textual answer. We consider MiniWoB++, which contains focused browser-control tasks. We retain the same fixed base model, reflection procedure, adapter training configuration, and adapter-first retry with base retry fallback used in the main experiments.

The benchmark provides the agent with a textual accessibility tree containing stable element identifiers. The agent receives no screenshots or visual pixels. The available tools support observing the page, executing BrowserGym actions, clicking, typing, selecting options, pressing keys, scrolling, waiting, and signaling completion.

\subsection{MiniWoB++}
\label{app:miniwob-results}

Many MiniWoB++ categories are already saturated by the base agent and provide little opportunity for learning. We, therefore, focus on four categories with meaningful remaining headroom. These include a harder multi-target social-media task, collapsible interface elements, and two variants of email-inbox interaction. Simpler categories, such as \texttt{navigate-tree} and \texttt{multi-layouts}, were excluded because the base model already solves nearly all evaluated instances.

Across the selected categories, the common initial base execution solves $42/100$ tasks. Three independent base attempts increase coverage to $55/100$, whereas one execution of the full \method loop achieves $54/100$. By nearly matching base pass@$3$, these results suggest that the same mechanism for learning from mistakes extends to browser interaction. The resulting adapters remain available for future reuse (see Table~\ref{tab:additional-browser-results}).

\begin{table*}[h]
      \centering
      \caption{\textbf{Additional results on browser use environments.} \method enables meaningful gains in tasks involving browser use.}
      \label{tab:additional-browser-results}
      \small
      \setlength{\tabcolsep}{7pt}
      \resizebox{\linewidth}{!}{%
      \begin{tabular}{@{}llrrrr@{}}
          \toprule
          Benchmark & Task group & Tasks & Initial base & Base pass@$3$ & Full loop \\
          & & & (Qwen3.6-27B) & (no lesson retention) & (lessons retained) \\
          \midrule
          MiniWoB++
              & \texttt{social-media-all}
              & 25 & $6/25$ ($24\%$) & $11/25$ ($44\%$) & $10/25$ ($40\%$) \\
          MiniWoB++
              & \texttt{click-collapsible-2}
              & 25 & $8/25$ ($32\%$) & $11/25$ ($44\%$) & $11/25$ ($44\%$) \\
          MiniWoB++
              & \texttt{email-inbox}
              & 25 & $15/25$ ($60\%$) & $18/25$ ($72\%$) & $18/25$ ($72\%$) \\
          MiniWoB++
              & \texttt{email-inbox-nl-turk}
              & 25 & $13/25$ ($52\%$) & $15/25$ ($60\%$) & $15/25$ ($60\%$) \\
          \cmidrule(lr){2-6}
          MiniWoB++
              & \textbf{Overall}
              & \textbf{100}
              & \textbf{$42/100$ ($42\%$)}
              & \textbf{$55/100$ ($55\%$)}
              & \textbf{$54/100$ ($54\%$)} \\
          % \midrule
          % OpenApps
          %     & Full suite
          %     & 270
          %     & $254/270$ ($94.1\%$)
          %     & --
          %     & $258/270$ ($95.6\%$) \\
          \bottomrule
      \end{tabular}
      }
\end{table*}

\clearpage

\section{Expanded Results Across Benchmarks}
\label{app:expanded-results}

The main text reports pooled and average results to emphasize the central empirical findings. Here, we present the stratification across GAIA difficulty levels and repeated executions of Claw-Eval and Terminal-Bench~2.1. These expanded results test whether the reported improvements are concentrated in particular task groups or isolated runs.

\subsection{\method Trains Lessons That Can Be Reused for Future Tasks}

We find that \method enables successful transfer of learned lessons to future related, unseen tasks as described in more detail in Section~\ref{sec:results-transfer}.

\begin{table*}[h]
  \centering
  \caption{\textbf{\method transfers learned lessons to unseen tasks.}
  Stored lessons generalize to new tasks in a continual deployment stream, harder tasks, and related task categories.}
  \label{tab:skill-transfer-summary}

  \small
  \renewcommand{\arraystretch}{1.34}
  \setlength{\tabcolsep}{5pt}

  \begin{tabularx}{\textwidth}{@{}
    >{\raggedright\arraybackslash}p{0.15\textwidth}
    >{\raggedright\arraybackslash}p{0.17\textwidth}
    >{\raggedright\arraybackslash}p{0.13\textwidth}
    >{\raggedright\arraybackslash}X
    >{\raggedleft\arraybackslash}p{0.18\textwidth}@{}}
    \toprule
    \rowcolor{TableHeader}
    \textbf{Generalization} &
    \textbf{Lessons learned from} &
    \textbf{Library} &
    \textbf{Evaluation tasks} &
    \textbf{Improvement} \\
    \midrule

    \rowcolor{white}
    \textbf{Tasks in a continual stream}\newline
    {\footnotesize\color{black!65}GAIA}
      & 3 GAIA tasks
      & 3 adapters %\newline {\footnotesize\color{black!65}(256\,MB)}
      & GAIA deployment stream ($n=165$)
      & \shortstack[r]{%
          {\color{GainColor}\textbf{+16.3\,pp}}\\[-1pt]
          {\scriptsize\color{black!65}$36.4\% \rightarrow 52.7\%$}} \\

    \rowcolor{TableStripe}
    \textbf{Harder tasks}\newline
    {\footnotesize\color{black!65}GAIA}
      & Easy Level~1 tasks
      & 28 adapters
      & Harder Level~2 tasks ($n=86$)
      & \shortstack[r]{%
          {\color{GainColor}\textbf{+10.47\,pp}}\\[-1pt]
          {\scriptsize\color{black!65}$40.7\% \rightarrow 51.2\%$}} \\

    \rowcolor{white}
    \textbf{Tasks across categories}\newline
    {\footnotesize\color{black!65}Claw-Eval}
      & Operations tasks
      & 3 adapters
      & Unseen workflow tasks ($n=24$)
      & \shortstack[r]{%
          {\color{GainColor}\textbf{+12.5\,pp}}\\[-1pt]
          {\scriptsize\color{black!65}$75.0\% \rightarrow 87.5\%$}} \\

    \bottomrule
  \end{tabularx}

  \vspace{2pt}
  % \begin{minipage}{\textwidth}
  %   \footnotesize\color{black!65}
  %   \textit{Note.} Bold values are absolute percentage-point gains;
  %   smaller values report accuracy before and after skill reuse.
  % \end{minipage}
\end{table*}

\subsection{GAIA Results by Difficulty Level}
\label{app:gaia-by-level}

\begin{figure*}[h]
    \centering
    \includegraphics[width=0.98\textwidth]{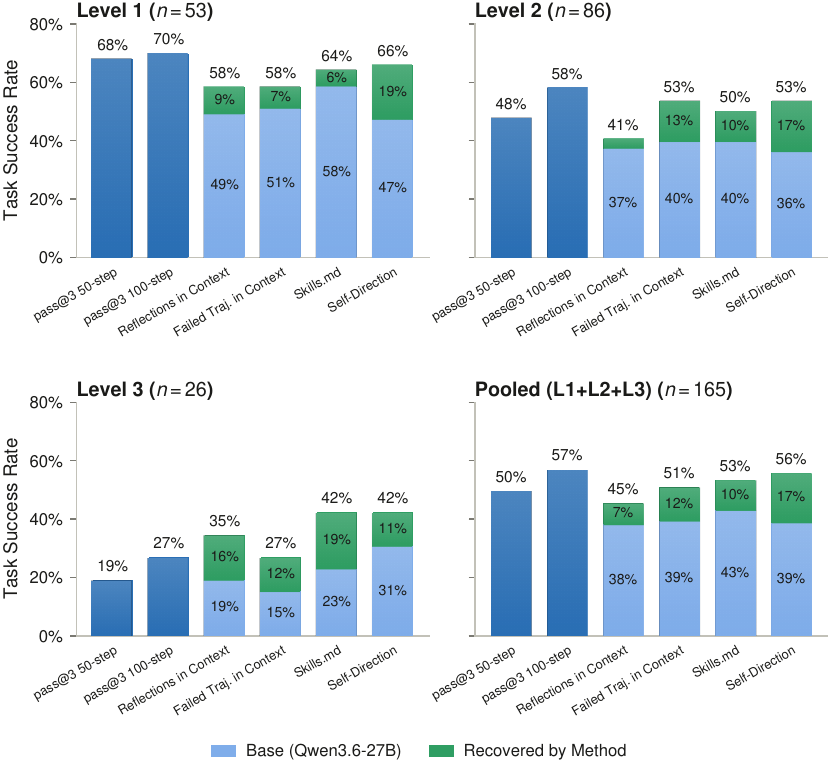}
    \caption{\textbf{Expanded GAIA recovery results by difficulty level.}
    Each panel compares \resampling, \trajretry, \icl, reflections in Skills.md, and SVD-LoRA. Stacked bars distinguish initial base successes from tasks recovered by each strategy. SVD-LoRA improves performance on all three levels and reaches $92/165$ tasks overall.}
      \label{fig:gaia-expanded}
  \end{figure*}

The pooled GAIA result presented in Fig.~\ref{fig:gaia-headline} summarizes how different recovery strategies respond to an initial failure. Fig.~\ref{fig:gaia-expanded} expands this comparison across GAIA Levels~1-3. We compare Base resampling, Base resampling with double the tool call budget, \trajretry, \icl, reflections accumulating in Skills.md, and SVD-initialized LoRA using the strategies defined in Section~\ref{sec:method}. All settings use up to 50 tool-call steps, except the 100-step budget ablation. The 100-step ablation demonstrates how sequential test-time scaling enables performance jumps. It should not be interpreted as a fair rollouts-matched comparison to other strategies and is reported for reference.

SVD-initialized LoRA improves performance at every difficulty level. Starting from the initial base execution, the full loop increases accuracy from $25/53$ to $35/53$ on Level~1, from $31/86$ to $46/86$ on Level~2, and from $8/26$ to $11/26$ on Level~3. Pooled across all levels, modular lesson adapters recover 21 failures and the base retry fallback recovers seven more, producing the headline result of $92/165$ ($55.8\%$).

\subsection{Expanded Per-Run Results on Claw-Eval and Terminal-Bench}
\label{app:three-run-results}

The aggregate results presented in Figs.~\ref{fig:headline} and \ref{fig:cross-domain-headline} report mean success across three runs. Here, we report all three complete executions of Claw-Eval and all eight executions of Terminal-Bench~2.1, together with task-level pass@$k$ and pass$^k$. Pass@$k$ measures coverage by counting tasks solved in at least one run, whereas pass$^k$ measures consistency by requiring success in all $k$ runs.

\begin{figure*}[h]
      \centering
      \includegraphics[width=0.98\textwidth]{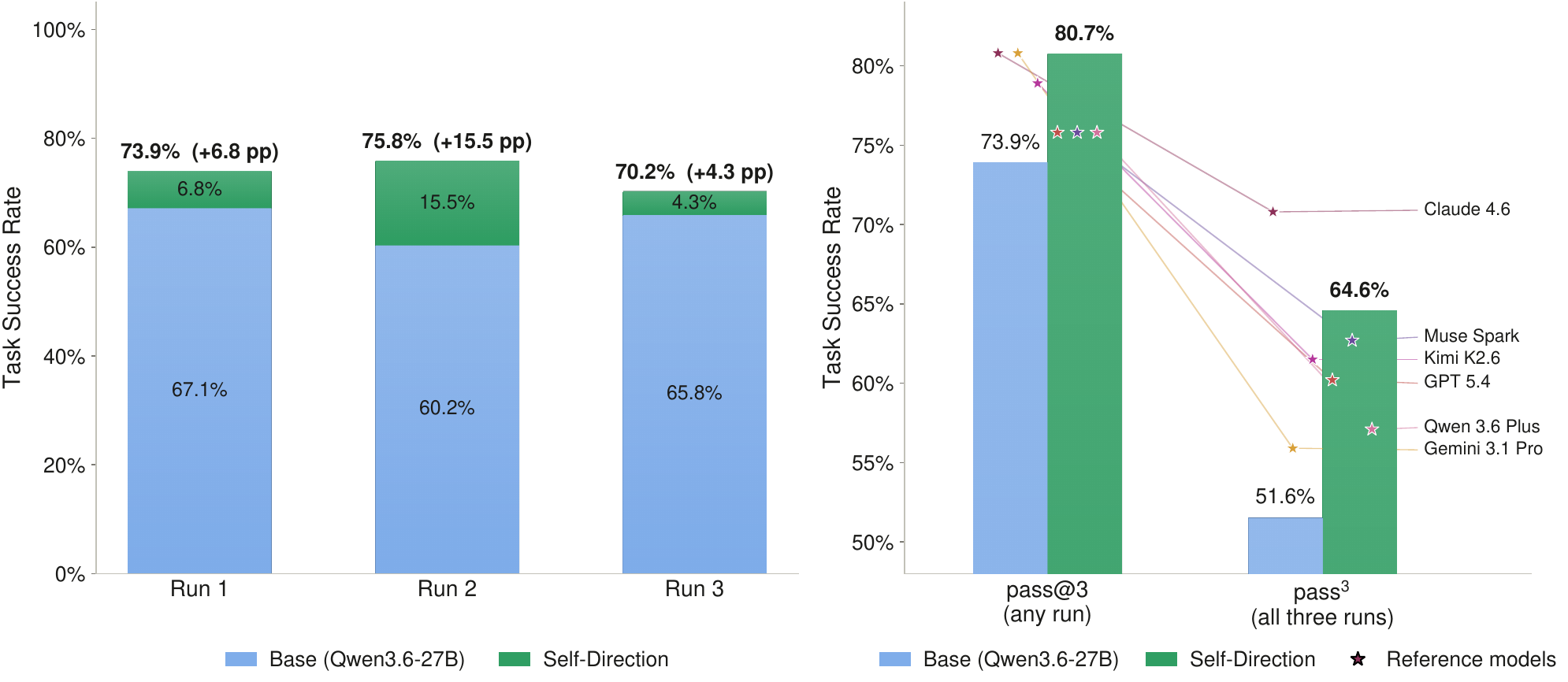}
      \caption{\textbf{Expanded three-run results on Claw-Eval.}
      The left panel decomposes each run into initial base successes, additional
      full-loop recoveries, and unresolved tasks. The right panel reports
      cross-run coverage (pass@$3$) and consistency (pass$^3$). The full loop
      improves both metrics and produces a positive accuracy gain in every run.
      Reference-model scores are included for context and may use different agent
      configurations.}
      \label{fig:claw-three-runs}
  \end{figure*}

The frontier-model reference points shown in Figs.~\ref{fig:terminal-three-runs} and \ref{fig:claw-three-runs} are reproduced from the corresponding public benchmark harnesses and should be interpreted as system-level results rather than controlled comparisons of the underlying models alone. Agent performance can vary substantially with the surrounding scaffold, including its prompts, tool interfaces, context management, stopping criteria, and retry policy. In contrast, all of our experiments use the same deliberately minimal ReAct harness described in Section~\ref{sec:method}, enabling the effect of the \method loop to be isolated within our setting. These comparisons, therefore, provide useful performance context but do not attribute differences solely to model capability. More broadly, sophisticated continual harnesses and \method are complementary: Harnesses improve how a fixed model is orchestrated, whereas \method gives an open model a mechanism to convert failures into persistent, modular in-weight lessons. The same learning loop could consequently be incorporated into richer agent harnesses.

\paragraph{Claw-Eval.}
The full loop improves accuracy in every Claw-Eval run, with gains ranging from $+4.3$ to $+15.5$ percentage points (see Fig.~\ref{fig:claw-three-runs}). Across the 161 tasks, pass@$3$ increases from $73.9\%$ to $80.7\%$, whereas pass$^3$ increases from $51.6\%$ to $64.6\%$. \method, therefore, improves both the range of tasks the agent can solve and the set it solves consistently. We also report appropriate frontier model comparisons for context; the results are reported as is from the official harness used by the original benchmark authors \citep{yeClawEvalTrustworthyEvaluation2026}. 

% \paragraph{Terminal-Bench~2.1.}
% The same pattern holds in interactive terminal environments. Across three complete 89-task runs, the full loop improves accuracy by $9.0$--$18.0$ percentage points (See Figure~\ref{fig:terminal-three-runs-app}). Pass@$3$ rises from $64.0\%$ to $74.2\%$, showing broader task coverage, while pass$^3$ rises from $34.8\%$ to $57.3\%$. The larger improvement in pass$^3$ demonstrates that \method primarily expands the set of tasks solved reliably across executions.

%   \begin{figure*}[t]
%       \centering
%       \includegraphics[width=0.98\textwidth]{new_figures/iclr_terminal_bench_2_1_three_runs.pdf}
%       \caption{\textbf{Expanded three-run results on Terminal-Bench~2.1.}
%       The left panel shows the outcome decomposition for each complete run. The
%       right panel reports task coverage and consistency across runs. \method
%       increases pass@$3$ from $64.0\%$ to $74.2\%$ and pass$^3$ from $34.8\%$ to
%       $57.3\%$. Reference scores are provided only for context because agent and
%       execution configurations may differ.}
%       \label{fig:terminal-three-runs-app}
%   \end{figure*}

\subsection{Terminal-Bench Reliability Across Eight Runs}
\label{app:terminal-eight-runs}

\begin{figure*}[t]
      \centering
      \includegraphics[width=0.98\textwidth]{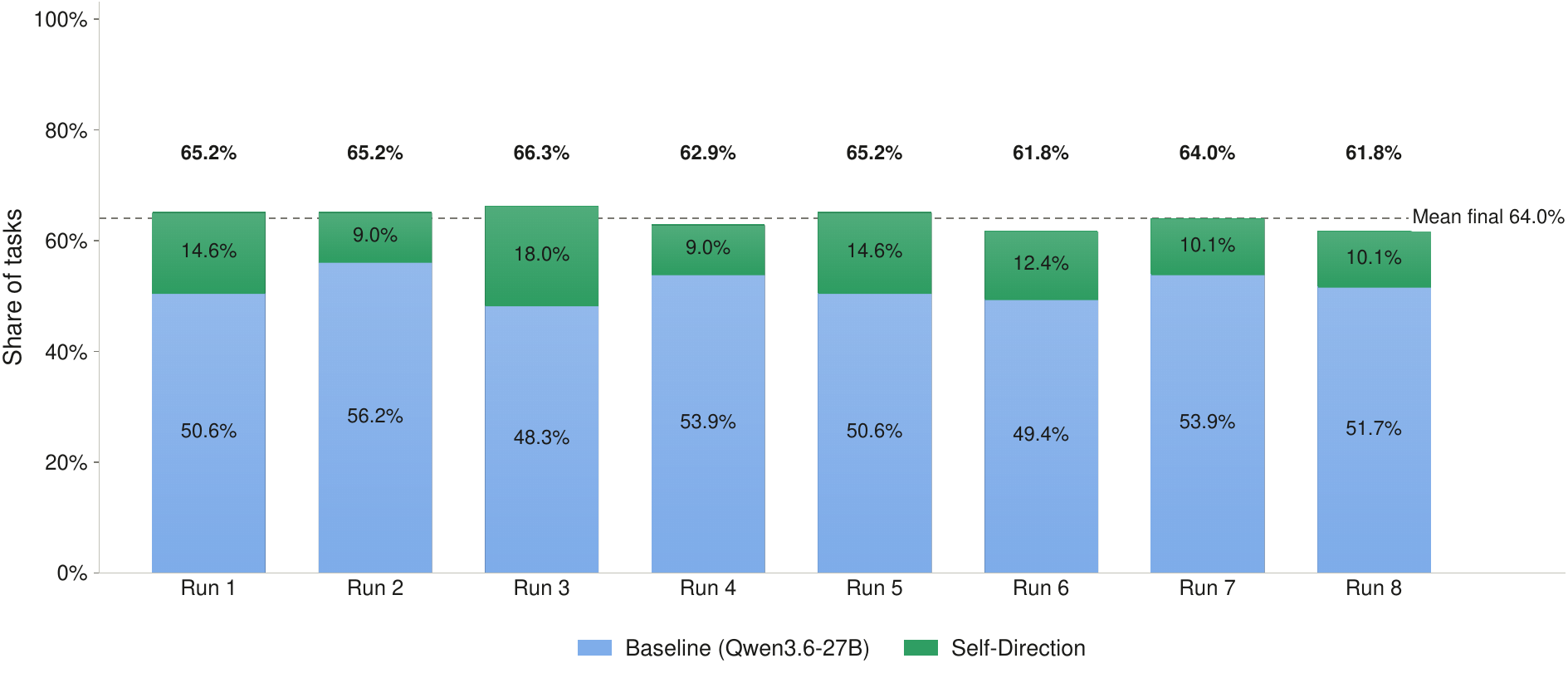}
      \caption{\textbf{Terminal-Bench~2.1 outcomes across eight complete runs.}
      Each bar decomposes success from the initial base execution and additional
      recovery by the full loop. \method improves all eight runs and raises
      mean accuracy from $51.8\%$ to $64.0\%$.}
      \label{fig:terminal-eight-runs}
  \end{figure*}

\begin{figure*}[htbp]
      \centering
      \begin{subfigure}[t]{0.48\textwidth}
          \centering
          \includegraphics[width=\linewidth]{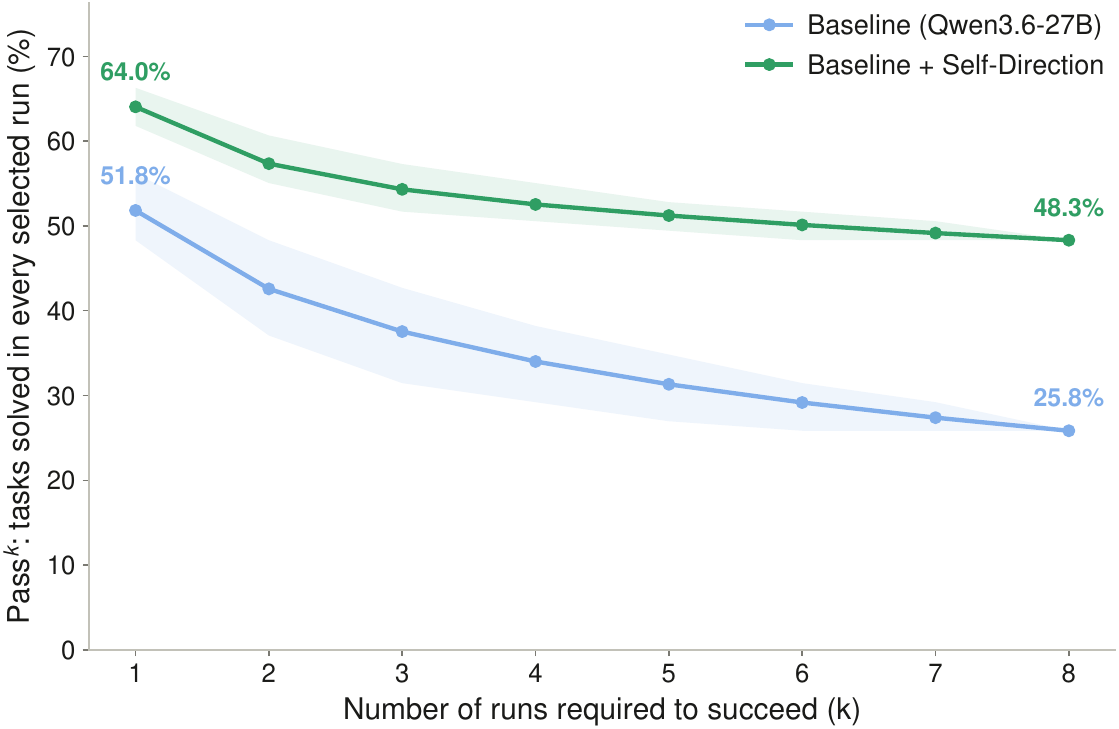}
          \caption{Pass$^k$ across repeated executions.}
          \label{fig:tb-pass-power-k}
      \end{subfigure}
      \hfill
      \begin{subfigure}[t]{0.48\textwidth}
          \centering
          \includegraphics[width=\linewidth]{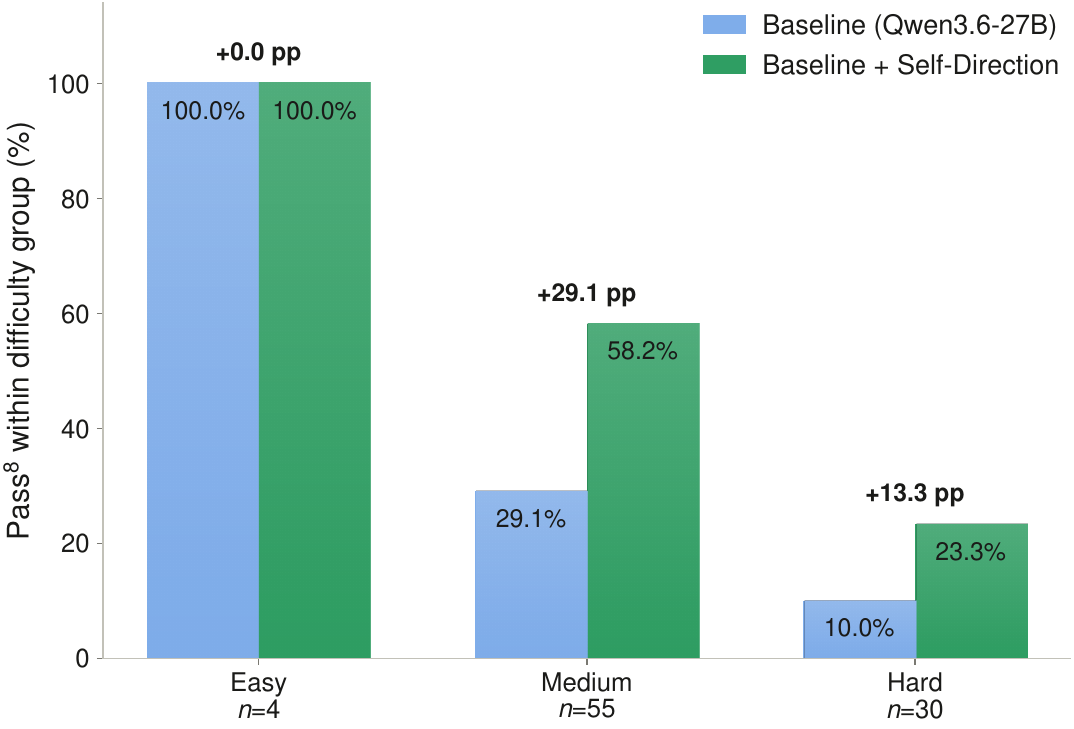}
          \caption{Pass$^8$ stratified by task difficulty.}
          \label{fig:tb-pass-power-difficulty}
      \end{subfigure}

      \caption{\textbf{\method improves repeated-run reliability on
      Terminal-Bench~2.1.}
      \textbf{Left:} pass$^k$, the fraction of tasks solved in each of $k$ runs,
      remains consistently higher for the full loop and increases from $25.8\%$
      to $48.3\%$ at $k=8$. Lines report the mean over all subsets of $k$ runs,
      with shaded bands showing the minimum and maximum. \textbf{Right:} The loop preserves
      $100\%$ pass$^8$ on easy tasks while increasing reliability from $29.1\%$
      to $58.2\%$ on medium tasks and from $10.0\%$ to $23.3\%$ on hard tasks.}
      \label{fig:tb-pass-8-comparison}
  \end{figure*}

Three runs establish that the improvement is repeatable, but a larger campaign provides a stronger view of run-to-run reliability. We, therefore, execute the complete loop eight times on the same 89 Terminal-Bench~2.1 tasks. \method improves accuracy in all eight runs, with gains between $9.0$ and $18.0$ percentage points. Mean success increases from $51.8\%$ for the initial base execution to $64.0\%$ for the full loop, an average gain of $12.2$ points (see Figs.~\ref{fig:terminal-eight-runs}, \ref{fig:tb-pass-8-comparison}).

The larger campaign also sharpens the distinction between coverage and reliability. The full loop increases pass@$8$ from $74.2\%$ to $80.9\%$, but its larger effect is on pass$^8$: the number of tasks solved in every run increases from $23/89$ to $43/89$, corresponding to a nearly doubled rate, from $25.8\%$ to $48.3\%$. The loop moves 20 tasks from intermittent success to 8/8 success. In addition, all 23 initially 8/8-reliable tasks remain reliable and six never-solved tasks gain some coverage. Thus, \method substantially enlarges the set of tasks solved consistently. 

This eight-run analysis measures benchmark-level consistency when the complete learning process is executed independently in every run. Since \method may use up to three rollouts per task, the comparison with the base agent may not be compute-matched. The compute-matched analysis on recovery from mistakes is presented in Section~\ref{sec:results-reliability}. It evaluates whether the same stored repair remains successful across repeated encounters.

In \Cref{tab:terminal-bench-2-1-qwen3-5-9b} 
%and \Cref{tab:terminal-bench-musacoder-27b} 
we show additional results for Qwen3.5-9B on Terminal-Bench.

\begin{table}[h]
\centering
\caption{Terminal-Bench 2.1 results for Qwen3.5-9B}
\label{tab:terminal-bench-2-1-qwen3-5-9b}
\resizebox{\columnwidth}{!}{%
\begin{tabular}{lrrrrr}
\hline
\textbf{Stage} & \textbf{Attempts} & \textbf{New wins} &
\textbf{Stage success} & \textbf{Cumulative wins} &
\textbf{Overall accuracy} \\
\hline
Phase 1                & 89 & 23 & 25.8\% & 23 & 25.8\% \\
Phase 4: adapter       & 65 &  7 & 10.8\% & 30 & 33.7\% \\
Phase 4: base fallback & 58 &  1 &  1.7\% & 31 & 34.8\% \\
\hline
\textbf{Final} & \textbf{89} & \textbf{31} & -- &
\textbf{31} & \textbf{34.8\%} \\
\hline
\end{tabular}
}
\end{table}

% \begin{table}[htbp]
% \centering
% \caption{Terminal-Bench results for
% \href{https://huggingface.co/MooreThreads/MusaCoder-27B}{MusaCoder-27B}.}
% \label{tab:terminal-bench-musacoder-27b}
% \resizebox{\textwidth}{!}{%
% \begin{tabular}{crrrrrrrr}
% \toprule
% \textbf{Shard} &
% \textbf{Tasks} &
% \textbf{Bootstrap wins} &
% \textbf{Bootstrap accuracy} &
% \textbf{Adapter wins / attempts} &
% \textbf{Base fallback wins / attempts} &
% \textbf{Final wins} &
% \textbf{Final accuracy} &
% \textbf{$\Delta$ accuracy} \\
% \midrule
% 0 & 12 & 6 & 50.0\% & 0 / 6 & 1 / 6 & 7 & 58.3\% & +8.3 pp. \\
% 1 & 11 & 3 & 27.3\% & 1 / 8 & 1 / 7 & 5 & 45.5\% & +18.2 pp. \\
% 2 & 11 & 2 & 18.2\% & 1 / 9 & 1 / 8 & 4 & 36.4\% & +18.2 pp. \\
% 3 & 11 & 6 & 54.5\% & 0 / 5 & 0 / 5 & 6 & 54.5\% & 0.0 pp. \\
% 4 & 11 & 3 & 27.3\% & 3 / 8 & 2 / 5 & 8 & 72.7\% & +45.5 pp. \\
% 5 & 11 & 3 & 27.3\% & 1 / 8 & 0 / 7 & 4 & 36.4\% & +9.1 pp. \\
% 6 & 11 & 5 & 45.5\% & 2 / 6 & 1 / 4 & 8 & 72.7\% & +27.3 pp. \\
% 7 & 11 & 5 & 45.5\% & 0 / 6 & 1 / 6 & 6 & 54.5\% & +9.1 pp. \\
% \midrule
% \textbf{Total} &
% \textbf{89} &
% \textbf{33} &
% \textbf{37.1\%} &
% \textbf{8 / 56} &
% \textbf{7 / 48} &
% \textbf{48} &
% \textbf{53.9\%} &
% \textbf{+16.9 pp.} \\
% \bottomrule
% \end{tabular}%
% }
% \end{table}

\subsection{Gains Across Difficulty, Category, and Task Structure}
\label{app:stratified-results}

Aggregate accuracy can conceal whether an improvement is concentrated in a narrow subset of tasks. We, therefore, stratify the results using the difficulty, category, and structural metadata provided by each benchmark. For every group, we report both mean accuracy across three runs and pass$^3$.

\paragraph{GAIA difficulty levels.}
Fig.~\ref{fig:gaia-expanded} separates the GAIA comparison into Levels~1-3. \method improves performance at every level, showing that the pooled gain is not driven only by easier tasks.

\paragraph{Claw-Eval categories and task structure.}
Table~\ref{tab:claw-eval-stratified} shows that the Claw-Eval improvement extends across heterogeneous workflows. On the two largest categories, Workflow and Ops, the full loop improves mean accuracy by $9.2$ and $9.7$ percentage points, respectively, while increasing pass$^3$ by $17.0$ and $16.2$ points. Multi-service tasks exhibit a similarly broad improvement, with mean accuracy increasing from $80.6\%$ to $91.1\%$ and pass$^3$ increasing from $66.7\%$ to $85.0\%$.

The gains are not limited to one execution structure. \method improves both state-changing and non-state-changing tasks, as well as safety-primary and non-safety-primary tasks. The task-structure rows represent three separate binary attributes and are, therefore, not mutually exclusive across attributes.

\begin{table*}[t]
      \centering
      \caption{\textbf{Claw-Eval results stratified by task category and
      structure.} Mean accuracy is computed across three independent runs, while
      pass$^3$ requires success in all three runs. Task categories partition the
      161 tasks. Task-structure rows describe three separate binary attributes:
      multi-service, state-changing, and safety-primary.}
      \label{tab:claw-eval-stratified}
      \scriptsize
      \setlength{\tabcolsep}{5pt}
      \renewcommand{\arraystretch}{1.15}

      \resizebox{\linewidth}{!}{%
      \begin{tabular}{@{}llcrrrr@{}}
          \toprule
          \textbf{Stratification}
          & \textbf{Group}
          & \textbf{N}
          & \multicolumn{2}{c}{\textbf{Mean Accuracy}}
          & \multicolumn{2}{c}{\textbf{Pass$^3$}} \\
          \cmidrule(lr){4-5}
          \cmidrule(lr){6-7}
          & &
          & \textbf{Initial base}
          & \textbf{Full loop}
          & \textbf{Initial base}
          & \textbf{Full loop} \\
          \midrule

          Overall
          & All tasks & 161 & 64.4\% & 73.3\% & 51.6\% & 64.6\% \\
          \midrule

          \multirow{10}{*}{Category}
          & Workflow         & 47 & 78.7\% & 87.9\% & 61.7\% & 78.7\% \\
          & Ops              & 31 & 89.2\% & 98.9\% & 80.6\% & 96.8\% \\
          & Finance          & 14 & 28.6\% & 40.5\% & 14.3\% & 28.6\% \\
          & Office QA        & 10 & 33.3\% & 43.3\% & 20.0\% & 20.0\% \\
          & Communication    & 8  & 75.0\% & 75.0\% & 75.0\% & 75.0\% \\
          & Productivity     & 7  & 61.9\% & 66.7\% & 42.9\% & 57.1\% \\
          & Operations       & 6  & 83.3\% & 88.9\% & 83.3\% & 83.3\% \\
          & Safety           & 5  & 53.3\% & 60.0\% & 40.0\% & 60.0\% \\
          & Terminal         & 5  & 13.3\% & 33.3\% & 0.0\%  & 0.0\%  \\
          & Other categories & 28 & 46.4\% & 54.8\% & 32.1\% & 46.4\% \\
          \midrule

          \multirow{6}{*}{Task structure}
          & Multi-service                  & 60  & 80.6\% & 91.1\% & 66.7\% & 85.0\% \\
          & Other general tasks            & 101 & 54.8\% & 62.7\% & 42.6\% & 52.5\% \\
          & State-changing action required & 39  & 71.8\% & 78.6\% & 59.0\% & 74.4\% \\
          & No expected state change       & 122 & 62.0\% & 71.6\% & 49.2\% & 61.5\% \\
          & Safety-primary                 & 65  & 71.3\% & 75.4\% & 58.5\% & 70.8\% \\
          & Not safety-primary             & 96  & 59.7\% & 71.9\% & 46.9\% & 60.4\% \\
          \bottomrule
      \end{tabular}%
      }
\end{table*}

\paragraph{Terminal-Bench difficulty and task categories.}
Table~\ref{tab:terminal-bench-2-1-stratified} shows that the improvement persists as the difficulty of Terminal-Bench 2.1 tasks increases. Easy tasks are already solved in every run by the base agent. On medium tasks, however, the full loop raises mean accuracy from $62.4\%$ to $78.2\%$ and pass$^3$ from $43.6\%$ to $70.9\%$. Hard-task accuracy increases from $25.6\%$ to $37.8\%$, whereas pass$^3$ increases from $10.0\%$ to $26.7\%$.

Furthermore, the category breakdown in Table~\ref{tab:terminal-bench-2-1-stratified} shows improvements across substantially different technical capabilities, including software engineering, system administration, data science, security, scientific computing, and file operations. Improvements in pass$^3$ are particularly pronounced for system administration, data science, security, and machine-learning tasks. The aggregate result, therefore, does not arise from repeatedly repairing one narrow class of terminal problems.

\begin{table*}[t]
      \centering
      \caption{\textbf{Terminal-Bench~2.1 results stratified by difficulty and
      task category.} Mean accuracy is computed across three independent runs,
      while pass$^3$ requires success in all three runs. \method improves
      both medium and hard tasks and produces gains across a broad range of
      technical categories.}
      \label{tab:terminal-bench-2-1-stratified}
      \scriptsize
      \setlength{\tabcolsep}{5pt}
      \renewcommand{\arraystretch}{1.15}

      \resizebox{\textwidth}{!}{%
      \begin{tabular}{@{}llcrrrr@{}}
          \toprule
          \textbf{Stratification}
          & \textbf{Group}
          & \textbf{N}
          & \multicolumn{2}{c}{\textbf{Mean Accuracy}}
          & \multicolumn{2}{c}{\textbf{Pass$^3$}} \\
          \cmidrule(lr){4-5}
          \cmidrule(lr){6-7}
          & &
          & \textbf{Initial base}
          & \textbf{Full loop}
          & \textbf{Initial base}
          & \textbf{Full loop} \\
          \midrule

          Overall
          & All tasks & 89 & 51.7\% & 65.5\% & 34.8\% & 57.3\% \\
          \midrule

          \multirow{3}{*}{Difficulty}
          & Easy   & 4  & 100.0\% & 100.0\% & 100.0\% & 100.0\% \\
          & Medium & 55 & 62.4\%  & 78.2\%  & 43.6\%  & 70.9\%  \\
          & Hard   & 30 & 25.6\%  & 37.8\%  & 10.0\%  & 26.7\%  \\
          \midrule

          \multirow{16}{*}{Category}
          & Software Engineering  & 26 & 38.5\%  & 47.4\%  & 26.9\%  & 42.3\%  \\
          & System Administration & 9  & 63.0\%  & 88.9\%  & 33.3\%  & 77.8\%  \\
          & Data Science           & 8  & 50.0\%  & 70.8\%  & 25.0\%  & 62.5\%  \\
          & Scientific Computing  & 8  & 29.2\%  & 37.5\%  & 12.5\%  & 37.5\%  \\
          & Security               & 8  & 75.0\%  & 91.7\%  & 62.5\%  & 87.5\%  \\
          & Debugging              & 5  & 86.7\%  & 93.3\%  & 60.0\%  & 80.0\%  \\
          & File Operations        & 5  & 33.3\%  & 60.0\%  & 20.0\%  & 40.0\%  \\
          & Data Processing        & 4  & 83.3\%  & 83.3\%  & 75.0\%  & 75.0\%  \\
          & Mathematics            & 4  & 25.0\%  & 41.7\%  & 0.0\%   & 25.0\%  \\
          & Model Training         & 4  & 66.7\%  & 83.3\%  & 50.0\%  & 75.0\%  \\
          & Machine Learning       & 3  & 66.7\%  & 88.9\%  & 33.3\%  & 66.7\%  \\
          & Data Querying          & 1  & 100.0\% & 100.0\% & 100.0\% & 100.0\% \\
          & Games                  & 1  & 0.0\%   & 33.3\%  & 0.0\%   & 0.0\%   \\
          & Optimization           & 1  & 100.0\% & 100.0\% & 100.0\% & 100.0\% \\
          & Personal Assistant     & 1  & 100.0\% & 100.0\% & 100.0\% & 100.0\% \\
          & Video Processing       & 1  & 0.0\%   & 0.0\%   & 0.0\%   & 0.0\%   \\
          \bottomrule
      \end{tabular}%
      }
\end{table*}

Together, these stratifications show that failure-conditioned \method remains useful across task types, execution structures, and difficulty levels. Results for very small groups should be interpreted descriptively, particularly categories containing fewer than five tasks.

\subsection{Controlling for the Number of Agent Rollouts}

\label{sec:results-efficiency}

\emph{Agent rollouts per task} measure computational cost, counting task-environment executions only and excluding reflection, synthesis, and adapter training. We compare the loop's accuracy and rollout count against the base pass@$k$ curve.

\begin{figure*}[t]
    \centering
    \includegraphics[width=\textwidth]{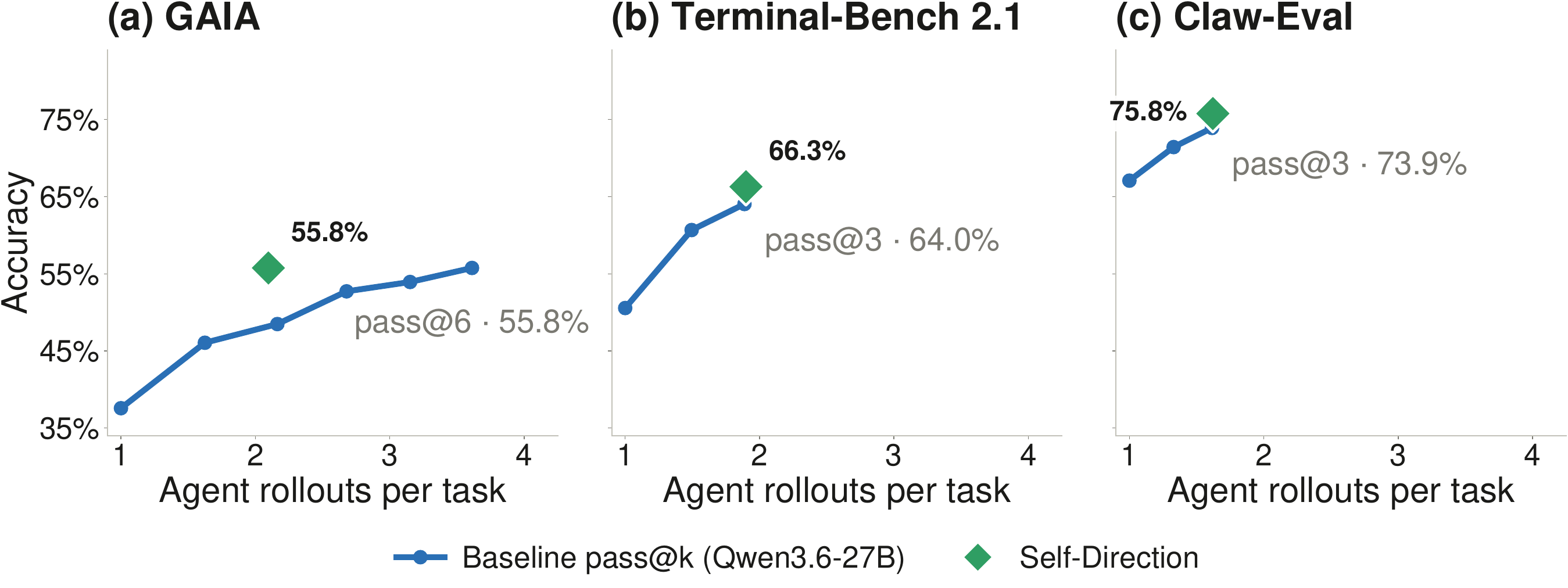}
    \caption{\textbf{\method reaches higher accuracy with comparable task-environment rollouts.} Relative to base pass@$k$, \method outperforms the pass@$k$ baseline across all three benchmarks while using a comparable number of full agent rollouts. Reflection, synthesis, and adapter optimization details are presented in Appendix~\ref{app:adapter-compute}.}
    \label{fig:rollout-efficiency}
\end{figure*}

\paragraph{\method improves accuracy at comparable task-environment rollout budgets.} 
As shown in Fig.~\ref{fig:rollout-efficiency}, the loop reaches $55.8\%$ accuracy on GAIA, with only $2.10$ agent rollouts per task, matching the accuracy achieved by base pass@$6$. On Terminal-Bench~2.1, it reaches $66.3\%$ accuracy with $1.90$ rollouts per task, exceeding the $65.3\%$ obtained by pass@$3$ using a similar number of rollouts. On Claw-Eval, the loop exceeds the accuracy of pass@$3$ while requiring a comparable average of $1.54$ rollouts per task. Across all three domains, the loop obtains higher accuracy than simple retrying while using comparable or fewer task-environment rollouts, showing that \method uses rollouts efficiently compared with standard resampling. While this reinforces the strength of spending additional test-time compute, simply resampling parallel rollouts does not retain any lessons; the agent is just as likely to repeat the mistake if the same task is encountered again. 
%Note that the reported \method numbers here are selected from the best run on the benchmark for context; the relevant comparisons with per benchmark means are displayed in Figs.~\ref{fig:headline} and \ref{fig:cross-domain-headline}.

\subsection{Adapters Trained in the GAIA Deployment Stream}
\label{app:gaia-adapters-visual}
\begin{figure}[h]
  \centering
  \includegraphics[width=\linewidth]{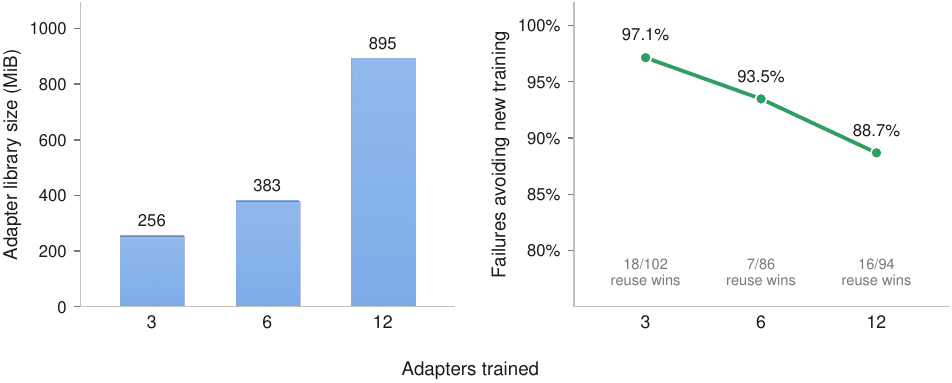}
  \caption{\textbf{A compact library of lessons amortizes learning across a deployment-like GAIA stream.} Existing lessons avoid new adapter training for $88.7$-$97.1\%$ of initial failures. Bars show adapters trained; annotations report storage and successful reuses; the green line shows training avoided
  through reuse.}
  \label{fig:gaia-amortization}
\end{figure}

Fig.~\ref{fig:gaia-amortization} reports the number of adapters trained for each of the P25, P50, and P75 thresholds in the GAIA deployment stream experiments, as described in Section~\ref{sec:results-transfer}. Each new adapter occupies a marginal $85$ MB of storage. Qualitative examples of recoveries from these adapters are further presented in Appendix~\ref{app:gaia-transfer-analysis}.

\subsection{Preserving Existing Lessons}
\label{sec:app-preserving}

\Cref{tab:cross_benchmark_thresholds} reports similarity thresholds that prevent retrieval of unrelated adapters in the evaluated cross-benchmark settings. A global threshold of 0.55 prevents all observed cross-benchmark retrievals with a full library of lessons from GAIA, Claw-Eval, and Terminal-Bench~2.1.%(Table~\ref{tab:cross_benchmark_thresholds}).

\begin{table*}[h]
\centering
\caption{\textbf{\method learns new lessons without disturbing existing capabilities.}
The safe threshold is the smallest two-decimal cutoff above the largest similarity observed across the pair. At this threshold, no evaluated task retrieves an adapter from the other domain in either direction, preventing cross-domain adapter application ($\checkmark$).}
\label{tab:cross_benchmark_thresholds}

\small
\renewcommand{\arraystretch}{1.2}
\resizebox{\linewidth}{!}{%
\begin{tabular}{lcccc}
\hline
\textbf{Benchmark Pair}
& \textbf{Max. Similarity}
& \textbf{Safe $\boldsymbol{\tau}$}
& \textbf{Cross-Domain Retrievals}
& \textbf{Capabilities Preserved} \\
\hline
GAIA--TB 2.1
& 0.546 & 0.55 & 0 / 145 & $\checkmark$ \\

GAIA--Claw
& 0.415 & 0.42 & 0 / 141 & $\checkmark$ \\

TB 2.1--Claw
& 0.306 & 0.31 & 0 / 84 & $\checkmark$ \\
\hline
All three, one global $\tau$
& 0.546 & 0.55 & 0 / 370 & $\checkmark$ \\
\hline
\end{tabular}
}
\end{table*}

\begin{figure}
    \centering
    \includegraphics[width=0.9\linewidth]{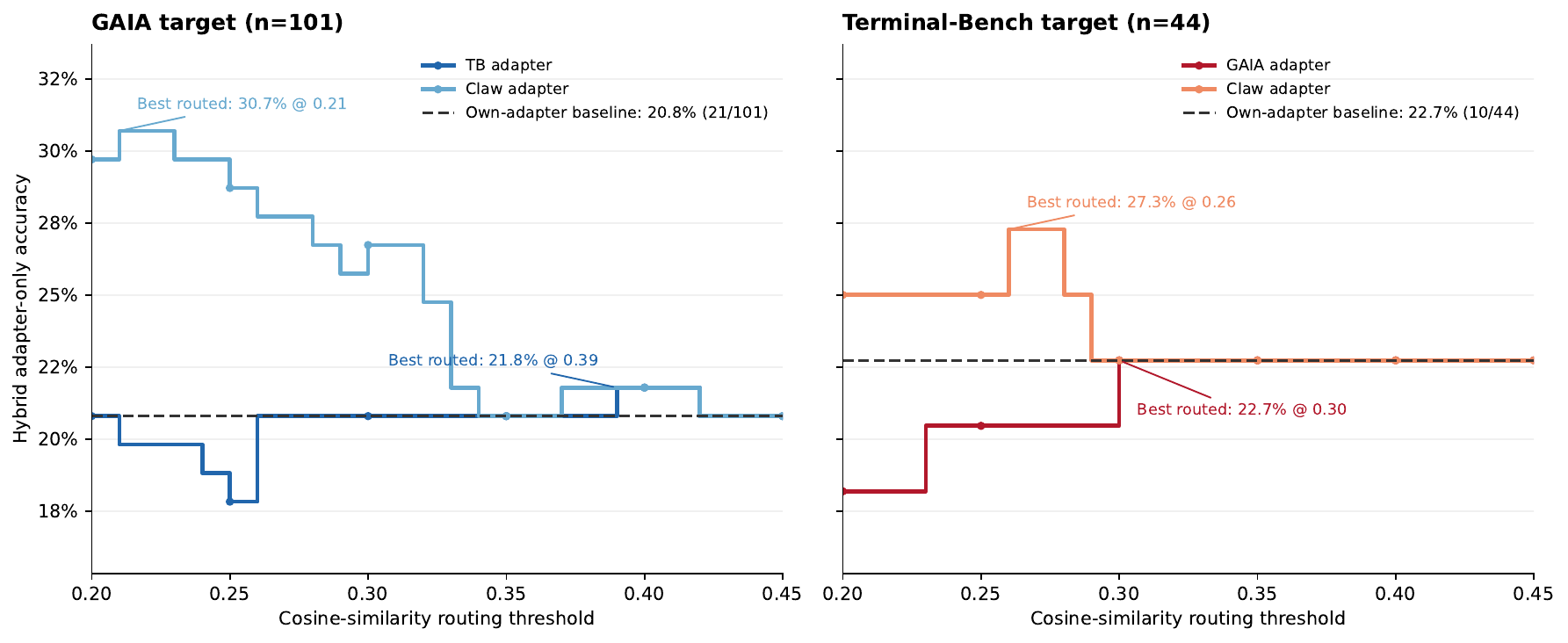}
    \caption{\textbf{Accuracy across router thresholds $\tau$ with a fixed library of cross-benchmark lessons.} \textbf{Left:} GAIA tasks failed by the base agent are retried with the corresponding top-matched adapter from Terminal-Bench and Claw-Eval libraries. \textbf{Right:} Terminal-Bench 2.1 tasks failed by the base agent are retried with the corresponding top-matched adapter from GAIA and Claw-Eval libraries. We see existing capabilities are preserved except for low thresholds dipping below 0.30, where adapters from other benchmarks in some cases can help or hurt accuracy on the new benchmark.}
    \label{fig:threshold_accuracy}
\end{figure}

\paragraph{Existing capabilities are preserved for reasonable threshold values.} In addition to evaluating preservation of existing capabilities with a global threshold, we set up several experiments where the threshold $\tau$ decreases below the cross-benchmark exclusion threshold, allowing selection and loading of adapters from other benchmarks. We vary $\tau$ from 0.2 to 0.6 in increments of 0.01 and present the results in Fig.~\ref{fig:threshold_accuracy}. We find that except for low thresholds (dipping below $0.30$), task success on GAIA and Terminal-Bench~2.1 is preserved even when the library contains lessons from the two other unrelated benchmarks. When $\tau$ is very low ($\tau < 0.35$), we find that adapters from other benchmarks can help or hurt, depending on benchmark transfer.

\subsection{Statistical Comparison on GAIA}
\label{app:t-test-gaia}
We compare \method with its initial base-agent execution using a paired, task-level $t$-test. The unit of analysis is an individual GAIA task, rather than an evaluation run or difficulty level.

For each of the $n=165$ tasks, let
  \[
  B_i \in \{0,1\}
  \]
  denote whether the initial base-agent attempt is correct and let
  \[
  S_i \in \{0,1\}
  \]
  denote whether the final output after executing the \method loop is
  correct. We define the paired difference
  \[
  D_i = S_i - B_i.
  \]

  The observed outcomes are:
  \begin{center}
  \begin{tabular}{lrr}
  \toprule
  Outcome & Count & Rate \\
  \midrule
  Base correct & 64 & 38.8\% \\
  \method correct & 92 & 55.8\% \\
  Recovered by \method & 28 & 17.0\% \\
  \bottomrule
  \end{tabular}
  \end{center}

Thus, 64 tasks are solved by both stages, 28 are solved only after \method, and 73 are unsolved by both. The mean paired difference is
  \[
  \bar{D} = \frac{28}{165} = 0.1697.
  \]

  We compute the test statistic as
  \[
  t = \frac{\bar{D}}{s_D/\sqrt{n}},
  \]
  where $s_D$ is the sample standard deviation of the 165 paired differences.
  The two-sided test gives
  \[
  t(164) = 5.79,
  \qquad
  p = 3.50 \times 10^{-8}.
  \]
  The corresponding 95\% confidence interval for the accuracy improvement is
  $[11.2,\,22.8]$ percentage points.

The results are pooled at the task level across GAIA Levels 1-3, comprising 53 Level-1, 86 Level-2, and 26 Level-3 tasks.

\clearpage

\section{\method with a Stronger Base Model}
\label{app:cross-model-results}

The benefits of \method should not depend on beginning with a weak base agent. We test this by applying the same loop to Qwen3.8-27B and comparing it with Qwen3.6-27B on identical GAIA, Claw-Eval, and Terminal-Bench~2.1 task sets.

As expected, Qwen3.8-27B solves more tasks during its initial execution on all three benchmarks. \method nevertheless provides substantial additional gains. On GAIA, it recovers 27 failures and raises accuracy from $59.4\%$ to $75.8\%$ ($+16.4$ pp). On Claw-Eval, it raises accuracy from $64.6\%$ to $76.4\%$ ($+11.8$ pp), whereas on Terminal-Bench~2.1, it improves accuracy from $55.1\%$ to $73.0\%$ ($+17.9$ pp). The corresponding Qwen3.6-27B runs also improve on every benchmark, with gains between $15.5$ and $18.0$ percentage points.

These results support the central role of failure as an intervention point. As the base model becomes more capable, fewer tasks enter the mistake subset but the remaining failures still provide useful
signals for targeted in-weight repair. \method, therefore, complements improvements in the underlying model rather than merely compensating for a weaker checkpoint. This experiment establishes
robustness across two capability levels within the same model family; broader generalization across model families and scales remains an important direction for future work.

\begin{figure*}[h]
    \centering
    \includegraphics[width=0.98\textwidth]{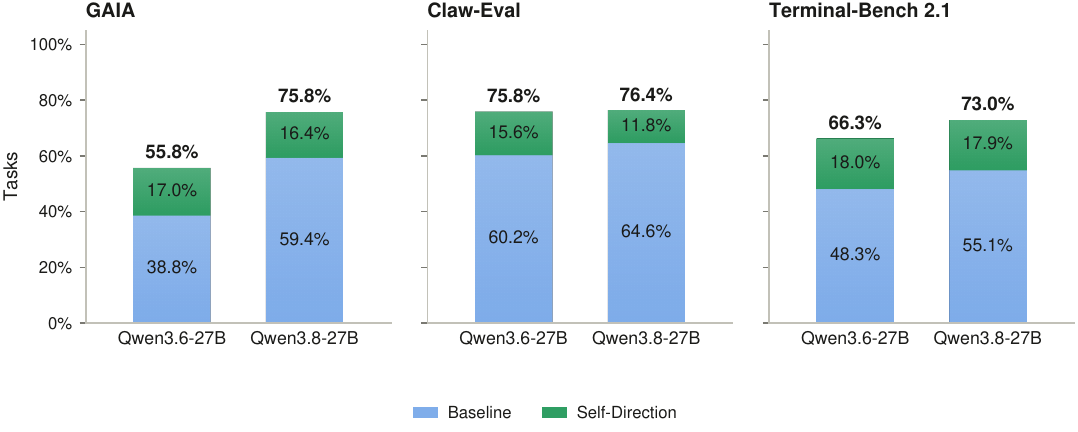}
    \caption{\textbf{\method continues to improve a stronger base model.}
    We apply the same failure-conditioned loop to Qwen3.6-27B and Qwen3.8-27B on identical benchmark tasks. Blue denotes tasks solved by initial base execution, green denotes additional tasks recovered by the complete loop. \method produces substantial gains for both models across GAIA, Claw-Eval, and Terminal-Bench~2.1.}
    \label{fig:cross-model-results}
\end{figure*}

\clearpage

\section{Why \method Uses Modular In-Weight Lessons}
\label{app:skill-representation}

\method relies on two design choices. Each failure produces a distinct lesson adapter, and relevant experience is retrieved on a later task. We examine both choices on GAIA. We first compare failure-specific adapters with pooled, mismatched, and randomly initialized LoRA alternatives. We then quantify why retaining the complete learning history directly in context does not scale.

\subsection{Lesson Specificity and Recovery}
\label{app:adapter-specificity}

\begin{figure*}[h]
      \centering
      \includegraphics[width=0.96\textwidth]{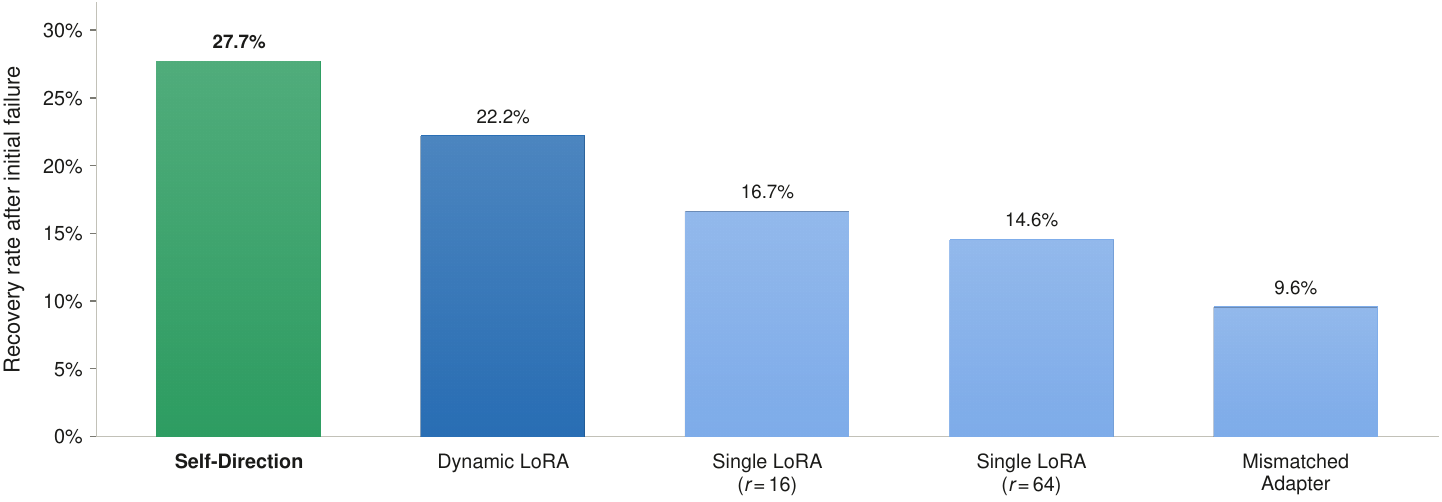}
      \caption{\textbf{Recovery after an initial failure on GAIA.}
      The complete \method loop recovers $27.7\%$ of the failures. Text memory stores lessons in \texttt{Skills.md}; the single-LoRA baselines pool experience across failures and the mismatched condition applies an unrelated adapter.}
      \label{fig:gaia-recovery-ablation}
\end{figure*}

Adding or modifying existing parameters is useful only when it captures a lesson relevant to the failure being repaired. Fig.~\ref{fig:gaia-recovery-ablation} compares the complete \method policy with several alternative representations of the same failure-derived experience.

The full loop recovers $27.7\%$ of the initial failures. A Dynamic LoRA adapter with Gaussian initialization instead of SVD initialization recovers $22.2\%$ of the initial failures. A textual \texttt{Skills.md} memory has a recovery rate of $18.1\%$, showing that explicit textual lessons can provide useful immediate guidance. However, a single pooled LoRA recovers only $16.7\%$ and $14.6\%$ of attempted failures at ranks 16 and 64, respectively. Deliberately applying a mismatched adapter by choosing the least similar one reduces recovery further to $9.6\%$, which is lower than the recovery rate of base retrying with pass@2.

These results offer evidence that the benefit does not arise from inserting an arbitrary parameter update or increasing adapter capacity. Recovery is strongest when the learned lesson remains aligned with the failure that produced it.

\subsection{Why Not Place Every Lesson in Context?}
\label{app:context-scaling}

Textual memory is useful but accumulating every lesson in the active context is not a scalable form of persistence. The 101 failures in the GAIA \method run produce 2,428 corrective examples: 672 from Level~1, 1,323 from Level~2, and 433 from Level~3. Using the Qwen3.6 tokenizer, their instruction-response content is estimated to contain approximately 362k tokens, implying that each self-reflection loop produces an additional $\sim3.6k$ tokens per failure. Including chat formatting and stored rationales increases this total to approximately 410k tokens.

This already exceeds the model's native 262,144-token context window. Moreover, the run reserves up to 16,384 tokens for generation, leaving at most 245,760 tokens for the input. Even the pooled synthesized responses across all failures require approximately 276k tokens, before including their instructions, the current task, system instructions, tool definitions, or the agent's ongoing trajectory.

The failed executions are larger still. Their full trajectories contain approximately 2.72M tokens. The compact representations supplied to lesson synthesis contain 188k tokens after truncation but combining even these compact failures with the generated lessons requires approximately 599k tokens. Retaining the complete failure trajectories and lessons would require roughly 3.13M tokens, or nearly twelve native context windows. The summary of this analysis is presented in Table~\ref{tab:gaia-context-capacity}.

\begin{table*}[t]
      \centering
      \caption{\textbf{Context required to retain the complete GAIA learning
      history.} Counts use the Qwen3.6 tokenizer. Totals are rounded because this section is intended as a capacity analysis.}
      \label{tab:gaia-context-capacity}
      \small
      \setlength{\tabcolsep}{8pt}
      \begin{tabular}{@{}lrr@{}}
          \toprule
          Stored content & Approximate tokens & Native context windows \\
          \midrule
          Full trajectories for 101 failures
              & $2.72$M & $10.4\times$ \\
          Truncated failure representations
              & 188K & $0.72\times$ \\
          2,428 lesson instruction-response pairs
              & 362K & $1.38\times$ \\
          Lessons with formatting and rationales
              & 410K & $1.56\times$ \\
          % Truncated failures and lessons
          %     & 599K & $2.29\times$ \\
          Full trajectories and lessons
              & $3.13$M & $11.9\times$ \\
          \bottomrule
      \end{tabular}
\end{table*}

The implication is not that textual memory is ineffective. The \texttt{Skills.md} baseline demonstrates that a small, relevant textual lesson can improve recovery. Rather, a continually growing agent cannot reinsert its entire learning history on every task. It must select a small amount of relevant experience. \method performs this selection over modular lesson adapters, enabling acquired lessons to persist without occupying the active prompt context.

\clearpage

\section{Agent Execution and Training Hyperparameters}
\label{app:hyperparameters}

We use the same agent-generation and adapter-training configuration across benchmarks unless stated otherwise. This section separates the parameters governing task execution from those used to construct each modular lesson adapter.

\subsection{Agent Generation and ReAct Execution}
\label{app:agent-hyperparameters}

The agent operates through a ReAct-style loop \citep{yao2022react} that interleaves model decisions with environment interaction. At each step, the model receives the task and accumulated trajectory, then emits either a tool call or a final response. A tool call is executed by the environment, its observation is appended to the trajectory, and the model is invoked again. A rollout terminates when the model provides a final answer or exhausts its interaction budget.

The same decoding configuration is used for the initial base execution, adapter-guided retry, and base fallback (see Tables~\ref {tab:agent-generation-hparams}, \ref{tab:lora-training-hparams}). The larger Terminal-Bench budget accommodates longer coding and environment-manipulation tasks. These values are upper bounds: A rollout terminates early whenever the agent completes the task or emits a final answer without any tool calls.

\begin{table}[h]
      \centering
      \caption{\textbf{Agent generation and interaction hyperparameters.}
      The token limit applies independently to each model turn within the ReAct
      loop, rather than to the complete task trajectory.}
      \label{tab:agent-generation-hparams}
      \small
      \setlength{\tabcolsep}{5pt}
      \begin{tabular}{@{}lcc@{}}
          \toprule
          Hyperparameter & GAIA / Claw-Eval & Terminal-Bench~2.1 \\
          \midrule
          Temperature & $0.6$ & $0.6$ \\
          Top-$p$ & $0.95$ & $0.95$ \\
          Maximum new tokens per turn & 16,384 & 16,384 \\
          Maximum ReAct steps & 50 & 100 \\
          Maximum tool calls & 100 & 100 \\
          Stop generation after tool call & Yes & Yes \\
          \bottomrule
      \end{tabular}
\end{table}

\subsection{Adapter Synthesis and Training}
\label{app:lora-hyperparameters}

Each failed execution produces up to 24 corrective examples. Adapter training begins only when at least eight valid examples are available. All base-model parameters remain fixed. Each lesson is represented by a rank-16 LoRA adapter initialized from the targeted base-model weights using the SVD procedure described in Section~\ref{sec:method}. The first LoRA factor is initialized from the leading right singular vectors and frozen, whereas the second factor begins at zero and is optimized on the corrective examples.

The target modules depend on the normalized failure reason rather than the benchmark. Failures involving planning or interaction limits update the attention projections, \texttt{q\_proj} and \texttt{v\_proj};  incorrect answers and execution exceptions update the multi-layer perceptron (MLP) \texttt{down\_proj} projection. This keeps the in-weight repair localized to the type of behavior implicated by the failure.

\begin{table*}[h]
      \centering
      \caption{\textbf{Adapter synthesis and training hyperparameters.}
      These settings are shared across GAIA, Claw-Eval, and Terminal-Bench~2.1.
      Unlisted \texttt{TrainingArguments} retain their library defaults.}
      \label{tab:lora-training-hparams}
      \small
      \setlength{\tabcolsep}{7pt}
      \begin{tabular}{@{}lll@{}}
          \toprule
          Component & Hyperparameter & Value \\
          \midrule
          Corrective data
              & Maximum examples per failure & 24 \\
          Corrective data
              & Minimum examples required & 8 \\
          Corrective data
              & Maximum assistant-response length & 16,384 tokens \\
          Corrective data
              & Trajectory steps retained for reflection & 24 \\
          \midrule
          LoRA
              & Rank $r$ & 16 \\
          LoRA
              & Scaling parameter $\alpha$ & 16 \\
          LoRA
              & Dropout & 0.05 \\
          LoRA
              & Bias parameters & None \\
          LoRA
              & Target modules & Failure-dependent \\
          \midrule
          SVD initialization
              & Initialized factor & $A$ from leading right singular vectors \\
          SVD initialization
              & Trainable factor & $B$ only; initialized to zero \\
          SVD initialization
              & Power iterations & 2 \\
          SVD initialization
              & Oversampling dimension & 8 \\
          SVD initialization
              & SVD device & CPU, cached across lesson adapters \\
          \midrule
          Optimization
              & Learning rate & $5\times10^{-4}$ \\
          Optimization
              & Training epochs & 2 \\
          Optimization
              & Per-device batch size & 1 \\
          Optimization
              & Gradient accumulation steps & 8 \\
          Optimization
              & Effective batch size & 8 \\
          Optimization
              % & Random seed & 42 \\
          Optimization
              & Gradient checkpointing & Enabled \\
          Optimization
              & Numerical precision & BF16 when supported \\
          \bottomrule
      \end{tabular}
\end{table*}

Reflection uses the same base model with temperature $0.6$. Its maximum generation length is 16,384 tokens for all benchmarks. No external teacher model is used.

\paragraph{Failure normalization.}
We normalize each failure using the agent's termination signal, the presence of an extractable response, and the binary verifier outcome. Budget exhaustion and execution exceptions take precedence over output-based labels. Otherwise, an empty response is assigned \texttt{no\_final\_response}, whereas a completed response rejected by the verifier is assigned \texttt{wrong\_answer}. Table~\ref{tab:normalized-failure-reasons} summarizes the resulting vocabulary.

\begin{table*}[t]
  \centering
  \caption{\textbf{Normalized failure reasons used by \method.}
  Each categorical reason determines which projections are updated during adaptation.}
  \label{tab:normalized-failure-reasons}
  \small
  \setlength{\tabcolsep}{4pt}
  \renewcommand{\arraystretch}{1.15}
  \resizebox{\linewidth}{!}{%
  \begin{tabular}{@{}p{0.25\linewidth}p{0.56\linewidth}p{0.18\linewidth}@{}}
  \toprule
  Reason & Operational definition & Target modules \\
  \midrule

  \texttt{step\_budget}
  & The agent reaches its maximum number of interaction steps before verified completion.
  & \texttt{q\_proj}, \texttt{v\_proj} \\

  \texttt{tool\_call\_budget}
  & The agent exhausts its permitted tool calls before completing the task.
  & \texttt{q\_proj}, \texttt{v\_proj} \\

  \texttt{no\_final\_response}
  & The execution ends without an extractable final response and was not terminated by a budget or exception.
  & \texttt{q\_proj}, \texttt{v\_proj} \\

  \texttt{wrong\_answer}
  & The agent produces a response but the verifier rejects the resulting answer or environment state.
  & \texttt{down\_proj} \\

  \texttt{exception}
  & An execution error prevents the agent from completing the task.
  & \texttt{down\_proj} \\

  \bottomrule
  \end{tabular}
  }
  \end{table*}

  The normalized reason, therefore, affects adaptation in two ways: It focuses reflection on the procedural deficiency exposed by the failure and determines where the resulting lesson is represented in the adapter. This routing is a lightweight localization prior rather than an assumption that planning and
  answer correctness reside in disjoint model components.

\paragraph{Information boundary.}
Normalized failure reasons are deliberately categorical and low-bandwidth. Let \(t\) denote the agent's termination status, \(e\) indicate whether an extractable response was produced, and  \(z=\mathcal{V}(x,\tau)\in\{0,1\}\) denote the verifier's binary decision. The reason exposed to reflection and routing is
  \[
  r=\operatorname{Normalize}(t,e,z).
  \]
The reference answer, hidden-test contents, and verifier-internal artifacts are not arguments to this function. In particular, \(r\) contains neither gold-answer tokens nor test outputs, judge scores, rubrics, rationales, or free-form judge feedback. The loop observes only the categorical outcome of verification, not the information used internally to compute it.

We additionally enumerated every persisted reason value in the reported GAIA, Claw-Eval, and Terminal-Bench~2.1 runs. The observed values of \(r\) were exclusively \texttt{step\_budget}, \texttt{wrong\_answer}, and \texttt{exception}; no stored reason contained task-specific text, answers, test results, scores, or judge feedback. Thus, the failure classification and module-routing decisions cannot communicate how to solve the task. They convey only a coarse description of how the preceding execution failed.

\subsection{Adapter Training and Compute Requirements}
\label{app:adapter-compute}

\method performs a small amount of training only after a mistake is detected, triggering self-reflection and adapter training. Each reflection call produces $\sim24$ structured instruction-response examples. Measured with the Qwen3.6-27B tokenizer and the actual chat template used for training, these examples contain an average of 3,622 tokens per adapter. Rationales generated during reflection are retained for analysis but are not included in the training loss.

The fixed base model is adapted using rank-16 LoRA with SVD initialization. The LoRA $A$ matrices are initialized from cached right-singular directions of the corresponding base-model weights and remain frozen; only the $B$ matrices are optimized. Training uses two epochs, a per-device batch size of one, gradient accumulation over eight examples, BF16 precision, and gradient checkpointing on one NVIDIA H200. Thus, an average adapter processes approximately 7,244 token presentations across both epochs.

\paragraph{Measurement protocol.}
We report synthesis costs calculated from analyzing 46 GAIA failures, 39 Claw-Eval failures, and 45 Terminal-Bench 2.1 failures (130 in total). All stages execute sequentially on one NVIDIA H200. Synthesis time covers failure reflection and example generation, optimization time is the recorded trainer runtime, and end-to-end time extends from the beginning of reflection until the adapter is saved. The initial task execution and subsequent task-environment rollouts are excluded from these timings and reported separately in Fig.~\ref{fig:rollout-efficiency}.

  \begin{table*}[h]
  \centering
  \caption{\textbf{Cost of constructing a modular lesson adapter.}
  Values are averages over 130 failures across the three benchmarks. End-to-end latency includes reflection, data preparation, optimization, and adapter serialization on one NVIDIA H200. Energy denotes measured GPU energy over the same interval.}
  \label{tab:adapter-compute}
  \small
  \setlength{\tabcolsep}{6pt}
  \begin{tabular}{@{}lrrrrrr@{}}
  \toprule
  Benchmark
  & Adapters
  & Training tokens
  & Synthesis
  & Optimization
  & End-to-end
  & GPU energy \\
  & & per adapter
  & (min)
  & (s)
  & (min)
  & (Wh) \\
  \midrule
  GAIA                & 46 & 4,059 & 8.51 & 80.9 & 9.89 & 37.0 \\
  Claw-Eval           & 39 & 3,709 & 5.82 & 80.3 & 7.19 & 26.5 \\
  Terminal-Bench 2.1  & 45 & 3,100 & 5.79 & 81.4 & 7.22 & 27.1 \\
  \midrule
  \textbf{Overall}    & \textbf{130} & \textbf{3,622}
  & \textbf{6.76} & \textbf{80.9} & \textbf{8.16}
  & \textbf{30.4} \\
  \bottomrule
  \end{tabular}
  \end{table*}

\paragraph{Latency and energy.}
Reflection and data synthesis take 6.76 minutes, on average, on our setup, followed by 80.9 seconds ($\sim1.5$ minutes) of adapter optimization. The complete post-failure construction process, therefore, takes 8.16 minutes, on average, and 6.95 minutes at the median. Average measured GPU energy is 30.4 Wh per adapter, comprising 26.6 Wh for synthesis and 3.68 Wh for optimization. Across the 130 measured adapters, this corresponds to approximately 17.7 H200 GPU-hours and 3.96 kWh of GPU energy. These measurements cover the GPU only and exclude host, storage, and datacenter overhead.

\paragraph{Trainable parameters and storage.}
  The number of trainable parameters depends on the normalized failure reason and its corresponding target modules. Attention adapters that target \texttt{q\_proj} and
  \texttt{v\_proj} optimize 3.41M parameters, or $0.013\%$ of the 26.90B-parameter base model. MLP adapters that target \texttt{down\_proj} optimize 5.24M parameters, or
  $0.0195\%$. Thus, each repair trains at most approximately $0.02\%$ additional parameters while leaving the base model fixed.

  Because the frozen SVD-initialized $A$ matrices must also be retained, the complete adapter tensors contain 6.03M parameters for attention repairs and 23.07M parameters for MLP
  repairs. Their serialized weight files occupy 23.0 MiB and 88.0 MiB, respectively. The recorded artifact directories occupy 42.1-107.1 MiB because each directory also
  contains an identical 19.1 MiB tokenizer file; this shared file can be stored once when maintaining an adapter library.

  \paragraph{Approximate optimization FLOPs.}
  Let $N=26.90$B denote the number of base-model parameters and 
  $D=2\times3{,}622=7{,}244$ denote the average number of token presentations
  across two epochs. A forward pass through the base model costs approximately
  $2ND$. Backpropagation must propagate activation gradients through the fixed
  base model but does not compute updates for its weights, giving an approximate
  training cost of
  \[
      C_{\mathrm{adapter}} \approx 4ND
      \approx 7.8\times10^{14}\ \text{FLOPs}
      = 0.78\ \text{PFLOPs}
  \]
  per adapter, before the comparatively small LoRA and optimizer overhead. For
  reference, applying the conventional full-parameter training approximation
  of $6ND$ yields a conservative upper bound of approximately $1.17$ PFLOPs. These are analytical estimates rather than hardware-counter measurements; the latency and energy values in Table~\ref{tab:adapter-compute} are measured
  directly.

\subsection{Benchmark Tool Interfaces}
\label{app:benchmark-tools}

Each benchmark is evaluated using its native interaction interface. The initial base execution, adapter-guided retry, and base fallback receive the same tools. Hence, improvements cannot arise from exposing additional capabilities after failure. During reflection, \method also receives the tools available in the failed execution, enabling it to generate corrective examples that refer to valid actions.

GAIA uses a fixed collection of general-purpose research and file-processing tools. Claw-Eval exposes task-specific subsets of a larger service-oriented tool collection and its nine sandbox tasks provide shell and file-manipulation tools in addition. Terminal-Bench~2.1 exposes a single persistent shell inside a fresh task container (see Table~\ref{tab:benchmark-tools}).

  \begin{table*}[t]
      \centering
      \caption{\textbf{Tool interfaces used by each benchmark.}
      The Claw-Eval row reports the union of tools across the evaluated tasks;
      each individual task exposes only its task-specific subset. Tool names are
      shown as presented to the agent.}
      \label{tab:benchmark-tools}
      \scriptsize
      \setlength{\tabcolsep}{7pt}
      \renewcommand{\arraystretch}{1.18}
      \begin{tabular}{@{}p{0.14\textwidth}p{0.81\textwidth}@{}}
          \toprule
          Benchmark & Available tools \\
          \midrule

          GAIA &
          \texttt{web\_search}, \texttt{web\_fetch},
          \texttt{python\_exec}, \texttt{shell},
          \texttt{file\_read}, \texttt{file\_open}, and
          \texttt{list\_dir}. These tools support web research, numerical
          computation, shell commands, directory inspection, and extraction from
          text, PDF, spreadsheet, HTML, and image files. \\

          \addlinespace
          Claw-Eval &
          \textbf{Email and contacts:}
          \texttt{gmail\_list\_messages},
          \texttt{gmail\_get\_message},
          \texttt{gmail\_save\_draft},
          \texttt{gmail\_send\_message},
          \texttt{contacts\_search}, \texttt{contacts\_get}, and
          \texttt{contacts\_send\_message}.
          \textbf{Calendar, tasks, and scheduling:}
          \texttt{calendar\_list\_events},
          \texttt{calendar\_get\_event},
          \texttt{calendar\_create\_event},
          \texttt{calendar\_get\_user\_events},
          \texttt{calendar\_delete\_event},
          \texttt{todo\_list\_tasks}, \texttt{todo\_get\_task},
          \texttt{todo\_update\_task}, \texttt{todo\_create\_task},
          \texttt{todo\_delete\_task},
          \texttt{scheduler\_list\_jobs},
          \texttt{scheduler\_get\_job},
          \texttt{scheduler\_job\_history},
          \texttt{scheduler\_create\_job},
          \texttt{scheduler\_update\_job}, and
          \texttt{scheduler\_delete\_job}.
          \textbf{Business services:}
          finance, CRM, inventory, reporting, helpdesk, and integration-management
          tools for listing, inspecting, updating, and creating benchmark
          entities.
          \textbf{Knowledge and content:}
          \texttt{kb\_search}, \texttt{kb\_get\_article},
          \texttt{kb\_update\_article},
          \texttt{notes\_list}, \texttt{notes\_get},
          \texttt{notes\_share},
          \texttt{rss\_list\_feeds}, \texttt{rss\_list\_articles},
          \texttt{rss\_get\_feed}, \texttt{rss\_get\_article},
          \texttt{rss\_publish}, \texttt{web\_search}, and
          \texttt{web\_fetch}.
          \textbf{Documents and media:}
          \texttt{documents\_extract\_text},
          \texttt{ocr\_extract\_text}, and
          \texttt{caption\_describe\_image}.
          The nine sandbox tasks additionally expose
          \texttt{Bash}, \texttt{Read}, \texttt{Write}, \texttt{Edit},
          \texttt{Glob}, and \texttt{Grep} inside an isolated workspace. \\

          \addlinespace
          Terminal-Bench~2.1 &
          \texttt{terminal\_shell}, which executes Bash commands inside a fresh,
          task-specific Podman container. Commands share a persistent filesystem
          for the duration of an attempt and return their exit code, standard
          output, and standard error. The shell supports inspecting files and
          processes, editing or creating artifacts, installing dependencies,
          compiling software, configuring services, and running agent-authored
          validation checks. \\

          \bottomrule
      \end{tabular}
  \end{table*}

  For Terminal-Bench, each adapter or base retry begins in a fresh container to prevent filesystem changes from a previous attempt from leaking into the next one. Claw-Eval similarly relies on the official benchmark services and audit logs. GAIA retries retain access to the same task attachments and general-purpose tools. Thus, \method changes how the agent uses an existing interface, rather than
  changing the interface itself.

\subsection{\method Loop}
\label{app:pseudocode}
The \method loop has three main steps: 1) reflect on a failure to produce a textual lesson, 2) distill the lesson into a LoRA adapter, and 3) load the adapter into the model for a retry. The following Python pseudocode implements the high-level steps, including retrieval of existing lessons and the base retry fallback:
\clearpage

\begin{lstlisting}
  lesson_library = []
  SIM_THRESHOLD = tau
  LIBRARY_CAP = C

  def self_direction(base_model, task, failure_trace):
      lesson, similarity = retrieve_best(
          task, lesson_library
      )

      if lesson is None or similarity < SIM_THRESHOLD:
          lesson = None

          # Learn only while the bounded library has capacity.
          if len(lesson_library) < LIBRARY_CAP:
              reason = normalize_failure(failure_trace)
              lessons = reflect(
                  base_model,
                  task,
                  failure_trace,
                  reason,
                  REFLECTION_PROMPT,
              )
    
              adapter = train_lora(
                  freeze(base_model), lessons
              )
              lesson = register(
                  adapter,
                  retrieval_key=embed(task),
                  source_task=task,
              )
              lesson_library.append(lesson)

      # Retry using either a retrieved or newly learned lesson.
      success = False
      if lesson is not None:
          adapted_model = load_adapter(
              base_model, lesson.adapter
          )
          success, _ = agent(
              adapted_model, task
          )
          unload_adapter(adapted_model)

      # Fall back to the clean, fixed base model.
      if not success:
          success, _ = agent(
              base_model, task, RETRY_PROMPT
          )

      return success
  \end{lstlisting}

\clearpage

\section{Prompt Templates}
\label{app:prompts}

\method uses the same base model for task execution and failure reflection. Prompts are specialized to each benchmark's tools and success conditions but share the common structure shown next. We present generic versions for readability first; benchmark-specific prompts are presented in the following subsection~\ref{app:benchmark-specific-prompts}.

\subsection{Task-Execution System Prompt}

The execution prompt defines a ReAct-style interaction protocol. At each turn, the agent either emits one tool call and waits for its result or terminates with a final answer. Tool observations are appended to the trajectory before the next model invocation.

\begin{promptbox}{Task-execution system prompt}
  You are a careful agent working in [ENVIRONMENT].
  Reason step by step and use the provided tools when they help.

  Tool-use protocol:
  - Emit one tool call per turn using the required tool-call format,
    then stop generating.
  - Wait for the tool result before deciding the next action.
  - Inspect tool outputs and errors before proceeding.
  - Avoid repeating actions that do not advance the task.
  - Respect the available step and tool-call budgets.

  Completion protocol:
  - Verify that the requested answer or environment state has been
    produced before stopping.
  - When the task is complete, end with:
    Final Answer: [ANSWER OR DONE]
\end{promptbox}

GAIA specializes this template for short-answer problem solving. Terminal-Bench exposes a container shell and emphasizes verifying filesystem or program state. Claw-Eval supplies task-specific tools, whereas MiniWoB++ describes accessibility-tree observations and browser actions.

\subsection{Failure-Reflection Prompt}

After a verified failure, the base model acts as its own teacher. It receives the task, normalized failure reason, available tools, and a compact rendering of the failed trajectory. The prompt includes neither the correct answer nor any hidden tests.

The self-reflection stage produces structured instruction--response examples rather than an unconstrained textual critique. These examples teach the procedure that was missing from the failed execution and become the training data for the modular lesson adapter. These examples are not full model rollouts.

\begin{promptbox}{Failure-reflection prompt}
  You are a teacher model for continual agent learning.

  A student agent attempted a task using tools and failed. Generate
  training examples that teach the PROCEDURE needed to solve this class
  of task. Do not teach the student to memorize the tools or answers.

  NO ANSWER LEAKAGE:
  - Do not copy a task-specific answer, hidden solution, verifier rule,
    or final artifact into a response.
  - Teach how to derive or verify the answer instead.

  Each example should teach:
  - which tools to use and in what order;
  - what queries, commands, or inputs to provide;
  - how to interpret observations and tool results;
  - how to recover from ineffective actions;
  - when enough evidence has been collected to stop.

  Use the failed trajectory as evidence of the missing skill. Vary the
  examples across related situations so that the lesson describes a
  general procedure rather than one trajectory.

  Return only a JSON array with exactly [N] objects:
  [
    {
      "instruction": "...",
      "response": "...",
      "rationale": "..."
    }
  ]

  TASK (x):
  [TASK]

  FAILURE REASON (r):
  [FAILURE REASON]

  FAILURE-SPECIFIC FOCUS:
  [GUIDANCE FOR WRONG ANSWER, STEP BUDGET, TOOL BUDGET, OR MISSING FINAL RESPONSE]

  AVAILABLE TOOLS:
  [TOOL NAMES OR BENCHMARK TOOL DESCRIPTIONS]

  FAILED TRAJECTORY (t):
  [COMPACT SEQUENCE OF ACTIONS, OBSERVATIONS, ERRORS, AND FINAL OUTPUT]
\end{promptbox}

% The generic prompt is further constrained for environments with hidden evaluators. For example, Terminal-Bench reflection cannot access or reconstruct hidden tests, and browser-task reflection must describe elements by accessibility role and name rather than memorizing transient element identifiers.

\subsection{Benchmark-Specific Prompt Specializations}
  \label{app:benchmark-specific-prompts}

  The generic templates above are instantiated by fixed benchmark-specific
  system prompts. These prompts are shared across tasks and runs within each
  benchmark; only the benchmark task, current observation, attachment, and
  task-local tool schema vary. To make these differences explicit, we reproduce
  the principal benchmark-specific clauses below. The bracketed benchmark names
  are editorial labels; the prompt text itself is reproduced verbatim.

  \begin{promptbox}{Benchmark-specific execution clauses}
    [GAIA]

    You are a careful problem solver working on GAIA benchmark tasks.
    Reason step by step and use the provided tools when they help.

    Final answer protocol:
    - When you are ready, respond with concise reasoning followed by a
      single line beginning with 'Final Answer:' on the last line.
    - The answer should be in the simplest form the question asks for
      (a number, a short string, or a comma-separated list).

    [Terminal-Bench 2.1]

    You are completing a Terminal-Bench task inside an isolated Linux
    container.

    Use `terminal_shell' to inspect the environment, create or edit files, install or configure software when necessary, and verify your work. Commands start in the task image's working directory and filesystem changes persist throughout this attempt.

    - Check command exit codes and inspect relevant outputs or artifacts.
    - Do not look for benchmark tests, verifier files, reward files, or
      solutions. They are intentionally unavailable during the attempt.

    When the requested state has been produced and checked, end with
    "Final Answer: done".

    [Claw-Eval]

    You are running Claw-Eval tasks through the official Claw-Eval harness.
    Use the provided benchmark tools directly and complete the user's
    request. The benchmark grades the full trace, service audit logs,
    generated artifacts, and final response, not just a short final answer.

    - Tool names are case-sensitive. Use exactly the names shown in the
      tool list.
    - Do not claim that you performed an action unless you actually used
      the relevant tool or verified the artifact/state.

    Stopping:
    - When the task is complete, provide the requested result clearly. If
      the task asks for a report, classification, draft, or plan, put that
      content in your final assistant message.

    [MiniWoB++]

    You are controlling a browser for a MiniWoB++ task. Reason step by step.

    Each observation contains:
      goal: <what to accomplish>
      url:, reward:, terminated:, truncated:[, last\_action\_error:]
      page: an accessibility tree where every interactive element has a bid.

    - Use bids that appear in the current `page:` tree; never invent bids.
      If `last_action_error:` is set, do not repeat the same call---choose
      a different bid or re-observe. Avoid repeated scroll/wait loops.

    Stopping:
    - When an observation shows `terminated: True` with `reward >= 0.5`,
      respond with a single line: `Final Answer: done`. Do not invent any
      other final answer.

  \end{promptbox}

The reflection prompt is specialized in the same way. GAIA uses the generic failure-reflection system prompt shown above without an additional benchmark-specific override. The other benchmarks replace it with prompts that encode their observable state and explicitly exclude information unavailable to the agent as shown below.

  \begin{promptbox}{Benchmark-specific reflection constraints}
    [Terminal-Bench 2.1]

    Use only the task instruction and the agent's terminal trajectory 
    as evidence. The hidden verifier, benchmark tests, oracle solution, and reward implementation are unavailable and must not be guessed or reconstructed.

    Vary examples across related tasks. Do not copy task-specific answers, secrets, exact final artifacts, hidden checks, or an oracle command sequence.

    [Claw-Eval]

    Use the failed trajectory as evidence of which agent skill was missing.

    CRITICAL RULES:
    - Do not copy a task's reference solution or grader rubric into the response.
    - Do not hardcode specific fixture answers unless the failed trajectory itself shows how to derive them from tools/files.

    [MiniWoB++]

    Use the trajectory as evidence of which browser skills the agent lacked or misapplied rather than guessing bids.

    CRITICAL RULES:
    - Do not copy specific numeric bids, autocomplete picks, or other seed-specific values from the trajectory into the response. Bids change across seeds; teach the agent how to FIND the right element by role + name, not which integer to type.

  \end{promptbox}

\subsection{Adapter Retry and Base Fallback}

If the adapter-guided retry is unsuccessful, a final retry is attempted with the base agent, with a short instruction appended to the original prompt:

\begin{promptbox}{Retry instruction appended for the base fallback}
  You already tried this task once and produced an incorrect answer.
  Please try again.
\end{promptbox}

For Claw-Eval, the final sentence also asks the agent to verify the result before stopping. During the adapter-guided retry, the selected modular adapter is active. If that attempt fails, the adapter is fully unloaded and the same task and retry suffix are presented to the base model. No reflection examples or adapter outputs are inserted into the fallback context. This isolates the fallback as an ordinary base-model retry while preserving the same failure-conditioned instruction.

\clearpage

\section{Qualitative Analysis of Forward Transfer}
\label{app:qualitative-transfer}

\begin{figure*}[h]
    \centering
    \includegraphics[width=\textwidth]{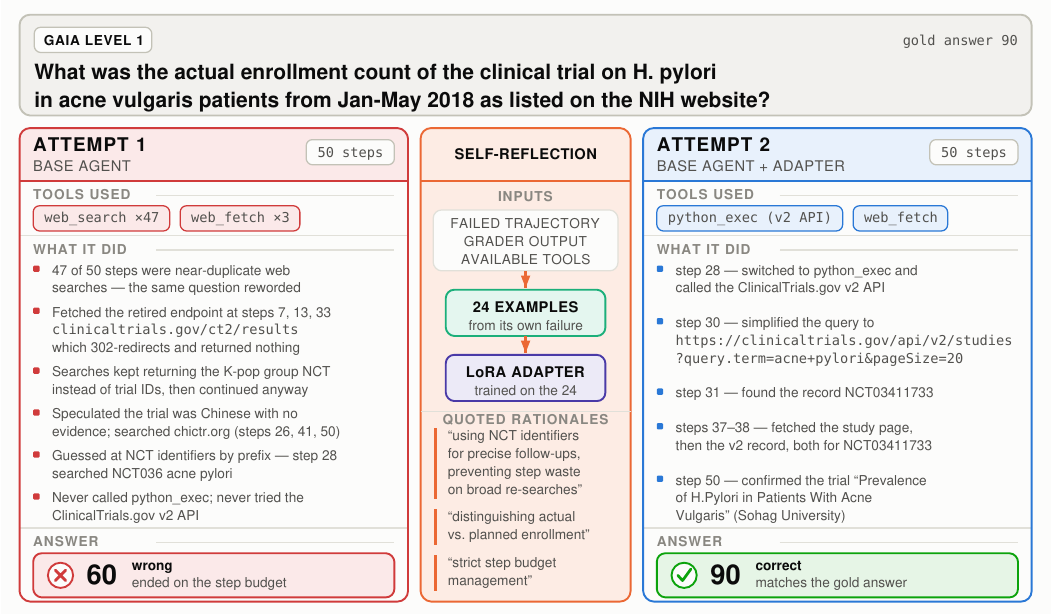}
    \caption{\textbf{\method recovers failing tasks.}
    A qualitative example showing how \method reflects on the failure, correctly identifies overuse of invalid tools, and generates corrective examples that ultimately help the agent succeed on a GAIA Level-1 task.}
    \label{fig:qual-gaia-example}
\end{figure*}

The quantitative transfer results show that stored lesson adapters can improve later tasks. We now examine what these adapters appear to transfer qualitatively. Across GAIA and Claw-Eval, the strongest examples share a procedural structure -- such as pagination, cross-system reconciliation, or exact-value verification -- even when their topics, difficulty levels, or task categories differ. Fig.~\ref{fig:qual-gaia-example} shows recovery with \method on a Level~1 task in GAIA. Examples of generated reflections are shown in the following section (Appendix~\ref{app:reflection-examples}).

\subsection{Transfer Within an Online GAIA Stream}
\label{app:gaia-transfer-analysis}

The GAIA deployment experiment in Section~\ref{sec:results-transfer} interleaves all 165 tasks across difficulty levels and enables the agent to retrieve only adapters learned earlier in the stream. Across the three routing thresholds, existing adapters are selected for 282 later failures and produce 41 successful adapter-conditioned retries. Thirty of these successes cross GAIA difficulty levels. Same-level and cross-level retrievals have nearly
identical success rates, $14.7\%$ and $14.5\%$, suggesting that benchmark difficulty is not the natural boundary of a reusable lesson.

Table~\ref{tab:gaia-qualitative-transfer} presents representative transfers. In each case, the source and target differ in surface content but require a related sequence of actions or verification strategy.

\begin{table*}[h]
      \centering
      \caption{\textbf{Representative transfers in the shuffled GAIA stream.}
      Every source task occurs before its target in the stream. The examples
      illustrate procedural reuse across topics, modalities, and difficulty
      levels.}
      \label{tab:gaia-qualitative-transfer}
      \small
      \renewcommand{\arraystretch}{1.15}
      \begin{tabularx}{\textwidth}{@{}p{0.19\textwidth}p{0.20\textwidth}Xp{0.20\textwidth}@{}}
          \toprule
          Source adapter & Later target & Shared procedure & Observed recovery \\
          \midrule
          ORCID publication aggregation
          (Level~3)
          &
          Wikipedia revision counting
          (Level~2)
          &
          Traverse a complete remote history, handle pagination, apply a date
          constraint, and verify completeness before counting.
          &
          Initial answer: 2610.
          Adapter answer: 2732 (correct).
          \\

          BBC Earth video lookup
          (Level~1)
          &
          Exact quotation from another video
          (Level~1)
          &
          Identify the correct video and extract a localized fact using
          descriptions, timestamps, comments, or other supporting metadata.
          &
          The initial execution exhausted its step budget; the adapter returned
          the exact quotation.
          \\

          Spreadsheet aggregation
          (Level~1)
          &
          Statistics over values extracted from an image
          (Level~2)
          &
          Extract structured values, inspect intermediate data, compute with a
          tool, verify the calculation, and format only after validation.
          &
          Initial answer: 2.404.
          Adapter answer: 17.056 (correct).
          \\

          Caesar-cipher decoding
          (Level~2)
          &
          Text-grid reconstruction
          (Level~1)
          &
          Apply a text transformation, check linguistic coherence, and preserve
          exact capitalization, spacing, and output constraints.
          &
          The initial execution recovered the meaning but violated formatting;
          the adapter returned the exact required sentence.
          \\
          \bottomrule
      \end{tabularx}
\end{table*}

These examples suggest that the adapters can encode procedures rather than task-specific guidance. The spreadsheet adapter, for example, transfers from tabular sales data to numerical values embedded in an image. The shared behavior is not the input modality but the discipline of extracting, checking, computing, and formatting. Similarly, the ORCID adapter transfers from publication records to Wikipedia revisions because both tasks require complete pagination and verified counting.

An adapter can also become useful even when it does not repair its source task. The first ORCID adapter fails its own adapter-guided retry, yet later produces 14 successful retries in the most permissive routing condition. Success on the source evaluation is, therefore, not a complete measure of whether the learned procedure will be useful elsewhere.

% \paragraph{Routing remains imperfect.}
% Task-text similarity does not reliably identify procedural compatibility. In the most permissive condition, the ORCID adapter is selected for 75 of 102 reuse attempts, creating an early routing hub. One high-similarity failure retrieves this adapter for an OpenReview author-counting task, but neither the base execution nor the adapter produces the correct result. More generally, similarity distributions for
% successful and unsuccessful retrievals overlap substantially.

% These failures motivate routers that consider required tools, input modality, failure reason, learned skill descriptions, and prior adapter outcomes rather than raw task-text similarity alone. The current similarity threshold is best understood as controlling library growth and reuse frequency, not as a calibrated estimate that an adapter will succeed.

\subsection{Cross-Category Transfer on Claw-Eval}
\label{app:claw-transfer-analysis}

The Claw-Eval experiment in Section~\ref{sec:results-transfer} provides a cleaner category shift. We freeze a library of three adapters learned from English operations (ops) tasks and evaluate it on 24 unseen workflow tasks in the original benchmark. The base model solves $18/24$ tasks. On the six initially encountered failures, retrieved operations adapters solve three, increasing final accuracy to $21/24$ ($87.5\%$).

The successful transfers involve related coordination procedures despite the source and target belonging to different benchmark categories. Table~\ref{tab:claw-qualitative-transfer} summarizes the three recovered workflows.

\begin{table*}[h]
      \centering
      \caption{\textbf{Qualitative cross-category transfer from Claw-Eval
      operations tasks to unseen workflows.} Scores are assigned by the
      Claw-Eval grader. No workflow adapter is trained before evaluation.}
      \label{tab:claw-qualitative-transfer}
      \small
      \renewcommand{\arraystretch}{1.15}
      \begin{tabularx}{\textwidth}{@{}p{0.20\textwidth}p{0.21\textwidth}Xp{0.15\textwidth}@{}}
          \toprule
          Source operations lesson & Target workflow & Shared procedure & Grader score change \\
          \midrule
          Scheduled-task management
          &
          Weekly meeting tracking
          &
          Collect status across multiple records, identify missing or incomplete
          actions, and produce a structured progress report.
          &
          $0.413 \rightarrow 0.946$
          \\

          Month-end reconciliation
          &
          Task backtracking
          &
          Reconcile evidence across emails, task lists, and meeting notes;
          identify discrepancies and report status, blockers, and next actions.
          &
          $0.674 \rightarrow 0.894$
          \\

          Scheduled-task management
          &
          Multi-project progress reporting
          &
          Aggregate task status across several projects, identify risks, and
          organize the results into separate actionable reports.
          &
          $0.586 \rightarrow 0.890$
          \\
          \bottomrule
      \end{tabularx}
\end{table*}

The scheduled-task adapter transfers to two workflows because all three tasks require auditing distributed state, identifying incomplete work, and producing a structured report. The reconciliation adapter similarly transfers beyond finance: Its useful behavior is gathering corresponding records from several systems, resolving inconsistencies, and summarizing the resulting status.

\paragraph{Interpretation.}
Together, these examples support the view that modular lesson adapters can preserve helpful procedural lessons and make them available in new contexts. The strongest cases are those in which the source and target share an identifiable workflow, even when their topics, modalities, difficulty levels, or benchmark categories differ.
\clearpage

\section{Qualitative Examples of Failure-Conditioned \method}
\label{app:reflection-examples}

The examples generated during reflection do not simply reproduce the failed task, nor are they full, expensive rollouts. Instead, they express a reusable corrective procedure that addresses the underlying failure across varied situations. Next, we trace one task from each primary benchmark for which the base execution failed and the adapter-guided retry succeeded. Excerpts are shortened for readability but are drawn directly from the saved run artifacts.

\textbf{GAIA: Structured extraction from audio.}

\emph{Task.} The agent was asked to transcribe an attached strawberry-pie recipe and return only the filling ingredients as an alphabetized, comma-separated list without measurements.

\emph{Failure.} The base agent returned generic ingredient names such as ``lemon juice'' and ``vanilla extract,'' omitting the required qualifiers ``freshly squeezed'' and ``pure,'' and introduced extraneous formatting. The answer, therefore, did not match the required list and was recorded as incorrect.

\emph{Generated lesson.} Self-reflection produced examples covering a general audio-to-structured-data procedure: Transcribe the recording, isolate the requested section, exclude irrelevant categories, remove quantities, normalize and sort the extracted items, and verify the required output format. One example asks the agent to extract only the dry ingredients from a baking recording while excluding measurements and returning an alphabetized list.

\emph{Outcome.} The resulting adapter returned the exact expected ingredient list and recovered the task.

\textbf{Claw-Eval: Completing evidence-grounded research.}

\emph{Task.} The agent was asked to evaluate a migration from Redis to Valkey, covering the Redis license change, Valkey's governance, technical compatibility, cloud support, migration experiences, and project health.

\emph{Failure.} The base execution spent its execution repeatedly searching and fetching sources but ended without synthesizing the requested report.

\emph{Generated lesson.} The reflective examples taught a more robust research workflow: Broaden queries when search returns no results, pivot to authoritative URLs when retrieval fails, cross-check technical claims, map collected evidence to every requested section, and reserve sufficient time for synthesis. For example, one generated instruction asks what to do when an initial technical search returns no results; its response recommends reformulating the query, trying authoritative sources directly, and avoiding unsupported conclusions.

\emph{Outcome.} The adapter-guided retry completed the requested technical evaluation and passed the Claw-Eval verifier.

\textbf{Terminal-Bench 2.1: Verifying a multi-service deployment.}

\emph{Task.} The agent had to configure a Git server over SSH, deploy the \texttt{main} and \texttt{dev} branches through a post-receive hook, and serve both through distinct HTTPS endpoints.

\emph{Failure.} The base execution exhausted its step budget without establishing a working end-to-end deployment. The subsequent test encountered an invalid repository state---including ``remote HEAD refers to nonexistent ref''---and failed.

\emph{Generated lesson.} Self-reflection produced examples emphasizing observable verification rather than assumed success: Inspect effective SSH and Nginx configurations, confirm canonical ports, validate the bare repository with \texttt{git ls-remote}, instrument the deployment hook, and reproduce the complete clone-commit-push-fetch workflow from a clean directory. One example explicitly instructs the agent to reproduce the complete end-to-end workflow and verify the returned page contents against the task requirements before declaring completion.

\emph{Outcome.} The adapter-guided retry correctly deployed both branches and the complete evaluator test passed.

Together, these cases illustrate that \method extracts different forms of procedural knowledge from different failures: precise information extraction on GAIA, evidence-grounded task completion on Claw-Eval, and systematic end-to-end verification on Terminal-Bench 2.1.

\clearpage

% \section{Why Adapter Skills Transfer: A Shared Repair Cone}
% \label{app:repair-cone}

% The ability of independently learned adapter skills to recover related failures raises a natural question: \emph{Why do task-specific repairs transfer at all?} To investigate this behavior, we compared the parameter updates learned across GAIA, Claw-Eval, and Terminal-Bench 2.1. We find that the adapters are neither identical nor scattered arbitrarily through parameter space. Instead, they occupy a partially shared region that we refer to as a \textbf{repair cone}: a common direction of behavioral correction accompanied by task-specific variation.

\section{Benefits of SVD Initialization}
\label{app:svd_lora}
Initializing matrix $A$ with SVD and keeping it frozen offers two benefits: It further reduces the number of trainable parameters in an already lightweight adapter and anchors adaptation in a high-energy subspace of the pretrained model \citep{ren2026scaling}. Initializing matrix $B$ to zero makes the adapter an exact no-op before training. Freezing the base model confines learning to a small, failure-specific set of adapter parameters. Only the adapter's $B$ matrices are trained, adding trainable parameters equivalent to at most approximately $0.02\%$ of the base model's parameter count. The failure type also determines where the repair is applied: Planning and budget failures target attention projections, whereas incorrect-answer failures target MLP projections.

In our final configuration, $A_\ell$ is initialized from a rank-$r$ high-energy right-singular subspace of the frozen base weight, $B_{i,\ell}$ is initialized to zero, and $A_\ell$ remains frozen. Only $B_{i,\ell}$ is optimized:
\[
  \mathcal{D}_i = \{(u_j,v_j)\}_{j=1}^{N_i},
  \qquad
  \phi_i^\star
  =
  \arg\min_{\phi_i}
  \frac{1}{N_i}
  \sum_{j=1}^{N_i}
  [-\log p_{\theta,\phi_i}(v_j \mid u_j)],
  \qquad \theta\ \text{fixed}.
  \]

Here, $\phi_i$ denotes the trainable parameters of the new adapter. For each of the $N_i$ corrective examples, $u_j$ denotes a training instruction and $v_j$ its desired procedural response. $p_{\theta,\phi_i}(v_j\mid u_j)$ is the probability the adapted model assigns to response $v_j$ given instruction $u_j$.

\paragraph{SVD provides a shared coordinate system for repair.}
For a base weight matrix \(W\), lesson adapter \(i\) induces the update
\[
W_i = W + \Delta W_i,
\qquad
\Delta W_i = B_i A,
\]
where \(A\) is initialized from a rank-16 slice of the base weight's high-energy spectral band and remains fixed during training. Only \(B_i\) is learned from the reflective examples for failure \(i\). The selected subspaces capture approximately \(16\)--\(22\) times as much base-weight energy as random subspaces of the same dimension. Thus, different lesson adapters are trained within a compact set of input features that are already prominent in the pretrained model.

Importantly, the learned update does not simply amplify the corresponding base weights. Although its input directions are inherited from the base model, its output directions are largely new. One interpretation is that the adapter recognizes familiar pretrained features but learns to route them toward behavior that would have prevented the observed failure.

\end{document}

%% file: math_commands.tex
\usepackage{amsmath,amsfonts,bm}

\def\eqref#1{equation~\ref{#1}}
\def\1{\bm{1}}

\DeclareMathAlphabet{\mathsfit}{\encodingdefault}{\sfdefault}{m}{sl}
\SetMathAlphabet{\mathsfit}{bold}{\encodingdefault}{\sfdefault}{bx}{n}

%% file: fair_template.bbl
\begin{thebibliography}{63}
\providecommand{\natexlab}[1]{#1}
\providecommand{\url}[1]{\texttt{#1}}
\expandafter\ifx\csname urlstyle\endcsname\relax
  \providecommand{\doi}[1]{doi: #1}\else
  \providecommand{\doi}{doi: \begingroup \urlstyle{rm}\Url}\fi

\bibitem[Agrawal et~al.(2026)Agrawal, Tan, Soylu, Ziems, Khare, Opsahl-Ong, Singhvi, Shandilya, Ryan, Jiang, Potts, Sen, Dimakis, Stoica, Klein, Zaharia, and Khattab]{agrawal2026gepa}
Lakshya~A. Agrawal, Shangyin Tan, Dilara Soylu, Noah Ziems, Rishi Khare, Krista Opsahl-Ong, Arnav Singhvi, Herumb Shandilya, Michael~J. Ryan, Meng Jiang, Christopher Potts, Koushik Sen, Alexandros~G. Dimakis, Ion Stoica, Dan Klein, Matei Zaharia, and Omar Khattab.
\newblock {GEPA}: Reflective prompt evolution can outperform reinforcement learning.
\newblock In \emph{Proceedings of the International Conference on Learning Representations}, 2026.
\newblock Oral. arXiv:2507.19457.

\bibitem[Aky{\"u}rek et~al.(2025)Aky{\"u}rek, Damani, Zweiger, Qiu, Guo, Pari, Kim, and Andreas]{akyurek2025surprising}
Ekin Aky{\"u}rek, Mehul Damani, Adam Zweiger, Linlu Qiu, Han Guo, Jyothish Pari, Yoon Kim, and Jacob Andreas.
\newblock The surprising effectiveness of test-time training for few-shot learning.
\newblock In \emph{Proceedings of the International Conference on Machine Learning}, volume 267 of \emph{Proceedings of Machine Learning Research}, pages 942--963. PMLR, 2025.

\bibitem[An et~al.(2024)An, Ma, Lin, Zheng, Lou, and Chen]{LearningMistakesMakes2024}
Shengnan An, Zexiong Ma, Zeqi Lin, Nanning Zheng, Jian-Guang Lou, and Weizhu Chen.
\newblock Learning from mistakes makes {LLM} better reasoner, March 2024.
\newblock arXiv:2310.20689 [cs.CL].

\bibitem[Anthropic(2026)]{Claude_skill_2026}
Anthropic, Jun 2026.
\newblock \url{https://claude.com/blog/complete-guide-to-building-skills-for-claude}.

\bibitem[Back et~al.(2026)Back, Lee, Kang, Lee, Hong, Gwon, and Ahn]{backUnderstandingLoRAKnowledg2026}
Seungju Back, Dongwoo Lee, Naun Kang, Taehee Lee, S.~K. Hong, Youngjune Gwon, and Sungjin Ahn.
\newblock Understanding {LoRA} as knowledge memory: An empirical analysis.
\newblock In \emph{Proceedings of the International Conference on Machine Learning}, June 2026.

\bibitem[Bandi et~al.(2026)Bandi, Dumitru, Hertzberg, Agarwal, Boo, Polakam, Hassaan, Da, Kim, Gupta, et~al.]{bandi2602mcp}
Chaithanya Bandi, Razvan-Gabriel Dumitru, Ben Hertzberg, Divyansh Agarwal, Geobio Boo, Tejas Polakam, Sami Hassaan, Jeff Da, HiJae Kim, Vipul Gupta, et~al.
\newblock {MCP-Atlas}: A large-scale benchmark for tool-use competency with real {MCP} servers, 2026.
\newblock arXiv:2602.00933 [cs.SE].

\bibitem[Brown et~al.(2024)Brown, Juravsky, Ehrlich, Clark, Le, R{\'e}, and Mirhoseini]{brown2024large}
Bradley Brown, Jordan Juravsky, Ryan Ehrlich, Ronald Clark, Quoc~V. Le, Christopher R{\'e}, and Azalia Mirhoseini.
\newblock Large language monkeys: Scaling inference compute with repeated sampling.
\newblock \emph{arXiv preprint arXiv:2407.21787}, 2024.

\bibitem[Chen et~al.(2026)Chen, Lu, Zhao, Meng, Teng, Li, Li, Liu, Liang, Zhang, et~al.]{chen2026harnessx}
Tingyang Chen, Shuo Lu, Kang Zhao, Weicheng Meng, Hanlin Teng, Tianhao Li, Chao Li, Xule Liu, Jian Liang, Zhizhong Zhang, et~al.
\newblock {HarnessX}: A composable, adaptive, and evolvable agent harness foundry.
\newblock \emph{arXiv preprint arXiv:2606.14249}, 2026.

\bibitem[Dong et~al.(2026)Dong, Gao, Li, Xu, Hua, and Yang]{dong2026agent}
Gen Dong, Yanjie Gao, Liqun Li, Tianyin Xu, Yu~Hua, and Fan Yang.
\newblock Agent skills can be harmful: An empirical study of skill-induced failures in {LLM} agents.
\newblock \emph{arXiv preprint arXiv:2608.11888}, 2026.

\bibitem[Gonzalez-Pumariega et~al.(2026)Gonzalez-Pumariega, Agashe, Yang, Li, and Wang]{gonzalez2026reliability}
Gonzalo Gonzalez-Pumariega, Saaket Agashe, Jiachen Yang, Ang Li, and Xin~Eric Wang.
\newblock On the reliability of computer use agents.
\newblock \emph{arXiv preprint arXiv:2604.17849}, 2026.

\bibitem[Gu et~al.(2024)Gu, Xu, Ma, Lu, Ling, Chang, and Peng]{gModelEditingHarms2024}
Jia-Chen Gu, Hao-Xiang Xu, Jun-Yu Ma, Pan Lu, Zhen-Hua Ling, Kai-Wei Chang, and Nanyun Peng.
\newblock Model editing harms general abilities of large language models: Regularization to the rescue, October 2024.
\newblock arXiv:2401.04700 [cs.CL].

\bibitem[Gulcehre et~al.(2023)Gulcehre, Paine, Srinivasan, Konyushkova, Weerts, Sharma, Siddhant, Ahern, Wang, Gu, Macherey, Doucet, Firat, and de~Freitas]{gulcehre2023reinforced}
Caglar Gulcehre, Tom~Le Paine, Srivatsan Srinivasan, Ksenia Konyushkova, Lotte Weerts, Abhishek Sharma, Aditya Siddhant, Alex Ahern, Miaosen Wang, Chenjie Gu, Wolfgang Macherey, Arnaud Doucet, Orhan Firat, and Nando de~Freitas.
\newblock Reinforced self-training ({ReST}) for language modeling.
\newblock \emph{arXiv preprint arXiv: 2308.08998}, 2023.

\bibitem[Gupta et~al.(2026)Gupta, Isola, Jegelka, Lopez-Paz, Ahuja, Ibrahim, and Pezeshki]{gupta2026reasoncache}
Sharut Gupta, Phillip Isola, Stefanie Jegelka, David Lopez-Paz, Kartik Ahuja, Mark Ibrahim, and Mohammad Pezeshki.
\newblock {ReasonCache}: Teaching {LLMs} to reason without weight updates.
\newblock \emph{arXiv preprint arXiv:2602.02366}, 2026.

\bibitem[Hardt and Sun(2024)]{hardt2024testtime}
Moritz Hardt and Yu~Sun.
\newblock Test-time training on nearest neighbors for large language models.
\newblock In \emph{Proceedings of the International Conference on Learning Representations}, 2024.

\bibitem[Hu et~al.(2022)Hu, Shen, Wallis, Allen-Zhu, Li, Wang, Wang, and Chen]{uLoRALowRankAdaptation2021}
Edward~J. Hu, Yelong Shen, Phillip Wallis, Zeyuan Allen-Zhu, Yuanzhi Li, Shean Wang, Lu~Wang, and Weizhu Chen.
\newblock {LoRA}: Low-rank adaptation of large language models.
\newblock In \emph{Proceedings of the International Conference on Learning Representations}, October 2022.

\bibitem[Hu et~al.(2025)Hu, Lu, and Clune]{hu2025adas}
Shengran Hu, Cong Lu, and Jeff Clune.
\newblock Automated design of agentic systems.
\newblock In \emph{Proceedings of the International Conference on Learning Representations}, 2025.

\bibitem[Jiang et~al.(2026)Jiang, Shi, Hong, Xu, Sun, Sun, Bashir, and Han]{jiang2026harness1}
Pengcheng Jiang, Zhiyi Shi, Kelly Hong, Xueqiang Xu, Jiashuo Sun, Jimeng Sun, Hammad Bashir, and Jiawei Han.
\newblock Harness-1: Reinforcement learning for search agents with state-externalizing harnesses.
\newblock \emph{arXiv preprint arXiv:2606.02373}, 2026.

\bibitem[Karten et~al.(2026)Karten, Zhang, Thomas, M\"{u}ller, and {Prime Intellect Team}]{primeintellect2026primeagent}
Seth Karten, Alex~L. Zhang, Kevin Thomas, Sebastian M\"{u}ller, and {Prime Intellect Team}.
\newblock Prime agent: A self-improving {RLM} harness.
\newblock Prime Intellect Blog, August 2026.
\newblock \url{https://www.primeintellect.ai/blog/prime-agent}.

\bibitem[Lee et~al.(2026)Lee, Nair, Zhang, Lee, Khattab, and Finn]{lee2026metaharness}
Yoonho Lee, Roshen Nair, Qizheng Zhang, Kangwook Lee, Omar Khattab, and Chelsea Finn.
\newblock Meta-harness: End-to-end optimization of model harnesses.
\newblock \emph{arXiv preprint arXiv:2603.28052}, 2026.

\bibitem[Lin et~al.(2026)Lin, Liu, Pan, Lin, Dou, Xi, Huang, Yan, Han, Gui, et~al.]{lin2026ahe}
Jiahang Lin, Shichun Liu, Chengjun Pan, Lizhi Lin, Shihan Dou, Zhiheng Xi, Xuanjing Huang, Hang Yan, Zhenhua Han, Tao Gui, et~al.
\newblock Agentic harness engineering: Observability-driven automatic evolution of coding-agent harnesses.
\newblock \emph{arXiv preprint arXiv:2604.25850}, 2026.

\bibitem[Liu et~al.(2023)Liu, Lin, Hewitt, Paranjape, Bevilacqua, Petroni, and Liang]{iuLostMiddleHow2023}
Nelson~F. Liu, Kevin Lin, John Hewitt, Ashwin Paranjape, Michele Bevilacqua, Fabio Petroni, and Percy Liang.
\newblock Lost in the middle: How language models use long contexts, November 2023.
\newblock arXiv:2307.03172 [cs.CL].

\bibitem[Lou et~al.(2026)Lou, L{\'a}zaro-Gredilla, Dedieu, Wendelken, Lehrach, and Murphy]{lou2026autoharness}
Xinghua Lou, Miguel L{\'a}zaro-Gredilla, Antoine Dedieu, Carter Wendelken, Wolfgang Lehrach, and Kevin~P. Murphy.
\newblock {AutoHarness}: Improving {LLM} agents by automatically synthesizing a code harness.
\newblock \emph{arXiv preprint arXiv:2603.03329}, 2026.

\bibitem[Lu et~al.(2022)Lu, Bartolo, Moore, Riedel, and Stenetorp]{uFantasticallyOrderedPrompts2022}
Yao Lu, Max Bartolo, Alastair Moore, Sebastian Riedel, and Pontus Stenetorp.
\newblock Fantastically ordered prompts and where to find them: Overcoming few-shot prompt order sensitivity, March 2022.
\newblock arXiv:2104.08786 [cs.CL].

\bibitem[Lu et~al.(2026)Lu, Yao, Wu, Han, Gu, Cai, Lu, Xiao, Zhuang, and Shen]{uSKILL0InContextAgentic2026}
Zhengxi Lu, Zhiyuan Yao, Jinyang Wu, Chengcheng Han, Qi~Gu, Xunliang Cai, Weiming Lu, Jun Xiao, Yueting Zhuang, and Yongliang Shen.
\newblock {SKILL0}: In-context agentic reinforcement learning for skill internalization, April 2026.
\newblock arXiv:2604.02268 [cs.LG].

\bibitem[Luo et~al.(2023)Luo, Yang, Meng, Li, Zhou, and Zhang]{uoEmpiricalStudyCatastrophic2025}
Yun Luo, Zhen Yang, Fandong Meng, Yafu Li, Jie Zhou, and Yue Zhang.
\newblock An empirical study of catastrophic forgetting in large language models during continual fine-tuning, January 2023.
\newblock arXiv:2308.08747 [cs.CL].

\bibitem[Lv et~al.(2026)Lv, E, Cai, Huang, Zhang, Ding, Wu, Wang, Zhang, Wang, and Duan]{lvFailForgeDistillingProcedural2026}
Dongyi Lv, Fushun E, Aichen Cai, Liang Huang, Ya~Zhang, Qiuyu Ding, Canhui Wu, Zhi Wang, Yuesong Zhang, Jiaqi Wang, and Nan Duan.
\newblock {FailForge}: Distilling procedural competence from persistent failures into code agents, August 2026.
\newblock arXiv:2608.08570 [cs.AI].

\bibitem[Madaan et~al.(2023)Madaan, Tandon, Gupta, Hallinan, Gao, Wiegreffe, Alon, Dziri, Prabhumoye, Yang, Gupta, Majumder, Hermann, Welleck, Yazdanbakhsh, and Clark]{madaanSelfRefineIterativeRefinement2023}
Aman Madaan, Niket Tandon, Prakhar Gupta, Skyler Hallinan, Luyu Gao, Sarah Wiegreffe, Uri Alon, Nouha Dziri, Shrimai Prabhumoye, Yiming Yang, Shashank Gupta, Bodhisattwa~Prasad Majumder, Katherine Hermann, Sean Welleck, Amir Yazdanbakhsh, and Peter Clark.
\newblock Self-{Refine}: Iterative refinement with self-feedback.
\newblock In \emph{Advances in Neural Information Processing Systems}, May 2023.

\bibitem[Majumder et~al.(2024)Majumder, Dalvi~Mishra, Jansen, Tafjord, Tandon, Zhang, Callison-Burch, and Clark]{majumder2023clin}
Bodhisattwa~Prasad Majumder, Bhavana Dalvi~Mishra, Peter Jansen, Oyvind Tafjord, Niket Tandon, Li~Zhang, Chris Callison-Burch, and Peter Clark.
\newblock {CLIN}: A continually learning language agent for rapid task adaptation and generalization.
\newblock In \emph{Proceedings of the Conference on Language Modeling}, 2024.

\bibitem[Meng et~al.(2024)Meng, Wang, and Zhang]{mengPiSSAPrincipalSingular2025}
Fanxu Meng, Zhaohui Wang, and Muhan Zhang.
\newblock {PiSSA}: Principal singular values and singular vectors adaptation of large language models.
\newblock In \emph{Proceedings of Advances in Neural Information Processing Systems}, April 2024.

\bibitem[Merrill et~al.(2026)Merrill, Shaw, Carlini, Li, Raj, Bercovich, Shi, Shin, Walshe, Buchanan, Shen, Ye, Lin, Poulos, Wang, Nezhurina, Jitsev, Lu, Mastromichalakis, Xu, Chen, Liu, Zhang, Chen, Kashyap, Uslu, Li, Wu, Yan, Bian, Sharma, Sun, Dillmann, Anand, Lanpouthakoun, Koopah, Hu, Guha, Dreiman, Zhu, Krauth, Zhong, Muennighoff, Amanfu, Tan, Pimpalgaonkar, Aggarwal, Lin, Lan, Zhao, Liang, Wang, Wang, Zhou, Heineman, Liu, Trivedi, Yang, Lin, Shetty, Yang, Omi, Raoof, Li, Zhuo, Lin, Dai, Wang, Chai, Zhou, Wahdany, She, Hu, Dong, Zhu, Cui, Saiyed, Kolbeinsson, Hu, Rytting, Marten, Wang, Dimakis, Konwinski, and Schmidt]{merrillTerminalBenchBenchmarkingAgents2026}
Mike~A. Merrill, Alexander~G. Shaw, Nicholas Carlini, Boxuan Li, Harsh Raj, Ivan Bercovich, Lin Shi, Jeong~Yeon Shin, Thomas Walshe, E.~Kelly Buchanan, Junhong Shen, Guanghao Ye, Haowei Lin, Jason Poulos, Maoyu Wang, Marianna Nezhurina, Jenia Jitsev, Di~Lu, Orfeas~Menis Mastromichalakis, Zhiwei Xu, Zizhao Chen, Yue Liu, Robert Zhang, Leon~Liangyu Chen, Anurag Kashyap, Jan-Lucas Uslu, Jeffrey Li, Jianbo Wu, Minghao Yan, Song Bian, Vedang Sharma, Ke~Sun, Steven Dillmann, Akshay Anand, Andrew Lanpouthakoun, Bardia Koopah, Changran Hu, Etash Guha, Gabriel H.~S. Dreiman, Jiacheng Zhu, Karl Krauth, Li~Zhong, Niklas Muennighoff, Robert Amanfu, Shangyin Tan, Shreyas Pimpalgaonkar, Tushar Aggarwal, Xiangning Lin, Xin Lan, Xuandong Zhao, Yiqing Liang, Yuanli Wang, Zilong Wang, Changzhi Zhou, David Heineman, Hange Liu, Harsh Trivedi, John Yang, Junhong Lin, Manish Shetty, Michael Yang, Nabil Omi, Negin Raoof, Shanda Li, Terry~Yue Zhuo, Wuwei Lin, Yiwei Dai, Yuxin Wang, Wenhao Chai, Shang Zhou, Dariush Wahdany, Ziyu She,
  Jiaming Hu, Zhikang Dong, Yuxuan Zhu, Sasha Cui, Ahson Saiyed, Arinbjörn Kolbeinsson, Jesse Hu, Christopher~Michael Rytting, Ryan Marten, Yixin Wang, Alex Dimakis, Andy Konwinski, and Ludwig Schmidt.
\newblock Terminal-{Bench}: Benchmarking agents on hard, realistic tasks in command line interfaces, January 2026.
\newblock arXiv:2601.11868 [cs.SE].

\bibitem[Mi et~al.(2026)Mi, Ma, Yang, Li, Wang, Zhang, and Wang]{miSkillProLearningReusable2026}
Qirui Mi, Zhijian Ma, Mengyue Yang, Haoxuan Li, Yisen Wang, Haifeng Zhang, and Jun Wang.
\newblock Skill-{Pro}: Learning reusable skills from experience via non-parametric {PPO} for {LLM} {Agents}.
\newblock In \emph{Proceedings of International Conference on Machine Learning}, February 2026.

\bibitem[Mialon et~al.(2023)Mialon, Fourrier, Swift, Wolf, LeCun, and Scialom]{mialonGAIABenchmarkGeneral2023}
Grégoire Mialon, Clémentine Fourrier, Craig Swift, Thomas Wolf, Yann LeCun, and Thomas Scialom.
\newblock {GAIA}: A benchmark for general {AI} assistants, November 2023.
\newblock arXiv:2311.12983 [cs].

\bibitem[Muennighoff et~al.(2025)Muennighoff, Yang, Shi, Li, Fei-Fei, Hajishirzi, Zettlemoyer, Liang, Candès, and Hashimoto]{muennighoffS1SimpleTesttime2025}
Niklas Muennighoff, Zitong Yang, Weijia Shi, Xiang~Lisa Li, Li~Fei-Fei, Hannaneh Hajishirzi, Luke Zettlemoyer, Percy Liang, Emmanuel Candès, and Tatsunori Hashimoto.
\newblock s1: {Simple} test-time scaling, February 2025.
\newblock arXiv:2501.19393 [cs].

\bibitem[Nie et~al.(2026{\natexlab{a}})Nie, Zhang, Cai, Cheung, Tian, and Han]{nie2026evolvenet}
Jun Nie, Yonggang Zhang, Qianshu Cai, Yiu-ming Cheung, Xinmei Tian, and Bo~Han.
\newblock {EvolveNet}: Collaborative harness evolution for agent self-improvement.
\newblock \emph{arXiv preprint arXiv:2608.04968}, 2026{\natexlab{a}}.

\bibitem[Nie et~al.(2026{\natexlab{b}})Nie, Zhang, Song, Cai, Yu, Guo, Tian, and Han]{nie2026tthe}
Jun Nie, Yonggang Zhang, Jun Song, Qianshu Cai, Dahai Yu, Yike Guo, Xinmei Tian, and Bo~Han.
\newblock {TTHE}: Test-time harness evolution.
\newblock \emph{arXiv preprint arXiv:2607.08124}, 2026{\natexlab{b}}.

\bibitem[Ning et~al.(2026)Ning, Tieu, Fu, Wei, Li, Bei, Zou, Ai, Liu, Li, et~al.]{ning2026code}
Xuying Ning, Katherine Tieu, Dongqi Fu, Tianxin Wei, Zihao Li, Yuanchen Bei, Jiaru Zou, Mengting Ai, Zhining Liu, Ting-Wei Li, et~al.
\newblock Code as agent harness.
\newblock \emph{arXiv preprint arXiv:2605.18747}, 2026.

\bibitem[Pan et~al.(2026)Pan, Zou, Guo, Ni, and Zheng]{pan2026nlah}
Linyue Pan, Lexiao Zou, Shuo Guo, Jingchen Ni, and Hai-Tao Zheng.
\newblock Natural-language agent harnesses.
\newblock \emph{arXiv preprint arXiv:2603.25723}, 2026.

\bibitem[{Qwen Team}(2026{\natexlab{a}})]{qwen3.6-27b}
{Qwen Team}.
\newblock {Qwen3.6-27B}: Flagship-level coding in a {27B} dense model, April 2026{\natexlab{a}}.
\newblock \url{https://qwen.ai/blog?id=qwen3.6-27b}.

\bibitem[{Qwen Team}(2026{\natexlab{b}})]{qwen38}
{Qwen Team}.
\newblock {Qwen3.8-Max}: A new bar for coding and cowork, August 2026{\natexlab{b}}.
\newblock \url{https://qwen.ai/blog?id=qwen3.8}.

\bibitem[Rabanser et~al.(2026)Rabanser, Kapoor, Kirgis, Liu, Utpala, and Narayanan]{rabanser2026towards}
Stephan Rabanser, Sayash Kapoor, Peter Kirgis, Kangheng Liu, Saiteja Utpala, and Arvind Narayanan.
\newblock Towards a science of {AI} agent reliability.
\newblock \emph{arXiv preprint arXiv:2602.16666}, 2026.

\bibitem[Reimers and Gurevych(2019)]{reimers2019sentence}
Nils Reimers and Iryna Gurevych.
\newblock {Sentence-BERT}: Sentence embeddings using {Siamese} {BERT}-networks.
\newblock In \emph{Proceedings of the Conference on Empirical Methods in Natural Language Processing and the 9th International Joint Conference on Natural Language Processing (EMNLP-IJCNLP)}, pages 3982--3992, 2019.

\bibitem[Ren et~al.(2026)Ren, Luo, Yang, Zhu, Huang, Wu, Chou, Ye, Liang, Li, et~al.]{ren2026scaling}
Tao Ren, Weiyao Luo, Hui Yang, Rongzhi Zhu, Xiang Huang, Yuchuan Wu, Bingxue Chou, Jieping Ye, Jiafeng Liang, Yongbin Li, et~al.
\newblock Scaling self-evolving agents via parametric memory.
\newblock \emph{arXiv preprint arXiv:2606.04536}, 2026.

\bibitem[Robeyns et~al.(2025)Robeyns, Szummer, and Aitchison]{robeyns2025sica}
Maxime Robeyns, Martin Szummer, and Laurence Aitchison.
\newblock A self-improving coding agent.
\newblock \emph{arXiv preprint arXiv:2504.15228}, 2025.

\bibitem[Sengupta and Wang(2026)]{sengupta2026harbor}
Biswa Sengupta and Jinhua Wang.
\newblock Harbor: Automated harness optimization.
\newblock \emph{arXiv preprint arXiv:2604.20938}, 2026.

\bibitem[Shinn et~al.(2023)Shinn, Cassano, Gopinath, Narasimhan, and Yao]{shinn2023reflexion}
Noah Shinn, Federico Cassano, Ashwin Gopinath, Karthik Narasimhan, and Shunyu Yao.
\newblock Reflexion: Language agents with verbal reinforcement learning.
\newblock \emph{Advances in Neural Information Processing Systems}, 36:\penalty0 8634--8652, 2023.

\bibitem[Snell et~al.(2025)Snell, Lee, Xu, and Kumar]{snell2025scaling}
Charlie Snell, Jaehoon Lee, Kelvin Xu, and Aviral Kumar.
\newblock Scaling {LLM} test-time compute optimally can be more effective than scaling model parameters.
\newblock In \emph{Proceedings of the International Conference on Learning Representations}, 2025.

\bibitem[Sun et~al.(2020)Sun, Wang, Liu, Miller, Efros, and Hardt]{sun2020testtime}
Yu~Sun, Xiaolong Wang, Zhuang Liu, John Miller, Alexei~A. Efros, and Moritz Hardt.
\newblock Test-time training with self-supervision for generalization under distribution shifts.
\newblock In \emph{Proceedings of the 37th International Conference on Machine Learning}, volume 119 of \emph{Proceedings of Machine Learning Research}, pages 9229--9248. PMLR, 2020.

\bibitem[Tang et~al.(2026)Tang, Rashtchian, Ferng, Tomkins, Juan, and Vu]{tangWikiSkillCompilingAgent2026}
Liyan Tang, Cyrus Rashtchian, Chun-Sung Ferng, Andrew Tomkins, Da-Cheng Juan, and Tu~Vu.
\newblock {WikiSkill}: Compiling agent experience into persistent knowledge for skill evolution, August 2026.
\newblock arXiv:2608.27454 [cs.AI].

\bibitem[Ven et~al.(2025)Ven, Soures, and Kudithipudi]{venContinualLearningCatastrophic2025}
Gido M. van~de Ven, Nicholas Soures, and Dhireesha Kudithipudi.
\newblock Continual learning and catastrophic forgetting, 2025.
\newblock arXiv:2403.05175 [cs.LG].

\bibitem[Wang et~al.(2024)Wang, Xie, Jiang, Mandlekar, Xiao, Zhu, Fan, and Anandkumar]{wangVoyagerOpenEndedEmbodied2023}
Guanzhi Wang, Yuqi Xie, Yunfan Jiang, Ajay Mandlekar, Chaowei Xiao, Yuke Zhu, Linxi Fan, and Anima Anandkumar.
\newblock Voyager: An open-ended embodied agent with large language models.
\newblock \emph{Transactions on Machine Learning Research}, October 2024.
\newblock arXiv:2305.16291 [cs.AI].

\bibitem[Wang et~al.(2026)Wang, Shi, Li, Li, Yu, Yang, Panaganti, Mi, Zhou, et~al.]{wang2026handbook}
Ruhan Wang, Yucheng Shi, Zongxia Li, Zhongzhi Li, Yue Yu, Junyao Yang, Kishan Panaganti, Haitao Mi, Dongruo Zhou, et~al.
\newblock Harness handbook: Making evolving agent harnesses readable, navigable, and editable.
\newblock \emph{arXiv preprint arXiv:2607.13285}, 2026.

\bibitem[Xu et~al.(2026)Xu, Wen, and Li]{xu2026adapt}
Tianshi Xu, Huifeng Wen, and Meng Li.
\newblock Adapting the interface, not the model: Runtime harness adaptation for deterministic {LLM} agents.
\newblock \emph{arXiv preprint arXiv:2605.22166}, 2026.

\bibitem[Yan et~al.(2026)Yan, le~Su, Zhang, Li, Zhang, Zhang, Chen, Bai, and Hu]{yan2026hoh}
Haoyang Yan, Min le~Su, Hangfan Zhang, Zhanhao Li, Chen Zhang, Shao Zhang, Yang Chen, Lei Bai, and Shuyue Hu.
\newblock Harness-of-harness: Multi-day autonomous software development with continual improvement.
\newblock \emph{arxiv preprint arxiv:2609.01481}, 2026.

\bibitem[Yao et~al.(2022)Yao, Zhao, Yu, Du, Shafran, Narasimhan, and Cao]{yao2022react}
Shunyu Yao, Jeffrey Zhao, Dian Yu, Nan Du, Izhak Shafran, Karthik Narasimhan, and Yuan Cao.
\newblock React: Synergizing reasoning and acting in language models.
\newblock \emph{arXiv preprint arXiv:2210.03629}, 2022.

\bibitem[Ye et~al.(2026)Ye, Li, Yang, Liu, Yao, Lv, Xie, An, Li, Kong, Liu, Sui, and Yang]{yeClawEvalTrustworthyEvaluation2026}
Bowen Ye, Rang Li, Qibin Yang, Yuanxin Liu, Linli Yao, Hanglong Lv, Zhihui Xie, Chenxin An, Lei Li, Lingpeng Kong, Qi~Liu, Zhifang Sui, and Tong Yang.
\newblock Claw-{Eval}: Towards trustworthy evaluation of autonomous agents, May 2026.
\newblock arXiv:2604.06132 [cs.AI].

\bibitem[Yu et~al.(2026)Yu, Li, Si, Zhang, Xu, Wang, Dong, Tuo, Zeng, Feng, Wang, Shi, Hu, Yue, Wang, and Yan]{yuDOPDDualOnpolicy2026}
Xinlei Yu, Gen Li, Qingyi Si, Guibin Zhang, Yuqi Xu, Congcong Wang, Shuai Dong, Kaiwen Tuo, Xiangyu Zeng, Kaituo Feng, Qunzhong Wang, Yang Shi, Xiaobin Hu, Xiangyu Yue, Jiaqi Wang, and Shuicheng Yan.
\newblock {DOPD}: Dual on-policy distillation, June 2026.
\newblock arXiv:2606.30626 [cs.AI].

\bibitem[Yuan et~al.(2026)Yuan, Zhou, Xiong, Wu, Sun, Song, Cui, Wang, Wu, Li, et~al.]{yuan2026osworld2}
Mengqi Yuan, Zilong Zhou, Xinzhuang Xiong, Weiming Wu, Jiayang Sun, Jiamin Song, Kaiqian Cui, Bowen Wang, Haoyuan Wu, Yitong Li, et~al.
\newblock Osworld2. 0: Benchmarking computer use agents on long-horizon real-world tasks, 2026.
\newblock arXiv:2606.29537 [cs.AI].

\bibitem[Yuan et~al.(2024)Yuan, Pang, Cho, Li, Sukhbaatar, Xu, and Weston]{yuanSelfRewardingLanguageModels2025}
Weizhe Yuan, Richard~Yuanzhe Pang, Kyunghyun Cho, Xian Li, Sainbayar Sukhbaatar, Jing Xu, and Jason Weston.
\newblock Self-rewarding language models.
\newblock In \emph{Proceedings of the International Conference on Machine Learning}, March 2024.

\bibitem[Zelikman et~al.(2022)Zelikman, Wu, Mu, and Goodman]{zelikmanSTaRBootstrappingReasoning2022}
Eric Zelikman, Yuhuai Wu, Jesse Mu, and Noah~D. Goodman.
\newblock {STaR}: Bootstrapping reasoning with reasoning.
\newblock In \emph{Proceedings of Advances in Neural Information Processing Systems}, May 2022.

\bibitem[Zhang et~al.(2026{\natexlab{a}})Zhang, Lu, Xie, Zhu, Wang, Xie, Yu, Liu, Sun, Li, et~al.]{zhang2026jitagent}
Guibin Zhang, Leo Lu, Fangzhou Xie, Kang Zhu, Junhao Wang, Zhifei Xie, Zhaochen Yu, Zihang Liu, Zhongxiang Sun, Qiankun Li, et~al.
\newblock {JIT-Agent}: Scaling harness intelligence via just-in-time harness evolution.
\newblock \emph{arXiv preprint arXiv:2608.25593}, 2026{\natexlab{a}}.

\bibitem[Zhang et~al.(2026{\natexlab{b}})Zhang, Zhang, Li, Zhang, Chen, Zhang, Bai, and Hu]{zhang2026selfharness}
Hangfan Zhang, Shao Zhang, Kangcong Li, Chen Zhang, Yang Chen, Yiqun Zhang, Lei Bai, and Shuyue Hu.
\newblock Self-harness: Harnesses that improve themselves.
\newblock \emph{arXiv preprint arXiv:2606.09498}, 2026{\natexlab{b}}.

\bibitem[Zhang et~al.(2026{\natexlab{c}})Zhang, Hu, Lu, Lange, and Clune]{zhang2026dgm}
Jenny Zhang, Shengran Hu, Cong Lu, Robert Lange, and Jeff Clune.
\newblock Darwin {G\"{o}del} machine: Open-ended evolution of self-improving agents.
\newblock In \emph{Proceedings of the International Conference on Learning Representations}, 2026{\natexlab{c}}.

\bibitem[Zhao et~al.(2024)Zhao, Huang, Xu, Lin, Liu, and Huang]{zhao2024expel}
Andrew Zhao, Daniel Huang, Quentin Xu, Matthieu Lin, Yong-Jin Liu, and Gao Huang.
\newblock Expel: {LLM} agents are experiential learners.
\newblock In \emph{Proceedings of the AAAI Conference on Artificial Intelligence}, volume~38, pages 19632--19642, 2024.

\end{thebibliography}
